\documentclass[a4paper,10pt,reqno,oneside]{amsart}
\usepackage[foot]{amsaddr}
\usepackage{graphicx} 
\graphicspath{{figures/}}
\usepackage[colorinlistoftodos]{todonotes} 
\usepackage{subcaption}
\usepackage[colorlinks=true,linkcolor=black,urlcolor=black,citecolor=black]{hyperref}
\usepackage{amsfonts}
\usepackage{amssymb,amsmath}
\usepackage{amsthm}
\usepackage{fixmath}

\usepackage{algorithm}
\usepackage{algpseudocode}
\usepackage{enumerate}
\usepackage{multirow}

\usepackage{mathtools}
\usepackage{xcolor}
\usepackage{subcaption}
\numberwithin{equation}{section}
\numberwithin{figure}{section}
\numberwithin{equation}{section}

\newcommand{\x}{{{x}}}
\newcommand{\R}{\mathbb{R}}
\newcommand{\xx}{{y}}
\renewcommand{\d}{\mathop{}\!\mathrm{d}}

\newcommand{\bbZ}{\mathbb{Z}}
\newcommand{\cker}{c_{\gamma}}
\newcommand{\cpot}{c_F}
\newcommand{\thetalog}{\theta_c}
\newcommand{\kernel}{\gamma}

\renewcommand{\tt}{\Delta t}
\newcommand{\D}{\Omega}

\newcommand{\Sh}{S_h}
\newcommand{\Shw}{\hat{S}_h}
\newcommand*{\normd}[1]{{\left\lVert#1\right\rVert}_2}

\newcommand*{\innerd}[1]{{\left\langle#1\right\rangle}}
\newcommand{\astcirc}{%
  \mathrel{\ooalign{\hfil$\ast$\hfil\cr\hfil\textcircled{\hfil}\hfil}}%
}

\theoremstyle{remark}

\begin{document}
\title[Deep Learning Method for Nonlocal Allen-Cahn and Cahn-Hilliard Models]{An End-to-End Deep Learning Method for Solving Nonlocal Allen-Cahn and Cahn-Hilliard Phase-Field Models}

\author[Y.  Geng, O. Burkovska, Lili Ju,  Guannan Zhang, Max Gunzburger]{Yuwei Geng${}^{1}$,  Olena Burkovska${}^{2}$,  Lili Ju${}^{1}$,  Guannan Zhang${}^{2}$,  Max Gunzburger${}^{3,4}$}
\address{${}^1$ Department of Mathematics, University of South Carolina, USA}
\address{${}^2$ Computer Science and Mathematics Division, Oak Ridge National Laboratory,USA}
\address{${}^3$ Department of Scientific Computing, Florida State University,USA}
\address{${}^4$ Oden Institute for Computational Engineering and Sciences, University of Texas at Austin,USA}
\thanks{
This manuscript has been authored by UT-Battelle, LLC, under contract DE-AC05-00OR22725 with the US Department of Energy (DOE). The US government retains and the publisher, by accepting the article for publication, acknowledges that the US government retains a nonexclusive, paid-up, irrevocable, worldwide license to publish or reproduce the published form of this manuscript, or allow others to do so, for US government purposes. DOE will provide public access to these results of federally sponsored research in accordance with the DOE Public Access Plan.
}


\maketitle

\begin{abstract}
We propose an efficient end-to-end deep learning method for solving nonlocal Allen-Cahn (AC) and Cahn-Hilliard (CH) phase-field models.
One motivation for this effort emanates from the fact that 
discretized partial differential equation-based AC or CH phase-field models result in diffuse interfaces between phases, with the only recourse for remediation is to severely refine the spatial grids in the vicinity of the true moving sharp interface whose width is determined by a grid-independent parameter that is substantially larger than the local grid size. In this work, we introduce non-mass conserving nonlocal AC or CH phase-field models with regular, logarithmic, or obstacle double-well potentials. Because of non-locality, some of these models feature totally sharp interfaces separating phases. The discretization of such models can lead to a transition between phases whose width is only a single grid cell wide. Another motivation is to use deep learning approaches to ameliorate the otherwise high cost of solving discretized nonlocal phase-field models. To this end, loss functions of the customized neural networks are defined using the residual of the fully discrete approximations of the AC or CH models, which results from applying a 
Fourier collocation method
and a temporal semi-implicit approximation. To address the long-range interactions in the models, we tailor the architecture of the neural network by incorporating a nonlocal kernel as an input channel to the neural network model. 
We then provide the results of extensive computational experiments to illustrate the accuracy, structure-preserving properties, predictive capabilities, and cost reductions of the proposed method. 

\end{abstract}

\section{Introduction}
Phase-field modeling is a robust method utilized to describe a wide range of physical processes, including nucleation and solidification in materials science \cite{kobayashi1993modeling,wang1993thermodynamically}, {fracture~\cite{DiehlLipton2022}}, image processing \cite{bertozzi2006inpainting,garcke2018cahn}. These models, particularly local phase-field models of the Allen-Cahn or Cahn-Hilliard type, rely on local partial derivatives and are commonly employed due to their versatility. Nevertheless, a significant limitation of these models is their tendency to produce diffuse interfaces which poses computational challenges, especially when thin or sharp interfaces are intrinsic features of the models.
For instance, in material solidification applications the interface between two immiscible substances is typically represented by a moving sharp interface that separates the distinct phases. In practice, implementing such models is difficult due to the complexities associated with the coupling at the moving interface. Instead, a phase-field method can be adopted wherein the sharp interface is approximated by a diffuse interface. This approach eliminates the need for explicit interface tracking and allows for the simulation of the entire phase transition process. However, accurately resolving very thin diffuse interfaces to approximate a sharp interface necessitates the use of very fine meshes and/or more structurally complex models, which greatly increases the computational demands. 

To ameliorate these deficiencies of local phase-field models based on partial differential equations (PDEs), we consider {\em nonlocal phase-field models}.
The essential difference between nonlocal phase-field models and local phase-field models is that the former involves a derivative-free integral operator that allows for pairs of points separated by a finite distance to interact with each other, in contrast to PDE models for which pairs of points interact only within the infinitesimal neighborhoods needed to define derivatives. 
Nonlocal models lead to improved physical realism through a better representation of long-range interactions and correlations, enhanced numerical stability when dealing with sharp interfaces, and superior multi-scale modeling capabilities. Nonlocal models also reduce mesh dependencies, more easily incorporate statistical effects and heterogeneities, and provide a more accurate representation of interfacial phenomena. These features collectively result in a more comprehensive and physically accurate framework for simulating complex material systems, particularly those involving multiple phases, long-range interactions, and multi-scale phenomena. On the other hand, the nonlocal phase-field model is much more difficult to solve numerically due to the existence of the nonlocal integral operator, which makes deep learning methods an attractive alternative.

The practical applications of phase-field models in materials science and engineering demand that deep learning models for AC or CH equations possess the capability to accurately and efficiently capture moving sharp interfaces across a diverse range of initial conditions. This requirement stems from the need to simulate complex phase transformations and microstructural evolution in real-world materials systems. The challenge lies in developing neural network architectures that can not only handle the nonlinear dynamics of these equations but also maintain high fidelity in representing sharp interfaces without excessive computational cost. Moreover, these models must be generalizable enough to perform well on various initial configurations, including those not seen during training. Despite significant progress in applying machine learning to PDEs, achieving this ideal balance of accuracy, efficiency, and versatility for phase-field models remains an open challenge in the field, with current solutions often trading off one aspect for another. For example, the standard Physics-Informed Neural Networks (PINNs) face significant challenges in accurately resolving sharp interfaces, as noted by \cite{de2024physics,krishnapriyan2021characterizing,mao2023physics,YOU2022111536}. These challenges arise due to the neural network's inherent tendency to smooth out steep gradients. 
Adaptive sampling methods were proposed to dynamically select collocation points in regions with steep gradients. For example,  \cite{zhao2020solving} proposed an adaptive sampling strategy that selects collocation points and implements a time-adaptive strategy within sub-time intervals to capture sharp transitions when solving AC and CH equations, but this method cannot generalize to a broad range of initial conditions. Operator Learning-enhanced PINN (OL-PINN) \cite{lin2023operator} involves pre-training a deep neural network (DeepONet) on a class of smooth problems before integrating them with PINNs to better capture sharp interface features. However, the PINN model still needs to be retrained for each initial condition, which makes it computationally expensive for the nonlocal phase-field model under consideration. {Alternative approaches based on neural operators and their generalizations~\cite{goswami2023physics,kovachki2023neural,lu2021learning,yu2024nonlocal} have been actively explored in the literature. Those methods, often do not require the knowledge of the underlying physics and are based on a data-driven loss. 
In~\cite{oommen2022learning}, the authors present a framework that combines a convolutional autoencoder architecture with a DeepONet to learn the microstructure evolution of the phase-field variable. In contrast to our approach, this method employs a more complex architecture and is entirely data-driven, as it relies on a data-driven loss. To increase the accuracy a hybrid approach is also adopted.}

Neural networks proposed by \cite{bretin2022learning} emulate the action of diffusion and reaction operators, where convolutional layers approximate the diffusion process and multilayer perceptrons capture nonlinear reaction dynamics, enabling efficient simulations of interface evolution in AC equations. Similarly, ``Phase-Field DeepONet'' \cite{li2023phase} is a physics-informed deep operator neural network framework designed for rapid simulations of pattern formation in systems governed by gradient flows, validated on the local AC and CH equations. Although these methods leverage operator learning and do not require retraining for different initial conditions, they are not directly applicable to nonlocal problems, which involve interactions across distant points in the domain.

In this paper, we propose a more general framework, referred to as the {\em nonlocal phase-field network} (NPF-Net) for solving nonlocal AC and CH equations with both smooth and sharp interfaces. Building on our previous work in  \cite{geng2024deep}, NPF-Net aims to learn fully discrete nonlocal operators that achieve numerical approximation accuracy between adjacent time steps for the target equations. The loss functions are formulated using a residual discretized by a Fourier collocation method in space with first- or second-order semi-implicit time stepping schemes.
In this case, no training data needs to be generated a priori to train the NPF-Net model. The key contributions of this work are threefold. First, we present novel network architectures tailored for solving nonlocal systems, with a specific focus on nonlocal AC and CH equations. This design enables more accurate and efficient simulations of long-range interactions in materials. Second, our model successfully handles equations with non-smooth potential functions, a challenging aspect in both traditional numerical methods and machine learning approaches. This capability allows for more realistic modeling of complex material behaviors. Lastly, and perhaps most significant, NPF-Net demonstrates the ability to accurately describe sharp interface dynamics, a critical feature for simulating phenomena such as phase boundaries and other discontinuous processes in materials.

The remainder of the paper is organized as follows. In Section \ref{sec:DFF}, we introduce space and time discretization methods for the nonlocal AC and CH equations. We then discuss, in Section \ref{sec:NN}, the loss function, training strategy, and network architectures used for simulating the dynamics of the AC and CH equations using deep learning methods. Hyperparameter optimization, numerical experiments, and comparisons are presented in Section \ref{sec: Results} which demonstrate the superior performance of our proposed method, including an evaluation of its prediction and generalization capabilities under various scenarios in two and three dimensions. Finally, concluding remarks are provided in Section \ref{sec:conclusion}.

\section{Nonlocal Allen-Cahn and Cahn-Hilliard phase-field models}\label{sec:prob}

In this section, we provide a brief introduction of the nonlocal AC and CH models for three types of potential functions and the use of deep learning strategies to lessen the costs of computational simulations. 

Let $\Omega\subset \mathbb{R}^d$, $d=2, 3$, be a bounded domain. Specifically, we consider the nonlocal problem given by
\begin{equation}
\begin{cases}
\partial_t u +A w = 0, \\
w=Bu+\partial F(u),
\end{cases}
\label{eq:general}
\end{equation}
and equipped with periodic boundary conditions and a suitable initial condition $u(0,x)=u_0(x)$. Here, $B$ denotes a nonlocal diffusion operator defined as
\begin{align*}
Bu&:=\int_{\Omega} \gamma(x-y)\left(u(x)-u(y)\right)dy=(\gamma \ast 1)u(x)-(\gamma \ast u)(x) = \cker u - \kernel*u, 
\end{align*}
where $\gamma(y-x)$ is a nonlocal interaction kernel which is assumed to be integrable, positive, and symmetric (i.e., $\gamma(y-x)=\gamma(x-y)$) with $\cker = \kernel*1 = \int_{\D}\kernel(\x-\xx)\d\x$ and with $*$ denoting the convolution operator. We define the operator $A=I-\beta\upDelta$ with $\beta\geq 0$, where $\beta= 0$ corresponds to a non-mass conserving nonlocal AC model and $\beta>0$ to a non-mass conserving nonlocal CH model. Under appropriate conditions on the kernel, we can recover a local operator in the limit of vanishing nonlocal interactions, i.e., $Bu\to -\epsilon^2\upDelta u$ with a local interface parameter $\epsilon^2 = \frac{1}{2 n}\int_{\Omega}|\zeta|^2\gamma(|\zeta|)d\zeta$ (see, e.g.,~\cite{Davoli2021}).
We also define an associated nonlocal energy, which is an analogous of the local Ginzburg-Landau energy:
\begin{equation}
    E(u)=\frac{1}{4}\int_{\Omega}\int_{\Omega}(u(\x)-u(\xx))^2\gamma(\x-\xx)\d\x\d\xx +\int_{\Omega}F(u(\x))\d\x.\label{eq:GL_energy}
\end{equation}

\begin{figure}[h!]
    \centering
\includegraphics[width=0.32\textwidth]{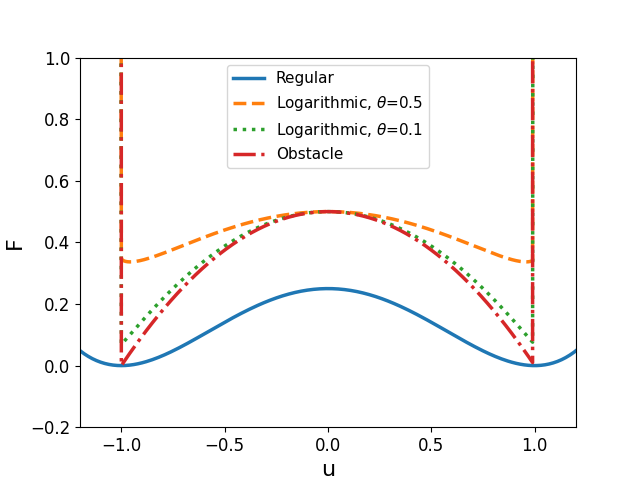}
        \includegraphics[width=0.32\textwidth]{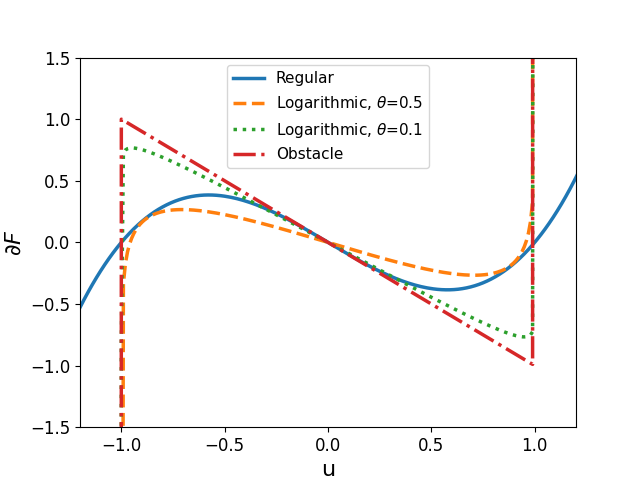}
        \includegraphics[width=0.32\textwidth]{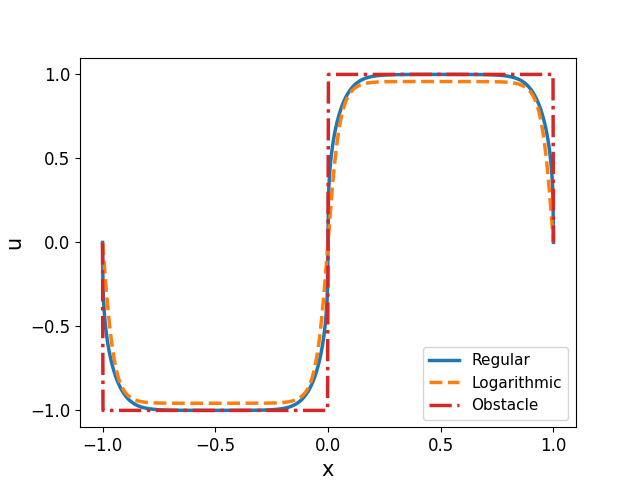}
    \caption{The plots of the different potential functions (left), their corresponding subdifferentials (derivatives) (middle), and the solutions of \eqref{eq:general} corresponding to those potentials. Here we use $\cpot=1$, $\xi=\cker-\cpot=0$, and the solution (right) for the logarithmic potential is depicted for $\thetalog=0.5$.}
    \label{fig:potential}
\end{figure}

In \eqref{eq:general} and~\eqref{eq:GL_energy}, $F$ denotes a double-well potential with $\partial F$ denoting a generalized subdifferential of $F$. We consider the three common types of potentials, namely the regular,  logarithmic, and obstacle potentials. Those potentials can be expressed as
\begin{equation}
    F(u)=\frac{\cpot}{2}(1-u^2)+\psi(u),\quad \cpot>0,
\end{equation}
where a {\em smooth regular potential} $\psi(u)$ is defined  as
\begin{equation}
    \psi_{\rm reg}(u)=\frac{\cpot}{4}\left(u^4-1\right),\quad \cpot>0, \label{regular_potential}
\end{equation}
a {\em logarithmic potential} is defined as,  for $0 <\theta < \cpot$,
\begin{equation}
    \psi_{\rm log}(u) =\ 
    \begin{cases}\displaystyle
        \frac{\thetalog}{2}\left((1+u)\ln(1+u)+(1-u)\ln(1-u)\right),\quad\text{if}\; {u\in[-1,1]},\\
+\infty,\qquad\qquad\qquad\qquad\qquad\qquad\qquad\quad \text{\,else},
    \end{cases}\label{eq:pot_logarithm}
    \end{equation}
where the values at $u=\pm 1$ are defined by taking a limit of the function resulting in $\thetalog \ln(2)$. Eventually, a {\em non-smooth obstacle potential} is defined as the convex indicator function
\begin{equation}
     \psi_{\rm obs}(u)=\mathbb{I}_{[-1,1]}(u)= 
    \begin{cases}
        0,~\text{\quad\,\,\, if } u\in[-1,1],\\
        +\infty, \text{\quad else.}
    \end{cases}
\end{equation}
The corresponding differential or subdifferential, as the case may be, of $F(u)$ is then defined as
\begin{equation}
    \partial F(u)=-\cpot u + \partial\psi(u),\label{eq:subdiff_potential}
\end{equation}
with $\partial\psi(u)$ given by
\begin{subequations}
       \begin{align}
        \partial\psi_{\rm reg}(u) = &\ \psi^\prime(u) =   \left\{\cpot u^3\right\},\label{eq:subdiff_reg}
        \\
 \partial\psi_{\rm log} (u)=&\ 
\begin{cases} 
\emptyset, &\text{\quad if } u\not\in[-1,1],\\
\displaystyle
\left\{\frac{\thetalog}{2}\left(\ln(1+u)-\ln(1-u)\right)\right\}, & \text{\quad if }u\in (-1,1),
\end{cases} \label{eq:subdiff_log}\\              
        \partial\psi_{\rm obs}(u)=&\ \partial \mathbb{I}_{[-1, 1]}(u)=
\begin{cases} (-\infty, 0], & \text{\quad if } u=-1,\\ 
0, & \text{\quad if } u \in (-1, 1),\\
[0, \infty), & \text{\quad if } u=1.
\end{cases}\label{eq:subdiff_obs}
 \end{align}
\label{eq:subdiff_all}
 \end{subequations}\\
We note that for the case of regular and logarithmic potentials ($u\in (-1,1)$) the subdifferential $\partial\psi$ reduces to a classical derivative of those potentials. Illustrations of the three potential functions and their subderivatives or subdifferentials are given in Figure \ref{fig:potential}.

Using the above definitions for the subderivative or subdifferential, we can rewrite the model in \eqref{eq:general} as\footnote{{By a slight abuse of notation and for the sake of exposition, we use an equality in~\eqref{eq:general1} and throughout the paper. However, this should be understood as an inclusion: $w-\xi u+\gamma\ast u -\partial \psi(u) \ni 0$, which reduces to the equality in the cases of the regular and logarithmic potentials.} }
\begin{equation}
\begin{cases}
\partial_t u +A w = 0,\\
w-\xi u+\gamma\ast u -\partial \psi(u) = 0,
\end{cases}
\label{eq:general1}
\end{equation}
where $\xi=\cker-\cpot$ is a nonlocal interface parameter and plays an analogous role to that of the local interface parameter $\epsilon$ when $Bu\approx -\epsilon^2\upDelta u$. Here, larger values of $\xi$ result in a more diffuse interface, as noted in \cite{burkovska2021nonlocal}. However, unlike the local setting for which a sharp interface is achieved only in the limit $\epsilon\to 0$, in nonlocal settings well-defined sharp interfaces with distinct pure phases can emerge when $\xi=0$ \cite{burkovska2021nonlocal}.
Alternatively, we can set $w = -A^{-1}(\partial_t u) = G(\partial_t u)$, where $G$ denotes the Green's function related to the operator $A$, so that then we obtain
\begin{eqnarray}
    -G(\partial_t u)+ \xi u-\gamma\ast u +\partial \psi(u)= 0.\label{eq:general3}
\end{eqnarray}

The choice of potential greatly affects the properties of the solution such as interface sharpness. 
The obstacle potential produces the sharpest interface, whereas the regular potential results in the most diffuse one. For the logarithmic potential, sharpness is influenced by $\theta$, with smaller $\theta$ yielding a sharper interface, because as $\theta$ decreases, the logarithmic potential function increasingly resembles the non-smooth obstacle potential function. 
These findings align with existing results in the literature which indicate that solutions of the nonlocal phase-field model with the obstacle potential can exhibit jump discontinuities, resulting in pure phases without intermediate transitions, during a whole time evolution (for the Cahn-Hilliard model) or at a steady-state (for the Allen-Cahn model)~\cite{burkovska2023non,burkovska2021nonlocal}.
In contrast, for the regular potential, interfaces in the solution, although possibly admitting discontinuities, are not jump discontinuities but may feature narrow transition regions between phases as has been studied in~\cite{DuJuLiQiao2019max,DuYang2016,DuYang2017} for the Allen-Cahn solutions at a steady-state.  To illustrate this, 
in Figure \ref{fig:potential}, in addition to  plotting three distinct potential functions (left) and the corresponding subdifferentials (middle), the plots corresponding to steady-state solutions for $\xi=0$, with the initial condition 
$u_0=0.1sin(2*\pi x)$, are shown in Figure~\ref{fig:potential} (right). 

\section{Discretization of the nonlocal Allen-Cahn and Cahn Hilliard equations} \label{sec:DFF}

In this section, we introduce full spatial and temporal discretizations of the nonlocal AC and CH equations governed by \eqref{eq:general1} which will be used in Section \ref{sec:loss} to define the loss functions. We consider first and second-order time stepping schemes combined with the Fourier collocation approach for the spatial discretization. 
For ease of illustration, we define the domain $\Omega=(-X,X)\times(-Y,Y)\subset\R^2$  and assume that $u$ satisfies periodic boundary conditions.

\subsection{\textbf{Temporal discretization} }\label{sec:temporal_AC}

We adopt first and second order time stepping schemes to discretize the model in \eqref{eq:general1}. We choose a uniform time step size $\Delta t>0$ and set $t_n = n\Delta t$ for $n=0,\dots,N$, and also denote $U_n=u(x,y,t_n)$ and $U^{i,j}_{n/2}:=(U_{n}+U_{n+1})/2$. For additional details about the study of the time-stepping schemes adopted here, we refer to~\cite{burkovska2023non,BDJ2025,burkovska2021nonlocal,BM2024}.

\subsubsection{Temporal discretization for the nonlocal Allen-Cahn model}

For the Allen-Cahn case $(\beta=0)$ we use a first order semi-implicit discretization with an explicit discretization of the convolution term, namely, given $U_0$, we seek $U_{n+1}$ for $n \geq 0$ such that
\begin{equation}
    \frac{1}{\tt}{(U_{n+1}-U_{n})}+\xi U_{n+1}-\gamma*U_{n}+\partial  \psi(U_{n+1})=0.\label{eq:AC_tk1}
\end{equation} 
Analogously, for the second-order scheme we use for the regular or logarithmic potentials is given by
\begin{equation}
    \frac{1}{\tt}{(U_{n+1}-U_{n})}+\xi U_{n/2}-\gamma*U_{n/2}+\frac{1}{2}\big({  \psi^\prime(U_{n+1})+ \psi^\prime(U_{n})}\big)=0,\label{eq:AC_tk2_regular}
\end{equation} 
and for the obstacle potential in \eqref{eq:subdiff_obs} we use
\begin{equation}
    \frac{1}{\tt}{(U_{n+1}-U_{n})}+\xi U_{n/2}-\gamma*U_{n/2}+\partial  \psi(U_{n+1})=0.\label{eq:AC_tk2_obstacle}
\end{equation}
We also comment on the solution algorithms for the above systems. For the case of the regular or logarithmic potentials, the resulting systems are nonlinear and we can adopt, e.g., a fixed-point iteration to solve them. The situation is more delicate for the case of an obstacle potential, as the system is nonlinear and non-smooth. However, the structure of the problem in \eqref{eq:general1} allows us to derive a characterization of the solution in terms of a representation formula given by proximal operators~\cite{BM2024}. Specifically, for example, for the first order scheme in \eqref{eq:AC_tk1},  the solution $U_n$ in \eqref{eq:AC_tk1} is equivalent to  
\begin{eqnarray}
    U_{n+1} = {\rm prox}_{\frac{1}{\lambda}\psi}\left(\frac{1}{\lambda}\left(\kernel*U_{n}+\frac{1}{\tt}U_{n}\right)\right),\quad \lambda=\xi+1/\tt>0,\label{eq:prox_representation_tk1}
\end{eqnarray}
and the proximal operator is defined as
${\rm prox}_{\eta\psi}:=\left(I+\eta\partial\psi\right)^{-1}$, $\eta>0$~\cite{boyd2013proximal}. For the regular or obstacle potential the proximal operator has an analytic representation obtained by Cardano's formula~\cite{witula2010cardano} or by pointwise projection, respectively. That is, 
for the regular potential, the solution of \eqref{eq:AC_tk1} has the form
\begin{equation}
    U_{n+1} = \sqrt[3]{\zeta_n+\sqrt{\zeta_n^2 + \left({\lambda}/{3\cpot}\right)^3}} + \sqrt[3]{\zeta_n-\sqrt{\zeta_n^2 + \left({\lambda}/{3c_F}\right)^3}},\label{eq:prox_formula_reg_order1}
\end{equation}
where $\zeta_n = \left(\kernel*U_{n}+\frac{1}{\tt}U_{n}\right)/(2\cpot)$
and for the obstacle potential, the proximal operator is a pointwise projection onto an admissible set and \eqref{eq:prox_representation_tk1} reduces to 
\begin{eqnarray}
    U_{n+1} = P_{[-1,1]}\left(\frac{1}{\lambda}\left(\kernel*U_{n}+\frac{1}{\tt}U_{n}\right)\right),\quad \lambda=\xi+1/\tt>0,\label{eq:proj_representation_tk1}
\end{eqnarray}
where $P_{[-1,1]}(\eta)=\eta$ if $|\eta|<1$ and $P_{[-1,1]}(\eta)=\pm 1$ if $\eta\geq 1$ or $\eta\leq -1$, respectively. Analogously, for the second-order scheme the solution of Eq.~\eqref{eq:AC_tk2_regular} and Eq.~\eqref{eq:AC_tk2_obstacle} has the form
\begin{eqnarray}
    U_{n+1} = {\rm prox}_{\frac{1}{2\lambda}\psi}\left(\frac{1}{\lambda}\left((\lambda-\xi)U_{n}+\kernel*U_{n/2}-\frac{1}{2}\psi^*\left(U_{n}\right)\right)\right),\quad \lambda=\xi/2+1/\tt>0,\label{eq:prox_representation_tk2}
\end{eqnarray}
where $\psi^*(u^{k-1})=\psi^\prime(u^{k-1})$ for the regular and logarithmic potentials and $\psi^*(u^{k-1})=0$ for the obstacle potential. Furthermore, for the obstacle potential this further reduces to 
\begin{eqnarray}
    U_{n+1} = P_{[-1,1]}\left(\frac{1}{\lambda}\left((\lambda-\xi)U_{n}+\kernel*U_{n/2}\right)\right),\quad \lambda=\xi/2+1/\tt>0.\label{eq:proj_representation_tk2_obs}
\end{eqnarray}
Whereas the first order scheme involves only explicit evaluation of the convolution that can be performed rather efficiently, the accuracy is only of first order. On the other hand, realizing the second-order scheme in \eqref{eq:AC_tk2_regular} and \eqref{eq:AC_tk2_obstacle} or equivalently \eqref{eq:prox_representation_tk2} and \eqref{eq:proj_representation_tk2_obs}
requires nonlinear iterative solvers  
such as the Picard iteration. For example, the Picard iteration for \eqref{eq:proj_representation_tk2_obs} proceeds as follows: initialize $U_{n+1,(0)}$ = $U_n$ and for $m=0,1,\ldots$, update $U_{n+1, (m+1)}$ by solving 
\begin{equation}
    U_{{n+1},m+1} =P_{[-1,1]} \left(\frac{1}{\lambda} \left((\lambda-\xi)U_{n}+\kernel*U_{n/2,m}\right)\right), \quad \lambda=\xi/2+1/\tt, \label{eq:AC_iter_1}
\end{equation}
where $ \kernel*U_{n/2,m}=\kernel\left(U_{n+1,m}+U_n\right)/2$ and the iteration continues until $U_{n+1} = U_{N+1,m^*}$ satisfies a predefined convergence criterion.

\subsubsection{Temporal discretization for the nonlocal CH equation} \label{sec:temporal_CH}

We briefly outline the temporal discretization for the Cahn-Hilliard model in \eqref{eq:general1} ($\beta>0$). In particular, we consider a first-order semi-implicit scheme, namely, given $U_0=u(0)$ we seek $U_{n+1}$, $n\geq 0$ such that
\begin{equation}
\begin{cases}
\frac{1}{\tt}\left(U_{n+1}-U_n\right) +A W_{n+1} = 0,\\
W_{n+1}-\xi U_{n+1}+\gamma*U_{n-1} -\partial \psi(U_{n+1}) = 0,
\end{cases}
\label{eq:general1_tk1}
\end{equation}
or alternatively
\begin{equation}
   - \frac{1}{\tt}G(U_{n+1}-U_{n})+\xi U_{n+1}-\gamma*U_{n}+\partial \psi(U_{n+1})=0.\label{eq:CH_tk1}
\end{equation}
For the regular or logarithmic potentials, the system in \eqref{eq:general1_tk1} can be solved directly, e.g., using a fixed-point iteration. 
Alternatively, we resort to the formulation in \eqref{eq:CH_tk1} where we also adopt an equivalent representation of the solution through the proximal operators~\cite{BM2024}, that is, the solution of \eqref{eq:CH_tk1} is equivalently expressed as
\begin{equation}
    U_{n+1} = {\rm prox}_{\frac{1}{\lambda}\psi}\left(\frac{1}{\lambda}\left(\frac{1}{\tt}\left(G+cI\right)\left(U_{n+1}-U_{n}\right)+\kernel*U_{n} + \frac{c}{\tt}U_{n}\right)\right),\quad \lambda=\xi+\frac{c}{\tt}>0,
    \label{eq:CH_prox_represent_tk1_obs}
\end{equation}
where for the regular potential the proximal operator can be computed using Cardano's formula and for the obstacle potential this leads to the projection formula
\begin{equation}
    U_{n+1} = P_{[-1,1]}\left(\frac{1}{\lambda}\left(\frac{1}{\tt}\left(G+cI\right)\left(U_{n+1}-U_{n}\right)+\kernel*U_{n} + \frac{c}{\tt}U_{n}\right)\right),\quad \lambda=\xi+\frac{c}{\tt}>0,
    \label{eq:CH_proj_represent_tk1_obs}
\end{equation}
where we have also added a stabilization term $\pm\frac{c}{\tt}\left(U_{n+1}-U_{n}\right)$, $c>0$, in~\eqref{eq:CH_tk1}; see~\cite{BDJ2025} for additional details. We use the above representations to compute solutions, where, similarly as in~\eqref{eq:AC_iter_1}, we perform a fixed point iteration to resolve implicit terms in~\eqref{eq:CH_prox_represent_tk1_obs}.

We also comment that there exist alternative time-stepping methods including unconditionally stable schemes for the nonlocal Cahn-Hilliard or Allen-Cahn equations, see, e.g.,~\cite{du2018stabilized,DuJuLiQiao2019max,guan2017convergence,guan2014second}, which can also be adopted in our learning framework. However, those approaches are focused solely on smooth potentials, are not applicable for the non-smooth obstacle potential.

\subsection{\textbf{Spatial discretization}}\label{sec: Fully_discrete}

{Several numerical methods exist for spatial discretizations of the nonlocal Allen-Cahn or Cahn-Hilliard models. Mostly the schemes are designed for smooth systems involving regular or logarithmic potentials. The schemes include finite difference discretizations~\cite{bates2009numerical,guan2017convergence,guan2014second}, spectral methods~\cite{ainsworth2017,DuYang2016,DuYang2017}, finite elements~\cite{acosta2021numerical}, etc; see also~\cite{DuFeng2020}. An efficient method, if applicable, is to use spectral or collocation approaches for the convolution operators, where the convolution is approximated by the Fourier transform with FFT implementation. This allows for efficient evaluation of the convolution, particularly in high dimensions. 
Here, we adopt a Fourier collocation method as described in~\cite{BM2024,du2018stabilized}. }

We define the discretization of $\Omega$ at collocation points which are equidistantly spaced in each dimension:
\begin{align*}
    \Omega_h = \left\{(x_i,y_j) : x_i=-X+ih_x, y_j=-Y+jh_y,\quad 
    0\le i\le N_x-1, ~0\le j\le N_y-1\right\},
\end{align*}
with uniform mesh sizes $h_x = 2X/N_x$, $h_y=2Y/N_y$,
where $N_x$ and $N_y$ are even numbers, and we set $h = \max\{h_x,h_y\}$. 
We next introduce index sets for the spatial and frequency domains:
\begin{eqnarray*}
    \Sh &=& \{(i,j)\in\bbZ^2\colon 0\le i\le N_x-1,\quad 0\le j\le N_y-1\},\\
    \Shw &=& \left\{(l,m)\in\bbZ^2\colon -\frac{N_x}{2}+1\le l\le \frac{N_x}{2},~-\frac{N_y}{2}+1\le m\le \frac{N_y}{2}\right\}.
\end{eqnarray*}
For all grid periodic functions $U,V\colon\Omega_h\to\mathbb{R}$, we define  
the discrete $L^2$ inner product $\innerd{\cdot,\cdot}$ and norm $\normd{\cdot}$
\begin{equation*}
\innerd{U,V}=h_xh_y\sum_{(i,j)\in\Sh}U^{ij}V^{ij},\qquad \normd{U}=\sqrt{\innerd{U,U}},
\end{equation*}
where $U^{ij}=u(x_i,y_j)$.
By $\astcirc$ we denote a discrete circular convolution for the periodic functions $U,V$:
\begin{equation*}
    (U\astcirc V)^{ij} = h_xh_y\sum_{(p,q)\in\Sh}U^{i-p,j-q}V^{pq},~~~~~(i,j)\in\Sh.\label{convolution_def}
\end{equation*}
We employ a discrete Fourier transform using FFT to efficiently compute the discrete convolution with $\mathcal{O}(N\log N)$ complexity. To this end, we define a discrete Fourier transform
\begin{eqnarray}
    \widehat{U}^{lm}=(\mathcal{F}_NU)^{lm}=\sum_{(i,j)\in\Sh}{U^{ij}}e^{- i\pi\left(\frac{lx_i}{X} + \frac{my_j}{Y}\right)},\quad(l,m)\in\Shw .\label{f_tranform}
\end{eqnarray}
and the corresponding inverse discrete Fourier transform 
\begin{eqnarray}
   {U^{ij}}=\left(\mathcal{F}_N^{-1}\widehat{U}\right)^{ij}=\frac{1}{N_xN_y}\sum_{(l,m)\in \widehat{S}_h}\widehat{U}^{lm}_ne^{ i\pi\left(\frac{lx_i}{X} + \frac{my_j}{Y}\right)},\quad (i,j)\in\Sh.\label{f_inverse}
\end{eqnarray}
Then, taking into account that
\begin{eqnarray*}
    \widehat{(U\astcirc V)}^{lm} = h_xh_y\widehat{U}^{lm}\widehat{V}^{lm},\quad (l,m)\in\Shw,
\end{eqnarray*}
we have that
    \begin{align*}
        (U \astcirc V)^{ij}=\left(\mathcal{F}_N^{-1}\widehat{(U\astcirc V)}\right)^{ij}=
        \frac{h_xh_y}{N_xN_y}\sum_{(l,m)\in \Shw}\widehat{U}^{lm}\widehat{V}^{lm}e^{ i\pi\left(\frac{lx_i}{X} + \frac{my_j}{Y}\right)},~~~~~(i,j)\in S_h.\label{discrete_conv}
    \end{align*}
Now, by taking $U\equiv 1$ and $V=\kernel_N$ in the above we can directly compute $\xi_N=\cker^N-\cpot$ with 
\begin{eqnarray*}
    \cker^N=\kernel_N\astcirc 1 = h_xh_y \widehat{\kernel_N}_{00}=h_xh_y\sum_{(i,j)\in\Sh}\kernel_{N,ij}.
\end{eqnarray*}
Then, for example, the fully discrete system for~\eqref{eq:AC_tk1} becomes:
\begin{equation*}
       \frac{1}{\tt}{(U^{i,j}_{n+1}-U^{i,j}_{n})}+\xi_N U^{i,j}_{n+1}-\gamma*U^{i,j}_{n}+\partial  \psi(U^{i,j}_{n+1})=0,\quad (i,j)\in\Sh,\label{eq:AC_tk1_ij}
\end{equation*}
where the solution also admits a discrete version of a representation formula~\eqref{eq:prox_representation_tk1}:
\begin{align*}
  U^{i,j}_{n+1} = {\rm prox}_{\frac{1}{\lambda_N}\psi}\left(\frac{1}{\lambda_N}\left(\kernel\astcirc U^{i,j}_{n}+\frac{1}{\tt}U^{i,j}_{n}\right)\right),\quad \lambda_N=\xi_N+1/\tt, \quad (i,j)\in\Sh.
\end{align*}
Analogously, we can derive a fully discrete system having second-order accuracy and a fully discrete system for the Cahn-Hilliard model.

\section{The nonlocal phase-field network (NPF-Net)} \label{sec:NN}
We now introduce a deep learning method aimed at learning the fully discrete operators $\mathcal{H}$, which map the input $U_n$ to the output $U_{n+1}$, without requiring ground-truth data for training. Specifically, we define $\mathcal{H}_{*}$, where $*\in \{AC,CH\}$ as the operator given in Section \ref{sec:DFF}. The objective is to accurately approximate these operators, thereby enabling rapid prediction of the dynamics of the nonlocal AC and CH equations under arbitrary initial conditions, without the necessity for retraining.

\subsection{\textbf{Network architecture}} \label{sec:Network}
\begin{figure}[!ht]
    \centering
    \begin{subfigure}{1\textwidth}
        \includegraphics[width=1\textwidth]{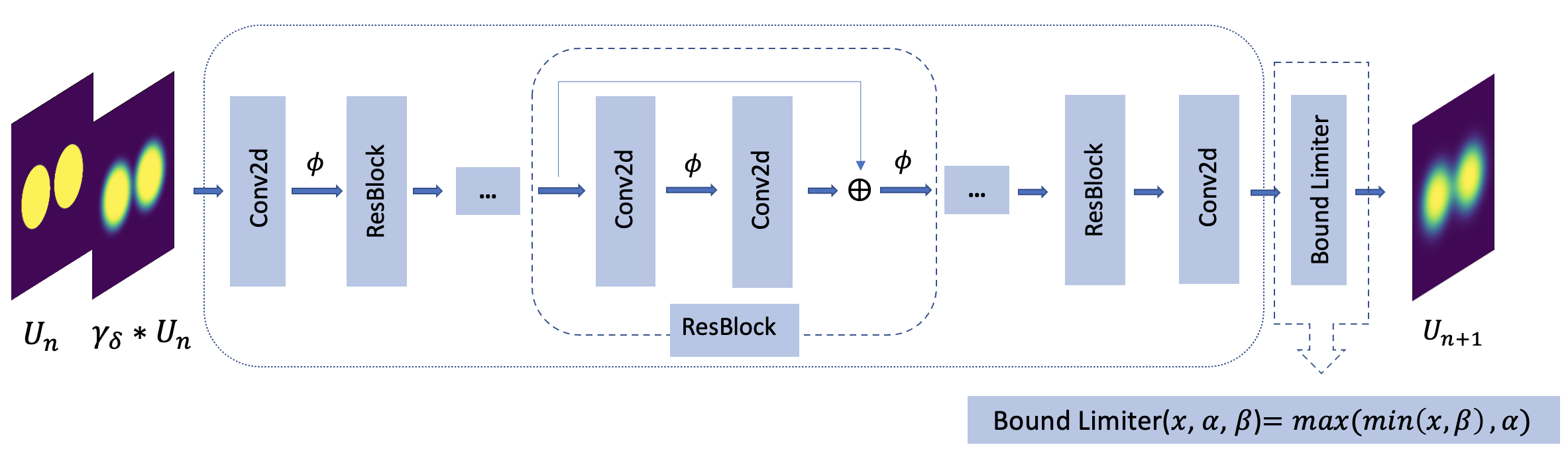}
    \end{subfigure}
    \caption{The network architecture of NPF-Net which learns the dynamics of non-mass conserving nonlocal phase field models \eqref{eq:general}. In this work, it is used to approximate the operator $\mathcal{H}_*$ with $U_{n+1}^{i,j,(l)}={\text{NPF-Net} }(U^{i,j,(l)}_{n};\Theta) \approx \mathcal{H}_*(U^{i,j,(l)}_{n})$, mapping $U_n$ to $U_{n+1}$, where $\Theta$ denotes the set of learnable parameters in NPF-Net.  }
    \label{fig:nCNN}
\end{figure}

Motivated by the network model in \cite{geng2024deep}, we propose a network architecture to model the nonlinear mappings $\mathcal{H}_{*}$ from $U_n$ to $U_{n+1}$ with the fixed time step size $\Delta t$. 
Figure \ref{fig:nCNN} illustrates the proposed ``NPF-Net'' architecture designed to learn the dynamics of the nonlocal phase-field models described by \eqref{eq:general}. This architecture includes a single convolutional layer, one or more successive Residual blocks (Resblocks) \cite{he2016deep}, and a final layer. The input to the network consists of $U_n$ and $\gamma_{\delta} *U_n$, represented as  $1 \times N_x \times N_y$ tensors. These are combined into a $2 \times N_x \times N_y$ tensor which serves as the input to predict the solution at the next time step, $U_{n+1}$, using a loss function based on the fully discrete scheme derived in Section \ref{sec:DFF}. The filter size in each convolutional layer is fixed as $K \times K$, and the number of filters $C$ is consistent across all intermediate layers. The first layer broadcasts the input tensor to the shape of $C \times N_x \times N_y$, whereas the last layer reduces it back to $1 \times N_x \times N_y$. Given the periodic boundary conditions of the target problem, the \textit{``circular"} padding \cite{8308186} model is applied across all convolutional layers. To preserve the spatial dimensions of the input feature map, the padding size is set according to the filter size used in each layer. 
ResBlocks are incorporated into the NPF-Net architecture to address the issue of vanishing gradients and improve gradient flow during training. Each ResBlock consists of two convolutional layers followed by an activation function 
$\phi$. It is important to note that violating the Maximum Bound Principle (MBP) may result in overflow errors when evaluating the loss function in the logarithmic potential and obstacle potential cases due to the presence of logarithmic terms and non-smoothness, respectively, leading to training failure. To uphold the MBP, we incorporate a bound limiter module \cite{geng2024deep} into the network architecture which ensures that the network output remains within the interval $[\alpha, \beta]$.

\subsection{\textbf{Loss function}}\label{sec:loss}

To learn the nonlinear mapping $\mathcal{H}_*$ by the discrete system, we define the loss function based on the residual of the underlying discretized system discussed in Section~\ref{sec:DFF}. Let the network input be denoted by  $\textbf{U}_n=\{U_n^{(1)},U_n^{(2)}, \ldots,U_n^{(S)}\}$ , where $U_n^{(l)}$ represents the solution at the time $t_n$ corresponding to the $l^{th}$ problem. The loss function with the regular and logarithm potential function is formulated to measure the violation \eqref{eq:AC_tk2_regular}:
\begin{equation}
{\small{\mathcal{L}_{\text{NPF-Net}} = \frac{1}{SN^2}\sum _{l=1}^{S} \sum _{i,j=0}^{N-1}\Bigg[U^{i,j,(l)}_{n+1} -U^{i,j,(l)}_{n}+\Delta t\Bigg(\xi_N U_{n/2}^{i,j,(l)}- \gamma\astcirc  U_{n/2}^{i,j,(l)} + {\left(\psi^\prime\left(U^{i,j,(l)}_{n}\right)+\psi^\prime\left(U^{i,j,(l)}_{n+1}\right)\right)/2} \Bigg)\Bigg]^2,}}
\label{eq:loss_AC_regular}
\end{equation}
where we recall $U_{n/2}^{i,j,(l)}=\left(U^{i,j,(l)}_{n+1}+ U^{i,j,(l)}_{n}\right)/{2}$. For the obstacle potential, to deal with the non-smooth function $\partial \psi_{obs}$, we define loss functions to measure the violation \eqref{eq:proj_representation_tk2_obs} with projection operator:
\begin{equation}
\mathcal{L}_{\text{NPF-Net}} = \frac{1}{SN^2}\sum _{l=1}^{S} \sum _{i,j=0}^{N-1}\Bigg[U^{i,j,(l)}_{n+1} - P_{[-1,1]}\left(\frac{1}{\frac{1}{\Delta t} + \frac{\xi_N}{2}} 
\Bigg[ \left(\frac{1}{\Delta t} - \frac{\xi_N}{2} \right)U^{i,j,(l)}_{n} + \gamma\astcirc  U_{n/2}^{i,j,(l)} \Bigg]\right)\Bigg]^2. \label{eq:loss_AC_obstacle}
\end{equation}
Thus, $U_{n+1}^{i,j,(l)}={\text{NPF-Net}}(U^{i,j,(l)}_{n};\Theta) \approx \mathcal{H}_{AC}(U^{i,j,(l)}_{n})$, where $\{U_{n+1}^{i,j,(l)}\}$ denotes the approximate solution of the $l^{th}$ problem. Note that adding the bound limiter can help us learn the nonlinear operator with a non-smooth potential function with $\{U_{n+1}^{i,j,(l)}\}_{i,j=1}^{N} \in [-1,1]$ by the bound limiter.

Similarly, for the CH equation with a smooth potential function, the loss function can be defined as 
\begin{equation}
\mathcal{L}_{\text{NPF-Net}} = \frac{1}{SN^2}\sum _{l=1}^{S} \sum _{i,j=0}^{N-1}\Bigg[ U^{i,j,(l)}_{n+1} -U^{i,j,(l)}_{n}+{\Delta t}A\Bigg(\xi_N(U^{i,j,(l)}_{n+1})- \gamma \astcirc  (U^{i,j,(l)}_{n}) + \psi'(U^{i,j,(l)}_{n+1}) \Bigg)\Bigg]^2
\label{eq:loss_CH_regular},
\end{equation} 
where $A=I-\beta\upDelta$, $\beta>0$, and for the obstacle potential function, the loss function is defined by
{\small{\begin{equation}
\mathcal{L}_{\text{NPF-Net}} = \frac{1}{SN^2}\sum _{l=1}^{S} \sum _{i,j=0}^{N-1}\Bigg[U^{i,j,(l)}_{n+1} - P_{[-1,1]} \left(\frac{1}{\frac{c}{\Delta t} + \xi_N}
\Bigg[ {\frac{1}{\Delta t}}(G+cI) \left(U^{i,j,(l)}_{n+1}-U^{i,j,(l)}_{n}\right) + \gamma \astcirc  U^{i,j,(l)}_{n} + \frac{c}{\Delta t} U^{i,j,(l)}_{n} \Bigg]\right) \Bigg]^2 ,\label{eq:loss_CH_obstacle}
\end{equation}}}
where $U_{n+1}^{i,j,(l)}={\text{NPF-Net}}(U^{i,j,(l)}_{n};\Theta) \approx \mathcal{H}_{CH}(U^{i,j,(l)}_{n})$.

It is important to note that, due to the specific selection of loss functions, NPF-Net can be trained without needing any ground truth data. This characteristic distinguishes these models, allowing them to learn effectively from the underlying dynamics of the problem rather than relying on pre-existing solution data.

\subsection{\textbf{Time-adaptive training strategy}}\label{sec: strategy}
\begin{figure}[!ht]
    \centering
\includegraphics[width=0.8\linewidth]{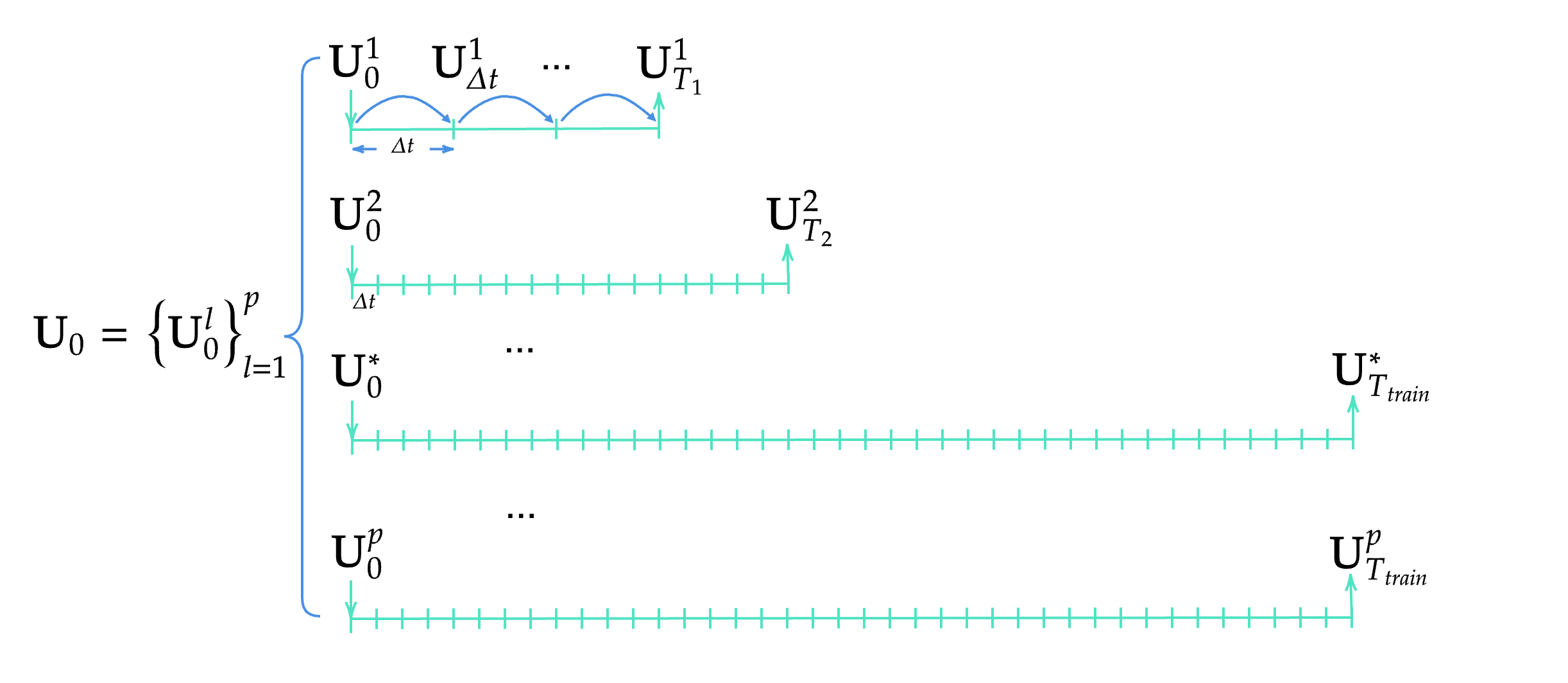}
    \caption{Illustration of the time adaptive training strategy, where training begins with a short time interval [0, $T_1$]and gradually expands as the training progresses until $T_{train}$.}
    \label{fig:adaptive training}
\end{figure}

Next, we describe the adaptive training strategy for NPF-Net. Consider a scenario with $S$ problems, each having different initial conditions, where $S$ can be expressed as $S=pq$. The initial data for these problems is organized into ${{\bf U}_0} = \{U_0^{(l)}\}_{l=1}^{S}$, with each value $\{U_0^{i,j,(l)}\}_{i,j=1}^{N}$ constrained within the interval $[-1,1]$. The training is conducted over the time interval $[0,T_{train}]$, where $T_{train} = a\Delta t$ and $a$ is a positive integer.  
Recall the training strategy in \cite{geng2024deep}; there, with fixed training initial conditions, the model was trained sequentially from initial conditions to $T_{train}$ for each subset in $\mathbf{U_0}$. In this paper, we introduced adaptive strategies in time to reduce the training time, meanwhile improving the accuracy. Figure \ref{fig:adaptive training} illustrates the detailed training with fixed $\Delta t$. 

\begin{enumerate}
\item[(i).] The initial set $\textbf{U}_0 = \left\{U_0^{(1)}, U_0^{(2)}, \ldots, U_0^{(S)}\right\}$ is randomly partitioned into $p$ subsets, denoted as $\{{\bf U}^l_0\}_{l=1}^p$, with each subset containing $q$ distinct initial conditions such that $S=pq$.

\item[(ii).] The neural network is initially trained using the first subset $\textbf{U}_0^{1}$ to learn the mapping from $\textbf{U}^1_0$ to $\textbf{U}^1_{T_1}$ sequentially with a specified learning rate, where $T_1 \leq T_{train}$. 

\item[(iii).] Subsequently, the neural network is provided with the second subset $\textbf{U}_0^2$ and the learning rate decay by some constant factor, and the training process continues from $\textbf{U}^2_0$ to $\textbf{U}^2_{T_2}$, ensuring that $T_1 \leq T_2 \leq T_{train}$. This process is iterated for each subset until the designated training time $T_{train}$ is achieved. If $T_1 =T_2=\cdots =  T_{train}$, this corresponds to the training strategy outlined in \cite{geng2024deep}.
\item[(iv).] Upon reaching $T_{train}$ with initial condition $\textbf{U}_0^*$, the model undergoes further training to learn the process from $\textbf{U}_0$ to $\textbf{U}_{T_{train}}$. Note, once the training time reaches $T_{train}$, it remains fixed, and training continues until the loss reaches an acceptable threshold or the maximum number of epochs is completed.
\end{enumerate}

\noindent This approach effectively balances training time and accuracy, preventing the adverse effects of prolonged sequential training.

\section{Numerical results} \label{sec: Results}

In this section, we conduct some hyperparameter optimizations on different network architectures and then test the trained models on benchmark problems for the nonlocal AC and CH equations. 
Throughout all the experiments, we fix the spatial domain $\Omega=[-1,1]^d$, $d=2,3$ and use $\Delta t =0.1$ for the nonlocal AC equation and $\Delta t = 0.01$ for the nonlocal CH equation. For the mesh grid, we fix $N_x=N_y=256$ for two-dimensional case ($d=2$) and $N_x=N_y=N_z=64$ for three-dimensional case ($d=3$). We set $\cpot = 1$ for all three types of potential functions and for the logarithm potential function we set $\thetalog= 0.5$. 
 The bounds for the AC equation are as follows: for the regular and obstacle potential functions, the bounding interval is $[-1,1]$.
In the case of the logarithmic potential, the bounds are $[-\rho_1, \rho_1]$ where $\rho_1 = 0.95750402$. For the nonlocal CH equation, there are no bounds for the regular potential function, whereas for the obstacle potential function, the bound is $[-1,1]$. We employ a smooth kernel function defined as the Gauss-type functions having the form
\begin{equation*}
    \gamma_{\delta}(\mathbf{x}) = \frac{4\epsilon^2}{\pi^{(d/2)}\delta^{(d+2)}} e^{{-|\mathbf{x}|^2}/{\delta^2}}, \quad\mathbf{x} \in \mathbb{R}^d,\, \delta >0,
\end{equation*}
which is appropriately scaled to match $Bu\approx-\epsilon^2\upDelta u$ for $\delta\to 0$, and we fix $\epsilon = 0.05$.
For the experiments, we select $\delta =0.05,0.075,0.095,0.1$  with corresponding values of $\xi=3,0.778,0.108,0$ respectively.

To train our networks, we follow the same parameter settings as outlined in \cite{geng2024deep} and the random initial conditions  $u_0(x,y)=0.95\, \textbf{rand}(x,y)$, where $\textbf{rand}(x,y)$ denotes the pseudo-random number generator producing a scalar value between $-1$ and $1$. Unless stated otherwise, we adopt the Adam optimizer \cite{DBLP:journals/corr/KingmaB14} with an initial learning rate of $0.001$ which decays by a factor of 0.6 after the training of each subset in $\mathbf{U_0}$.  Additionally, we set the maximum number of epochs to 2 and the loss tolerance to 
1e-10. Our models are implemented in PyTorch \cite{paszke2017automatic}, and all experiments are conducted on a server equipped with a V100 GPU card with 32GB of memory. The training time for each model in the experiments presented in this section range over 9 hours for two-dimensional problems and 50 hours for three-dimensional problems.

\begin{figure}[h!]
    \centering
    \begin{subfigure}{0.3\textwidth}
        \includegraphics[width=0.95\textwidth]{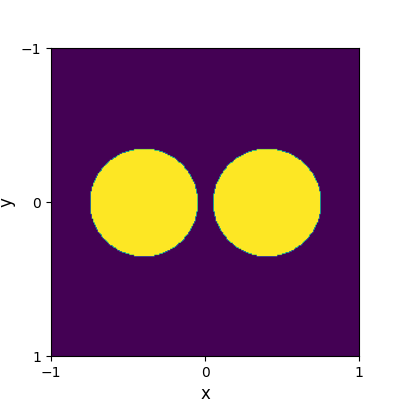}
        \caption*{\small{$u_0^{1}$}}
    \end{subfigure}
    \begin{subfigure}{0.3\textwidth}
        \includegraphics[width=0.95\textwidth]{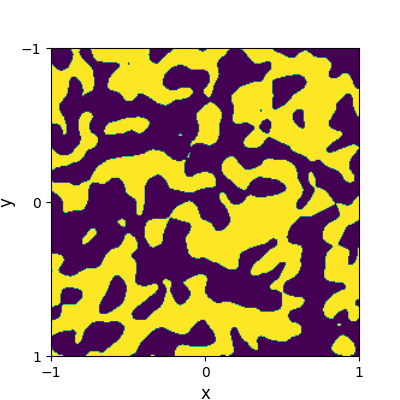}
        \caption*{\small{$u_0^{2}$}}
    \end{subfigure}
    \begin{subfigure}{0.3\textwidth}
        \includegraphics[width=0.95\textwidth]{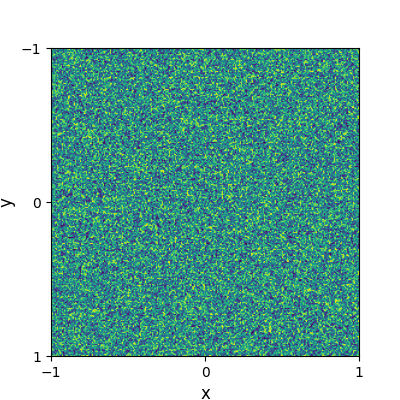}
        \caption*{\small{$u_0^{3}$}}
    \end{subfigure}
    \caption{Three different types of initial conditions. $u_0^1$ is for the bubbles merging case, $u_0^2$ is for sharp-colored noise, and $u_0^3$ is for white noise initial condition. }
    \label{fig:initials}
\end{figure}

To evaluate the accuracy of our trained models, we calculate the prediction errors using the relative $L_2$-norm
\begin{equation*}
    Error = \frac{\lVert \mathbf{U}_{\mathcal{NN}}-\mathbf{U}_{ref}\rVert_{2}}{\lVert\mathbf{U}_{ref}\rVert_{2}},
\end{equation*}
where $\mathbf{U}_{\mathcal{NN}}$ is the predicted solution generated by the trained neural network models and $\mathbf{U}_{ref}$ is the reference solution produced by the corresponding fully-discrete schemes. The reference solution is computed with the same spatial mesh and the sufficiently small time step size $\Delta t=0.001$ for AC and $\Delta t=1e-5$ for CH. The resulting nonlinear systems are solved using a Picard iteration with FFT-based implementation at each time step as outlined in Section~\ref{sec:DFF}. We consider three different types of initial conditions to test our network for solving two-dimensional nonlocal problems. The performance of these different initial conditions is shown in Figure \ref{fig:initials}. 
For the bubbles-merging initial condition $u_0^{1}$ we define
\begin{eqnarray*}
u_0^{1}(x,y)=
\begin{cases}
    1, ~~ \,\,\,\,\text{ if }  (x-0.4)^2+y^2\leq 0.35^2~~ \text{and}~~ (x+0.4)^2+y^2\leq 0.35^2,\\
    -1,~~\text{ else}
\end{cases}.
\end{eqnarray*}
For the sharp-colored noise initial condition $u_0^{2}$, we first generate the smooth Gaussian noise $G(x,y)$ by using the Python tools GSTools \cite{gmd-15-3161-2022} and then we generate the sharp-colored noise initial condition $u_0^{2}$,  where the values are restricted to either $1$ or $-1$, as follows:
\begin{eqnarray*}
u_0^{2}(x,y)=
\begin{cases}
    1, ~~\,\,\,\, \text{ if } ~~G(x,y)\geq0,  \\
    -1,~~\text{ if}  ~~G(x,y)<0.
\end{cases}
\end{eqnarray*}
Finally, the white noise initial condition $u_0^{3}$ is selected from $u_0^{3}(x,y) = 0.95\, \textbf{rand}(x,y)$. To assess the generalizing ability of our network, we calculate the average $L_2$ error across 50 different initial conditions for both white and sharp-colored noise, as both types of initial conditions can be generated randomly.
\vspace{-0.3cm} 
\subsection{\textbf{Hyperparameter optimization} }\label{sec:ab}
In this section, we perform hyperparameter optimizations to evaluate how improved training strategies and network architecture affect a model's prediction performance. For this analysis, we focus solely on the two-dimensional nonlocal AC equations.
\vskip5pt
\noindent{\textbf{\em Effect of the training strategy.}}
\begin{figure}[h!]
    \centering
    \begin{subfigure}{0.32\textwidth}
        \includegraphics[width=1.1\textwidth]{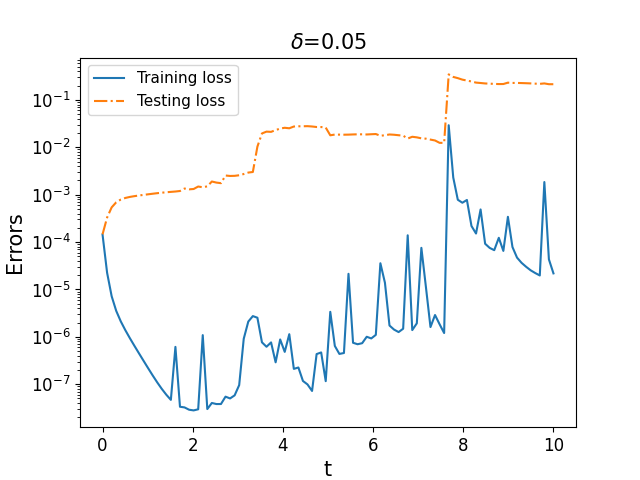}
    \end{subfigure}
    \begin{subfigure}{0.32\textwidth}
        \includegraphics[width=1.1\textwidth]{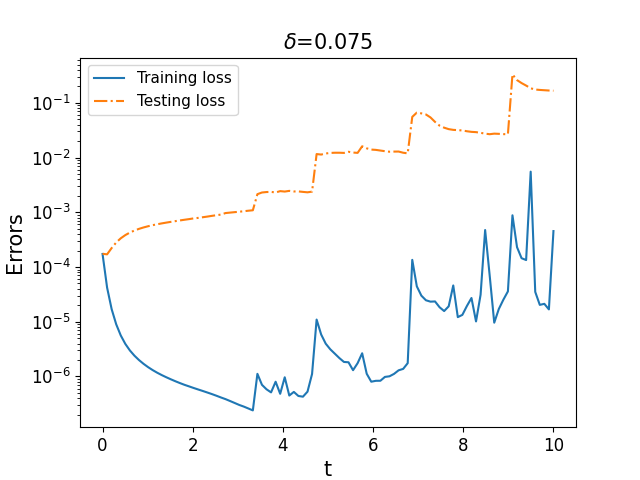}
    \end{subfigure}
   \begin{subfigure}{0.32\textwidth}
        \includegraphics[width=1.1\textwidth]{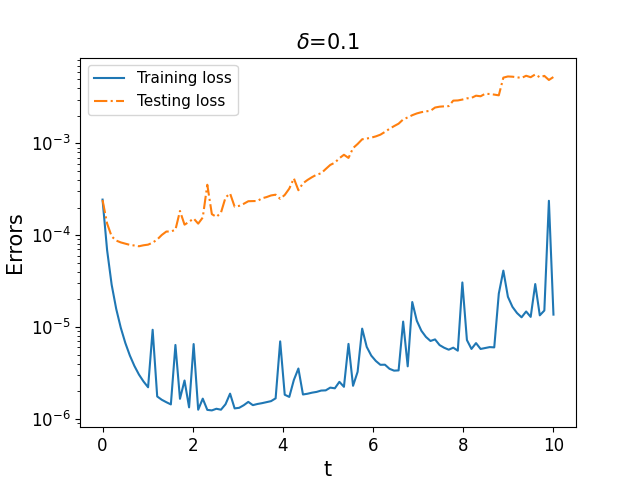}
    \end{subfigure}
    \caption{Training loss and testing loss for training the steps from $U_0$ to $U_{T_{train}}$ sequentially with different $\delta$. The testing set is 10 random initial conditions outside the training set. For all choices of $\delta$, both training loss and test loss increase sharply, especially after $t=2$ (20 steps).}
    \label{fig:Loss}
\end{figure}
Our first objective is to investigate how the improved training strategy affects the performance of the proposed methods for solving the nonlocal AC problem. First, we follow the sequential training strategy with $T_1=T_{train}=10$ on a training subset $U^1_0$ which contains $q=5$ random initial conditions. Then, we test our model on 10 random initial conditions outside of the training set. We observe that as the model progressed to later time steps for the nonlocal AC equation, the training loss and testing loss increase significantly, especially after $t=2$, as shown in Figure \ref{fig:Loss}. Therefore, we propose a time-adapted training strategy aimed at reducing training time while maintaining accuracy. Instead of training the model directly to $T_{train}$, we select $T_1 =2$, $T_2=4$ until to $T_{train}=10$ by adding $\Delta T=2$ for each training subset. In the next section, we compare the accuracy of the sequential training method and time-adaptive training strategies on different network architectures.

\vskip5pt
\noindent{\textbf{\em Effect of the network architecture.}}
\begin{figure}[h!]
\centering
\begin{subfigure}{1\textwidth}
\centering
\includegraphics[width=0.34\textwidth]{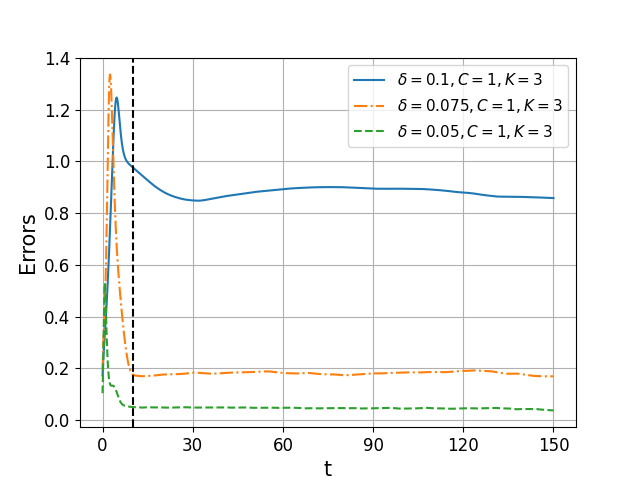}
\hspace{-0.4cm}\includegraphics[width=0.34\textwidth]{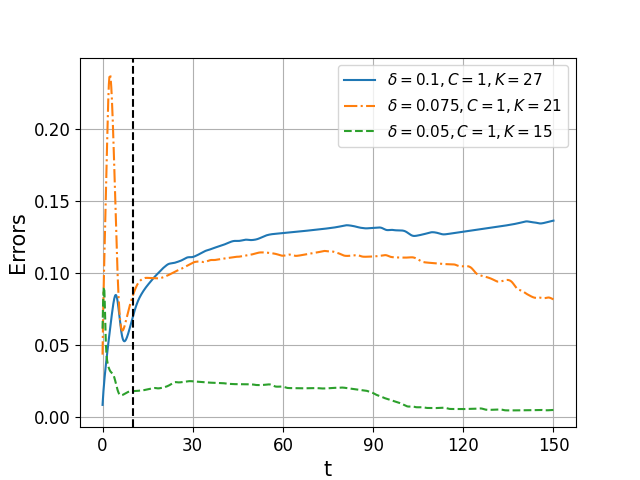}
\hspace{-0.4cm}\includegraphics[width=0.34\textwidth]{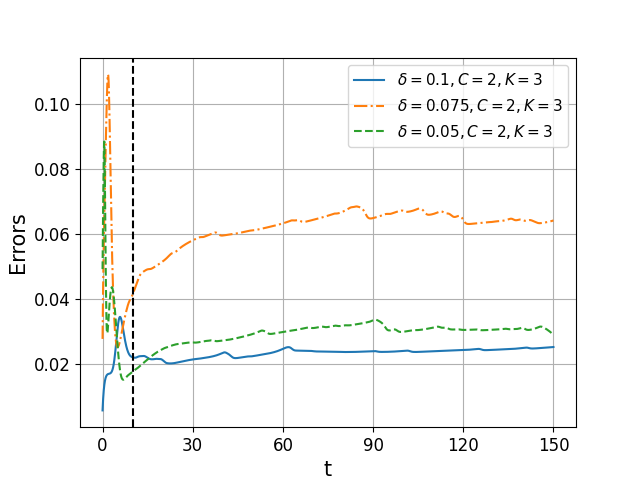}
\caption*{\scriptsize{Sequential training strategy}}
\end{subfigure}
\begin{subfigure}{1\textwidth}
\centering
\includegraphics[width=0.34\textwidth]{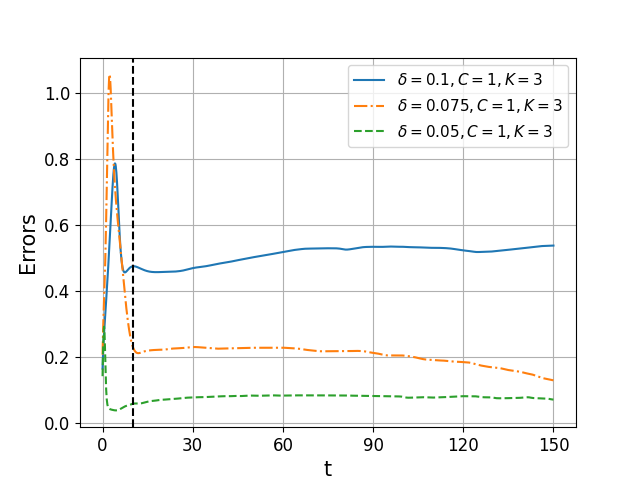}
\hspace{-0.4cm}\includegraphics[width=0.34\textwidth]{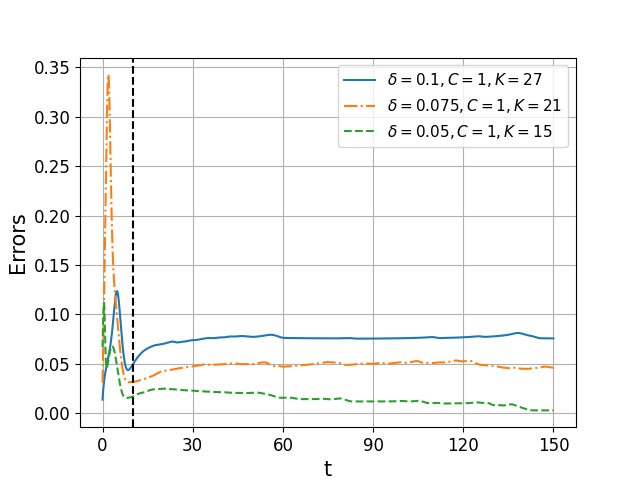}
\hspace{-0.4cm}\includegraphics[width=0.34\textwidth]{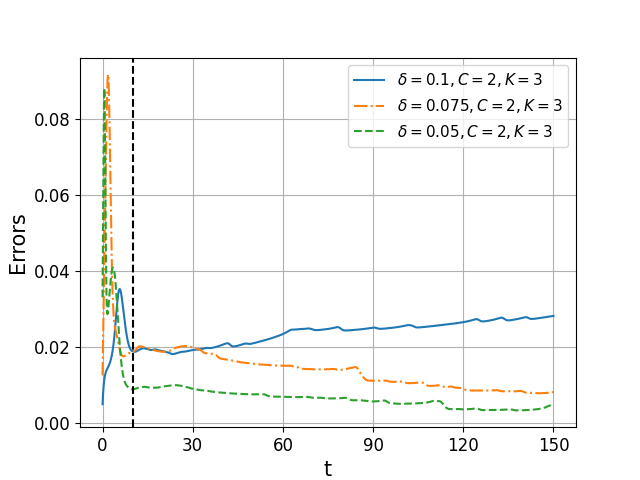}
\caption*{\scriptsize{Time-adaptive training strategy}}
\end{subfigure}
\caption{Evolution of the model prediction errors for the regular potential case under different training strategies. A network using $U_n$ as input with a fixed kernel size performs poorly, but the accuracy improves as the filter size $K$ of the input layer increases. Ultimately, the network using both $U_n$ and $\gamma * U_n$ as inputs achieves the best performance, with stable simulation errors across different choices of $\delta$. In addition, for all network settings, the simulation errors under the time-adaptive training strategy are significantly lower than those using the sequential training method. The vertical black dashed lines indicate that the end of the training time is $T_{train}=10$.}
\label{fig:Errors_DW_AB1_train}
\end{figure}
To illustrate the advantage of employing $\gamma * U_{n}$ in NPF-Net, we compare the performance of ACNN to NPF-Net. Figure \ref{fig:Errors_DW_AB1_train} presents the average $L_2$ error over 50 white noise initial conditions for the sequential training strategy (top) and time-adaptive training process (bottom), respectively. 
For the ACNN model with the input containing one channel, we first used 3 resblocks 16 mid-channels and a fixed filter size 3 for all layers, corresponding to $C=1$ and $K=3$ in Figure \ref{fig:Errors_DW_AB1_train} (left).   
From the results, we observe that the errors are significantly large, especially for a large choice of $\delta$ with a long interaction domain. To solve that problem, we then consider increasing the input filter size corresponding to the different options of $\delta$. For instance, with a mesh size $256\times 256$ over the domain $[-1,1]^2$ ($h = {2}/{256}$) and $\delta=0.1$, we select the filter size to be 27 and a padding number of 13, corresponding to $\frac{\delta}{h}=12.8$. Similarly, the input filter size can be determined based on the choice of $\delta$ in Table \ref{tab:kernel_size}. 
\begin{table}[ht!]
\centering
\small
\begin{tabular}{|c|c|c|c|}
\hline
kernel, padding & $\delta =0.05$ & $\delta =0.075$ &  $\delta =0.1$  \\ \hline
$N=$256   & 15,7             & 21,10          & 27,13           \\ \hline
\end{tabular}
\caption{The convolutional kernel size and padding correspond to different choices of~$\delta$.}
\label{tab:kernel_size}
\end{table}
The results, shown in Figure \ref{fig:Errors_DW_AB1_train} (middle), indicate a significant error reduction with increased receptive fields by increasing the filter size. However, for large $\delta$, the errors, although reduced, remain notable. Additionally, small mesh sizes $h$ necessitate adjustments in filter sizes to enhance receptive fields, which consequently increases the model complexity and the risk of overfitting. To address these challenges, we propose NPF-Net for the nonlocal problem which enhances performance without increasing the receptive fields by adding the convolutional kernel $\gamma * U_n$ to the input channel with $C=2$ and $K=3$. Figure \ref{fig:Errors_DW_AB1_train} (right) demonstrates that accuracy is significantly improved for large $\delta$ values and that the accuracy remained consistent across different $\delta$ values.
 In addition, the time-adaptive training strategy is more accurate and stable than the original training method for different network architectures. Furthermore, while the sequential training took 13 hours, the adaptive training strategy reduced the training time to 9 hours (a reduction of approximately 30\%). This time reduction is even more pronounced for larger-scale problems. Therefore, we use NPF-Net with the adaptive training strategy in all subsequent experiments. 
\vskip5pt
\noindent{\textbf{\em Effect of different training sets.}}
\begin{figure}[h!]
\centering
\begin{subfigure}{1\textwidth}
\centering
\includegraphics[width=0.34\textwidth]{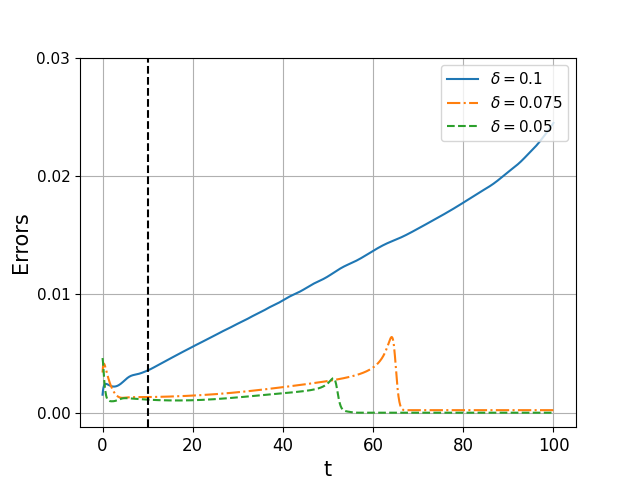}
\hspace{-0.4cm}\includegraphics[width=0.34\textwidth]{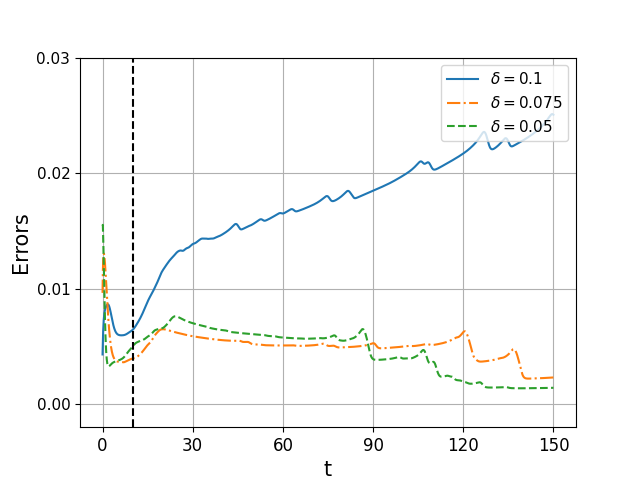}
\hspace{-0.4cm}\includegraphics[width=0.34\textwidth]{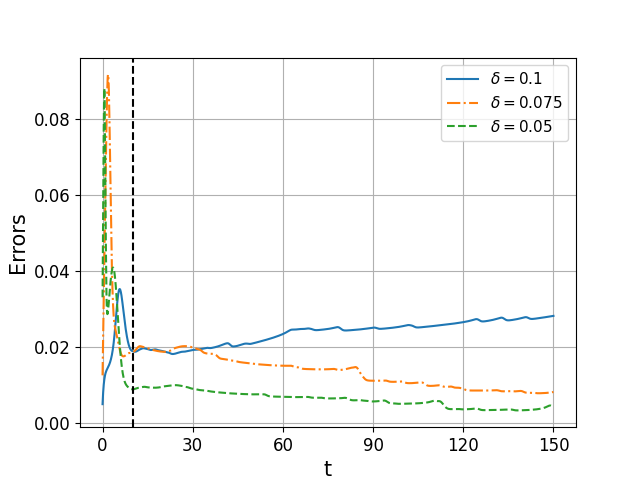}
\caption*{\scriptsize{Regular: $\mathbf{U}_0$ contains white noise $u^3_0$ only}}
\end{subfigure}
\begin{subfigure}{1\textwidth}
\centering
\includegraphics[width=0.34\textwidth]{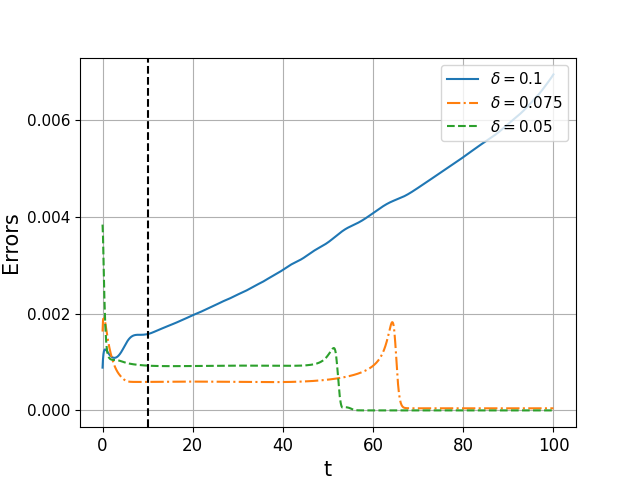}
\hspace{-0.4cm}\includegraphics[width=0.34\textwidth]{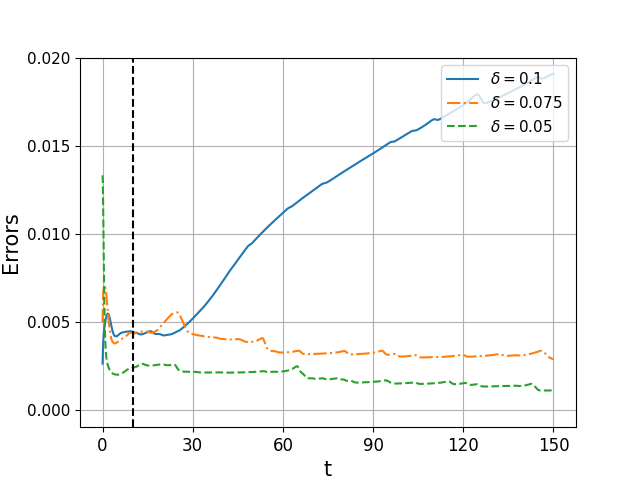}
\hspace{-0.4cm}\includegraphics[width=0.34\textwidth]{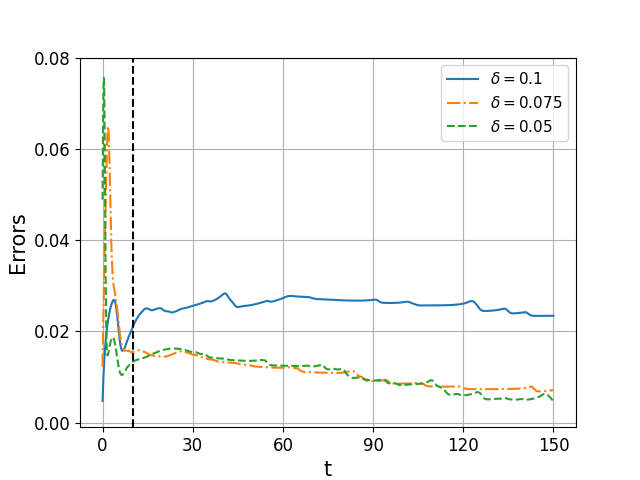}
\caption*{\scriptsize{Regular: $\mathbf{U}_0$ contains sharp-Colored Noise $u_0^2$ and white Noise $u_0^3$}}
\end{subfigure}
\caption{Evolution of model prediction errors for the nonlocal AC equation with the regular potential function, tested with different initial conditions: $u_0^1$ (left), $u_0^2$ (middle), and $u_0^3$ (right). Top row: the NPH-Net model is trained using an initial training set $\mathbf{U}_0$ containing only $u_0^3$. Bottom row: the network model is trained with the initial training set $\mathbf{U}_0$ incorporated with sharp-colored noise $u_0^2$. Comparing the two training choices, it is evident that incorporating sharp-colored noise into the initial training set significantly improves the model’s accuracy across different initial conditions. The vertical black dashed line indicates that the end of the training time is $T_{train}=10$.}
\label{fig:Errors_regular}
\end{figure}
\begin{figure}[h!]
\centering
\begin{subfigure}{1\textwidth}
\centering
\includegraphics[width=0.34\textwidth]{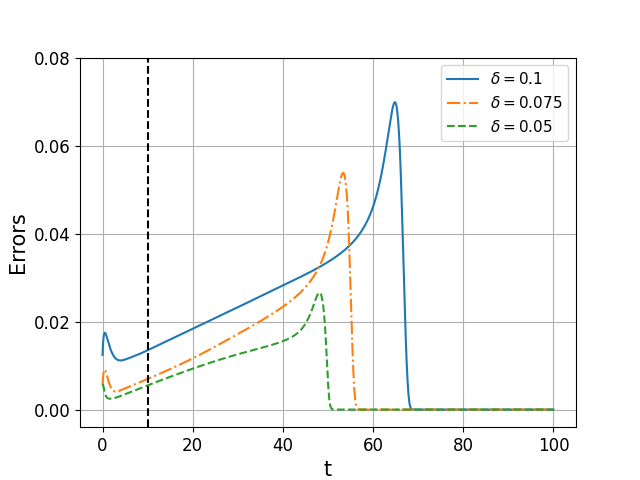}
\hspace{-0.4cm}\includegraphics[width=0.34\textwidth]{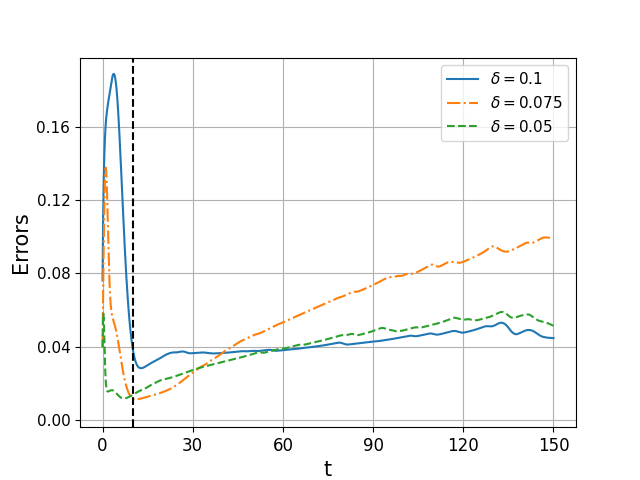}
\hspace{-0.4cm}\includegraphics[width=0.34\textwidth]{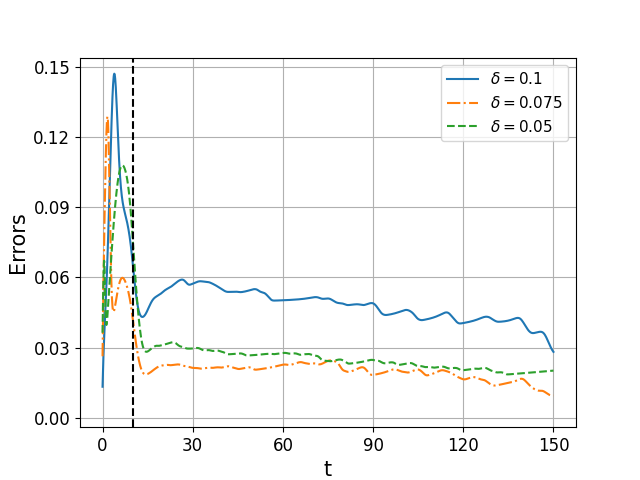}
\caption*{\scriptsize{Logarithmic: $\mathbf{U}_0$ contains white noise $u^3_0$ only}}
\end{subfigure}
\begin{subfigure}{1\textwidth}
\centering
\includegraphics[width=0.34\textwidth]{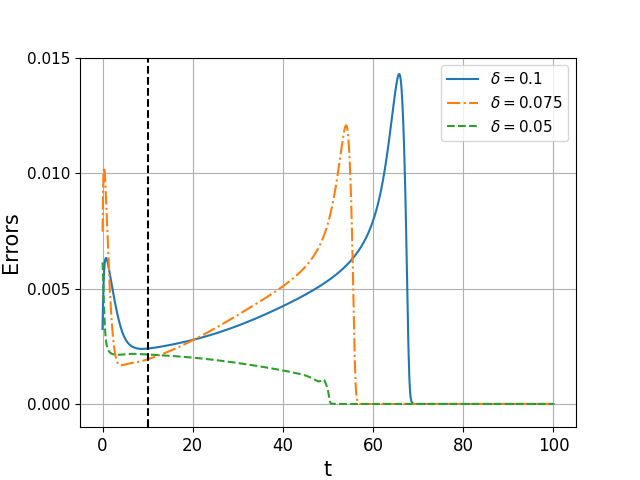}
\hspace{-0.4cm}\includegraphics[width=0.34\textwidth]{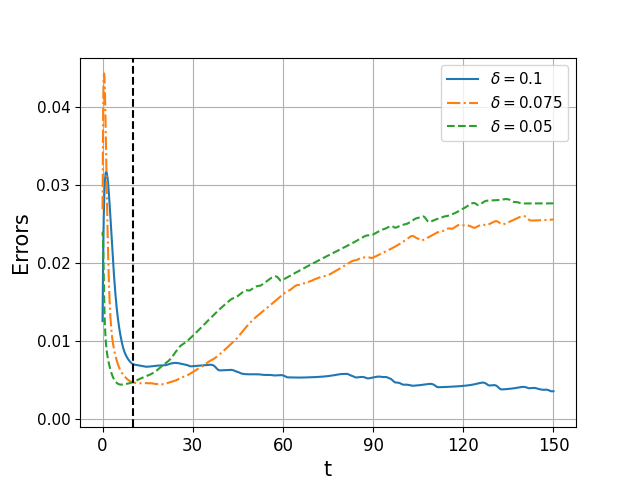}
\hspace{-0.4cm}\includegraphics[width=0.34\textwidth]{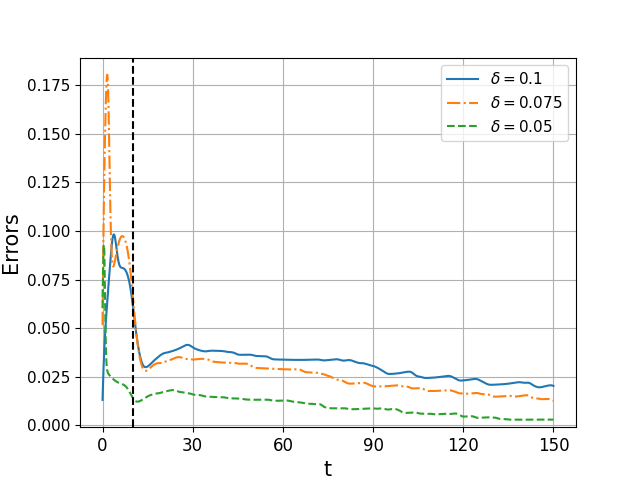}
\caption*{\scriptsize{Logarithmic: $\mathbf{U}_0$ contains sharp-Colored Noise $u_0^2$ and white Noise $u_0^3$}}
\end{subfigure}
\caption{Evolution of model prediction errors for the nonlocal AC equation with the logarithmic potential function, tested with different initial conditions: $u_0^1$(left), $u_0^2$ (middle), and $u_0^3$ (right). The vertical black dashed line indicates that the end of the training time is $T_{train}=10$.}
\label{fig:Errors_loga}
\end{figure}
\begin{figure}[h!]
\centering
\begin{subfigure}{1\textwidth}
\centering
\includegraphics[width=0.34\textwidth]{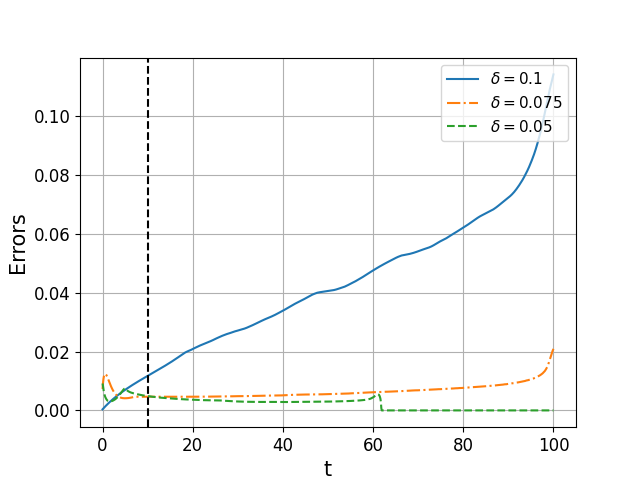}
\hspace{-0.4cm}\includegraphics[width=0.34\textwidth]{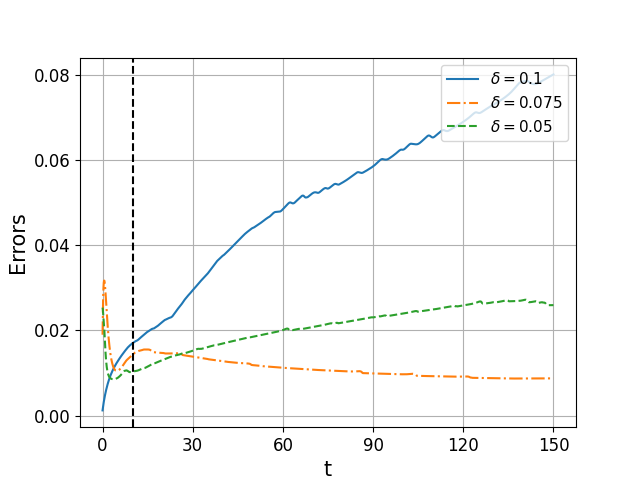}
\hspace{-0.4cm}\includegraphics[width=0.34\textwidth]{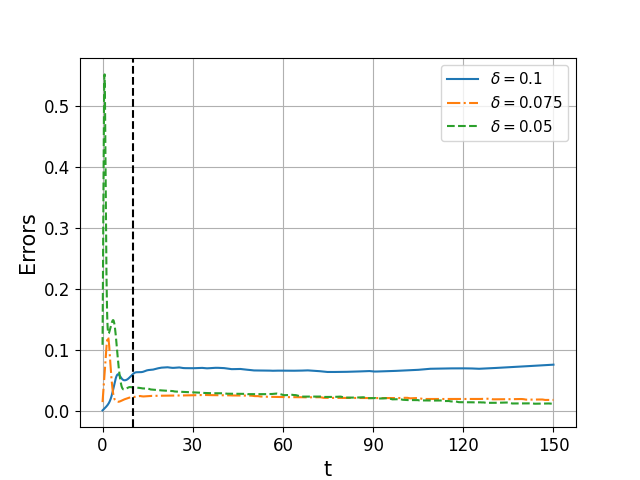}
\caption*{\scriptsize{Obstacle: $\mathbf{U}_0$ contains white noise $u^3_0$ only}}
\end{subfigure}
\begin{subfigure}{1\textwidth}
\centering
\includegraphics[width=0.34\textwidth]{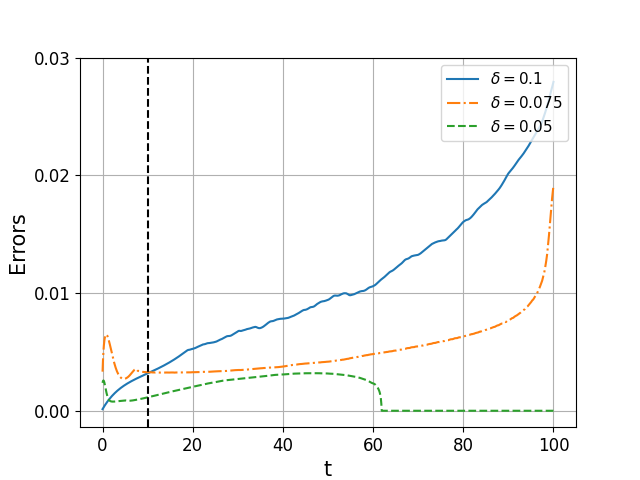}
\hspace{-0.4cm}\includegraphics[width=0.34\textwidth]{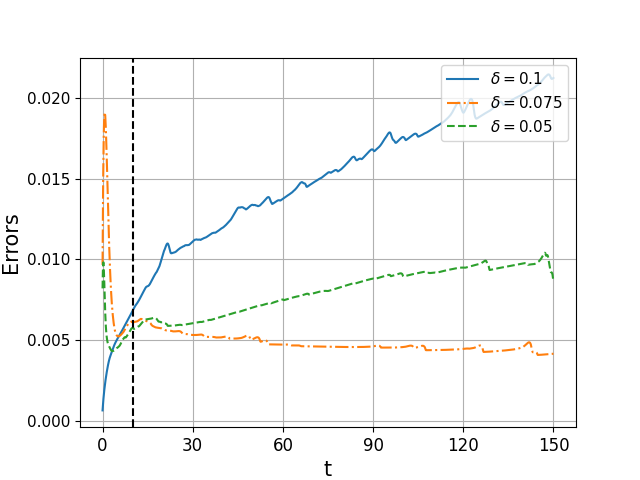}
\hspace{-0.4cm}\includegraphics[width=0.34\textwidth]{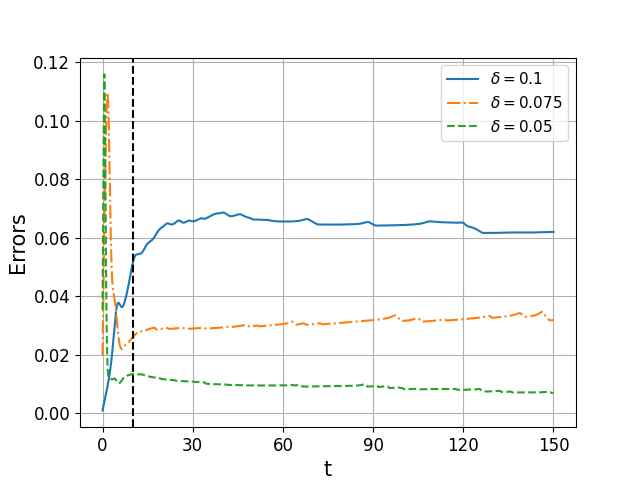}
\caption*{\scriptsize{Obstacle: $\mathbf{U}_0$ contains sharp-Colored Noise $u_0^2$ and white Noise $u_0^3$}}
\end{subfigure}
\caption{Evolution of model prediction errors for the nonlocal AC equation with the obstacle potential function, tested with different initial conditions:  $u_0^1$(left), $u_0^2$ (middle), and $u_0^3$ (right). The vertical black dashed line indicates that the end of the training time is $T_{train}=10$.}
\label{fig:Errors_Obs}
\end{figure}

To evaluate the network's performance across different types of initial conditions, we first tested our NPF-Net model training with $\mathbf{U}_0$, which contains 20 random initial conditions ($u_0^3$) only. Following this, we simulated solutions with various initial conditions. The top rows in Figures \ref{fig:Errors_regular}, \ref{fig:Errors_loga}, and \ref{fig:Errors_Obs} shows the error for the nonlocal AC equation with regular, logarithmic, and obstacle potential functions, respectively. We observe that the model exhibited significant errors, especially when handling sharp profiles with logarithmic and obstacle potential functions, e.g., bubbles merging $u_0^1$ and sharp-colored noise $u_0^2$ initial conditions. To address this issue, we incorporated sharp-colored noise $u_0^2$ into $\mathbf{U}_0$ alongside the random initial conditions. For training, we selected a $\mathbf{U}_0$=$\{{\bf U}^l_0\}_{l=1}^p$ containing 20 white noise initial conditions ($u_0^3$) and 20 sharp-colored noise conditions ($u_0^2$). For each training subset, we randomly selected $\textbf{U}_0^l$ containing 4 white noise initial conditions and 4 sharp-colored noise initial conditions. As the bottom row in these figures shows, this approach effectively halved the error across various initial conditions. Thus, integrating sharp Gaussian noise into $\mathbf{U}_0$ significantly enhances the model's generalizing capabilities and accuracy, particularly for handling sharp interfaces. Consequently, we include both white and sharp-colored initial conditions in $\mathbf{U}_0$ for all subsequent experiments.

\subsection{\textbf{Two-dimensional bubbles merging}}
Next, we demonstrate, through bubble-merging examples, the excellent performance of our proposed methods, both in terms of accuracy and efficiency. These examples illustrate how well the NPF-Net model learns the dynamics of the AC and CH equations. To evaluate the accuracy of our model, we compare the results using NPF-Net with the numerical scheme discussed in Section \ref{sec:DFF}  using the same mesh size $N_x=N_y$ and time step size $\Delta t$. The benchmark solutions are generated by solving the numerical scheme with a finer time step and double the mesh grid resolution. In this aspect of the study, we focus solely on the bubble-merging case because doubling the mesh grid for cases with random initial conditions would lead to different solution behaviors. Additionally, we observe that solving the nonlocal phase-field model with a regular potential behaves similarly to that using a logarithmic potential. Therefore, for simplicity, in the subsequent sections we focus exclusively on systems with regular and obstacle potential functions.

\vskip5pt
\subsubsection{\textbf{Nonlocal Allen-Cahn equation}}
\begin{figure}[h!]
    \centering
    \begin{subfigure}{1\textwidth}
        \includegraphics[width=0.34\textwidth]{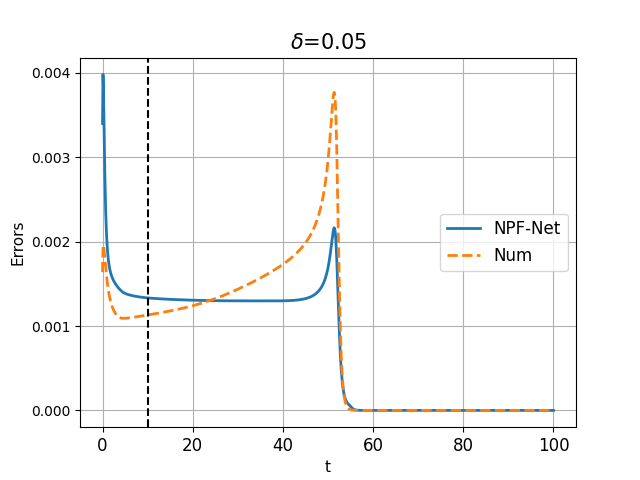}
        \hspace{-0.4cm}\includegraphics[width=0.34\textwidth]{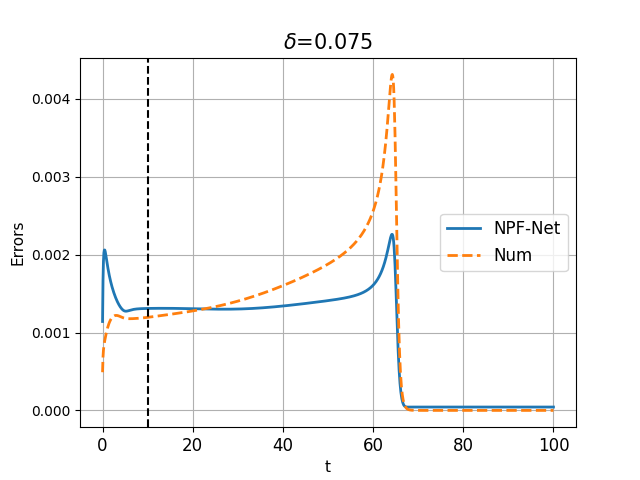}
       \hspace{-0.4cm} \includegraphics[width=0.34\textwidth]{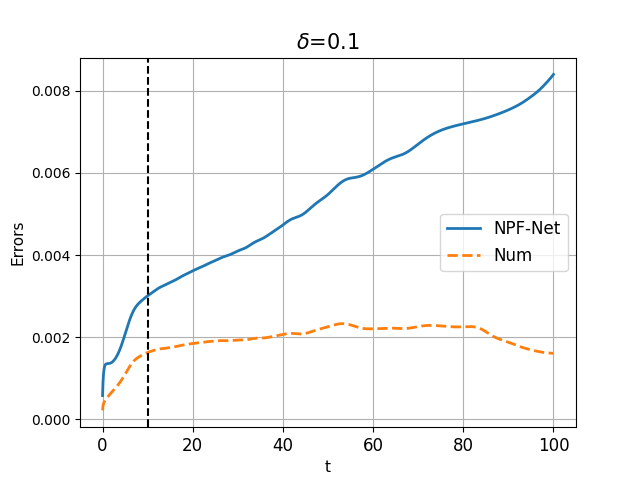}
        \caption*{\scriptsize{Regular}}
    \end{subfigure}
    \begin{subfigure}{1\textwidth}
        \includegraphics[width=0.34\textwidth]{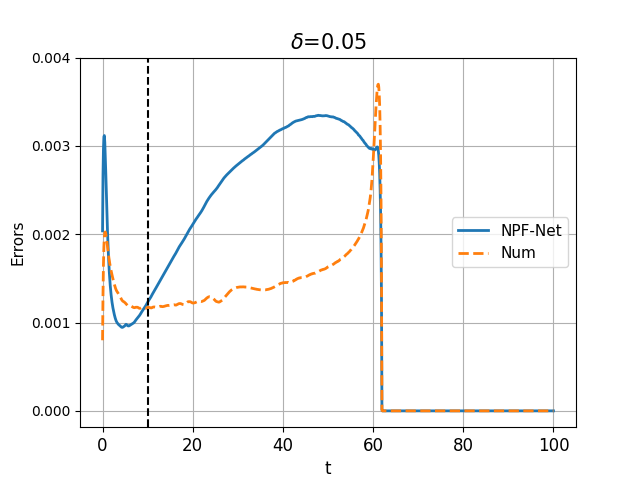}
        \hspace{-0.4cm}\includegraphics[width=0.34\textwidth]{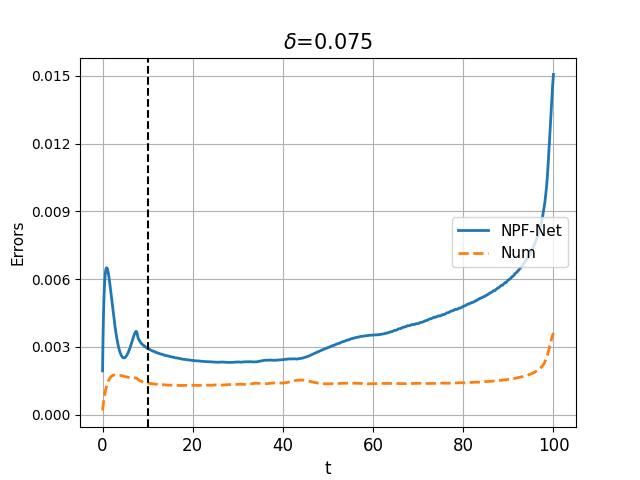}
        \hspace{-0.4cm}\includegraphics[width=0.34\textwidth]{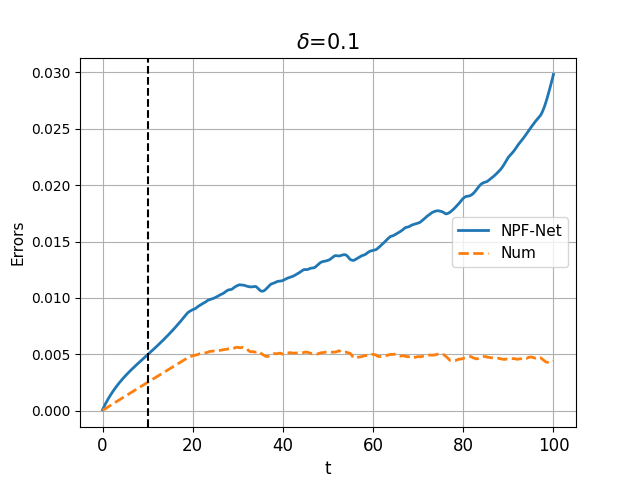}
        \caption*{\scriptsize{Obstacle}}
    \end{subfigure}
    \caption{
    Evolution of model prediction errors for the nonlocal AC equation when using the NPF-Net model and the Crank-Nicolson scheme with $N_x=N_y=256$, $\Delta t=0.1$, and for different choices of $\delta$. The vertical black dashed line indicates that the end of the training time is $T_{train}=10$.}
    \label{fig:AC_2D_mesh_test}
\end{figure}
\begin{figure}[h!]
\centering
\begin{subfigure}{1\textwidth}
\centering
\includegraphics[width=0.49\textwidth]{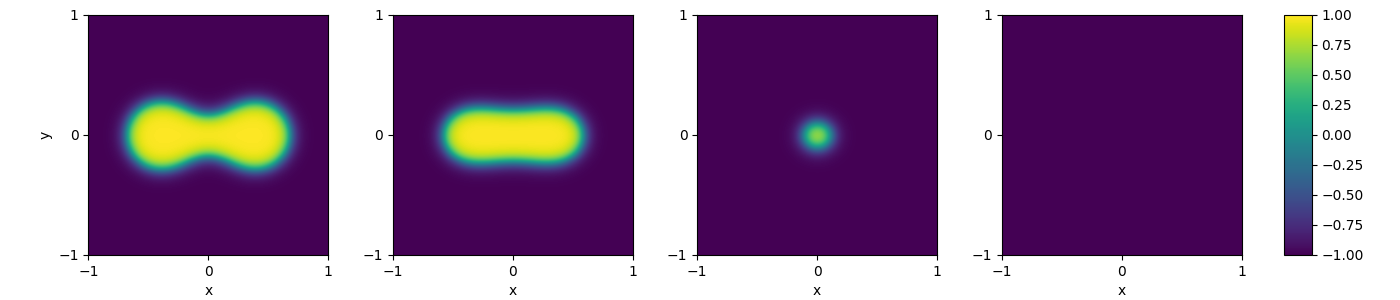}
\includegraphics[width=0.49\textwidth]{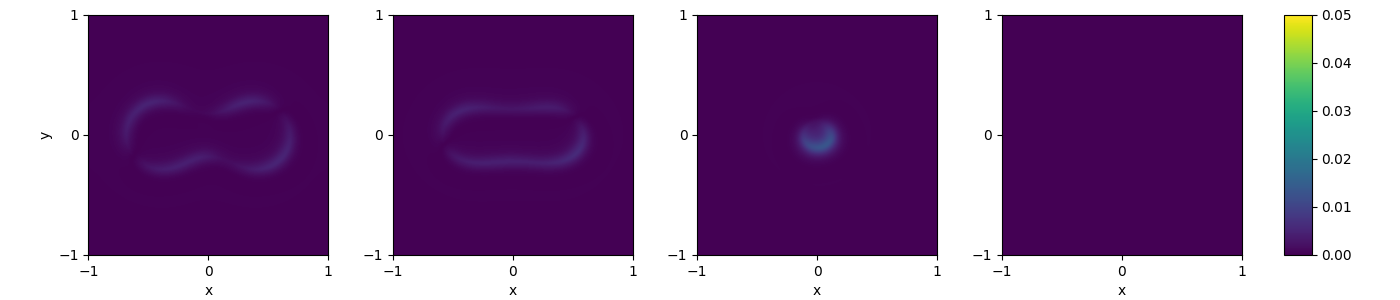}
\includegraphics[width=0.49\textwidth]{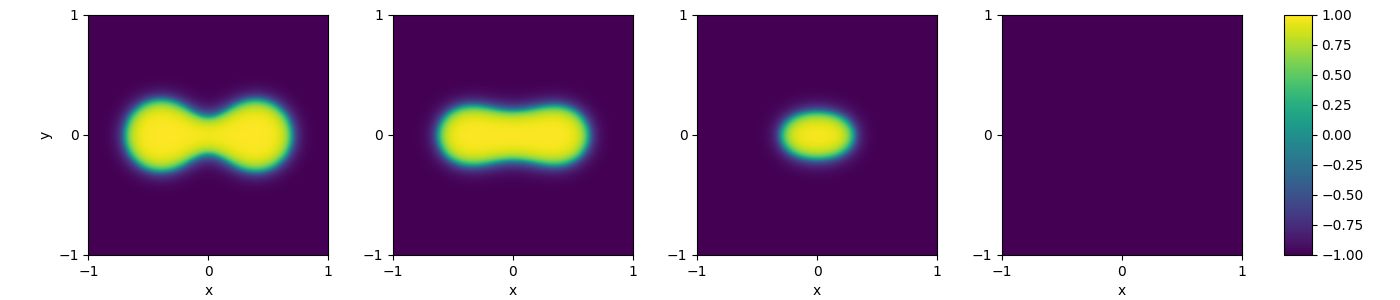}
\includegraphics[width=0.49\textwidth]{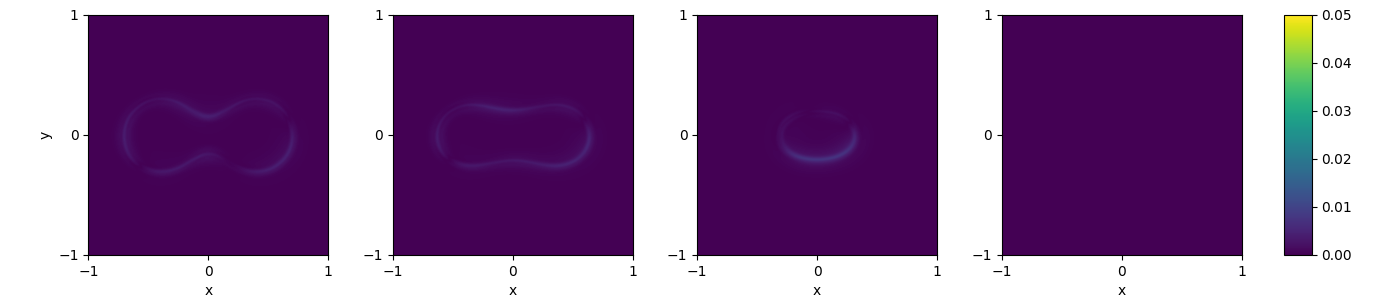}
\includegraphics[width=0.49\textwidth]{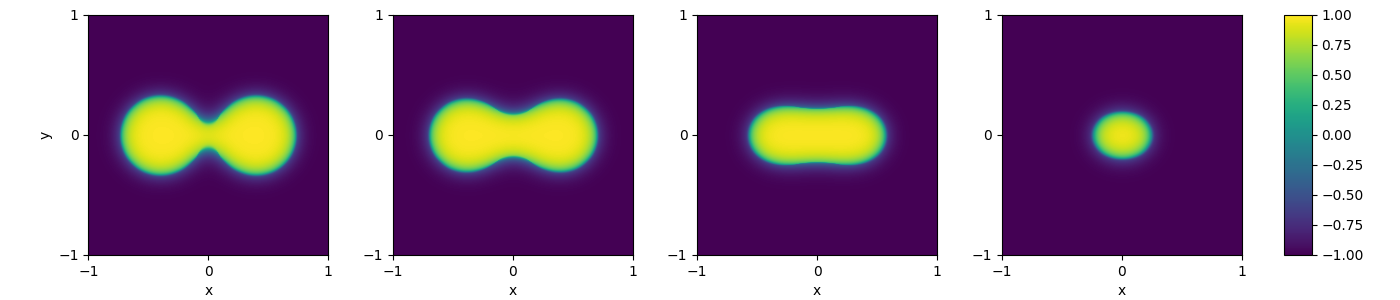}
\includegraphics[width=0.49\textwidth]{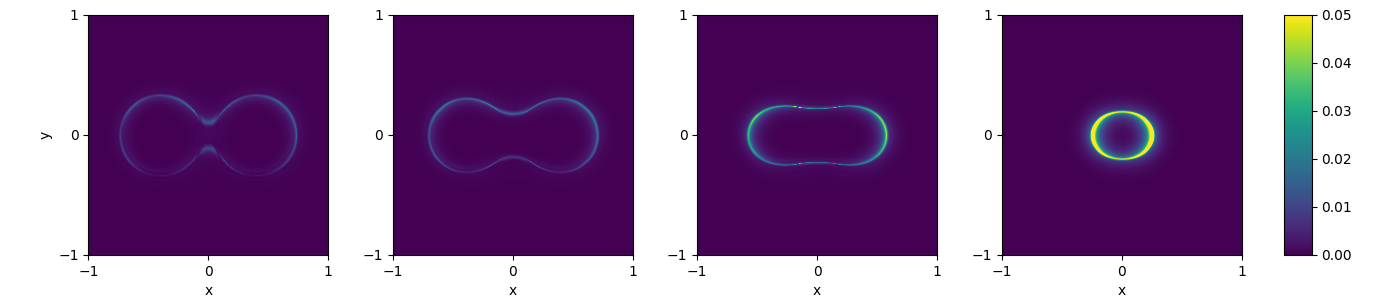}
\caption*{\scriptsize{Regular}}
\end{subfigure}

\begin{subfigure}{1\textwidth}
\centering
\includegraphics[width=0.49\textwidth]{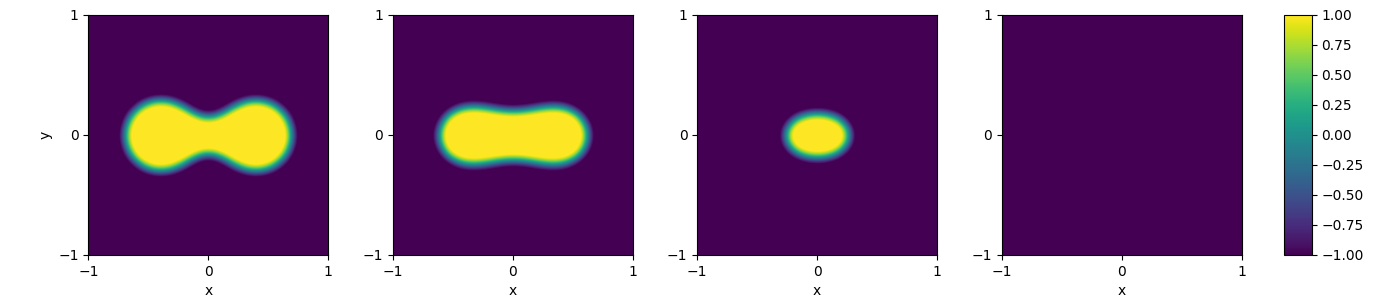}
\includegraphics[width=0.49\textwidth]{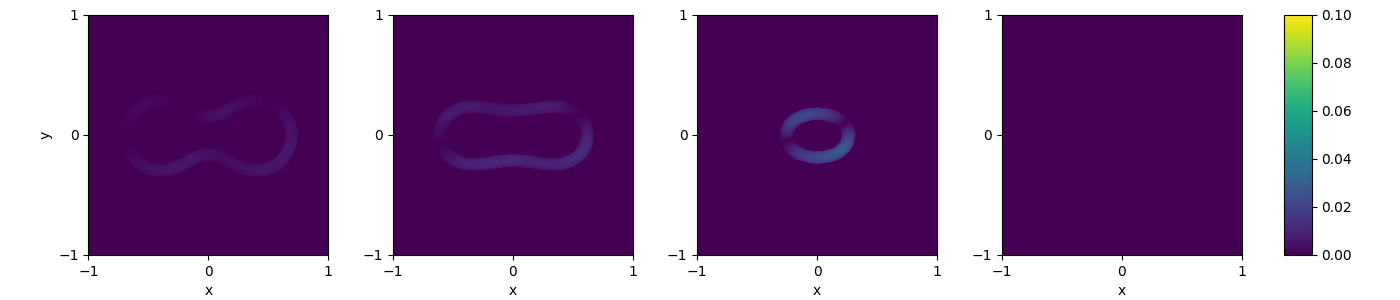}
\includegraphics[width=0.49\textwidth]{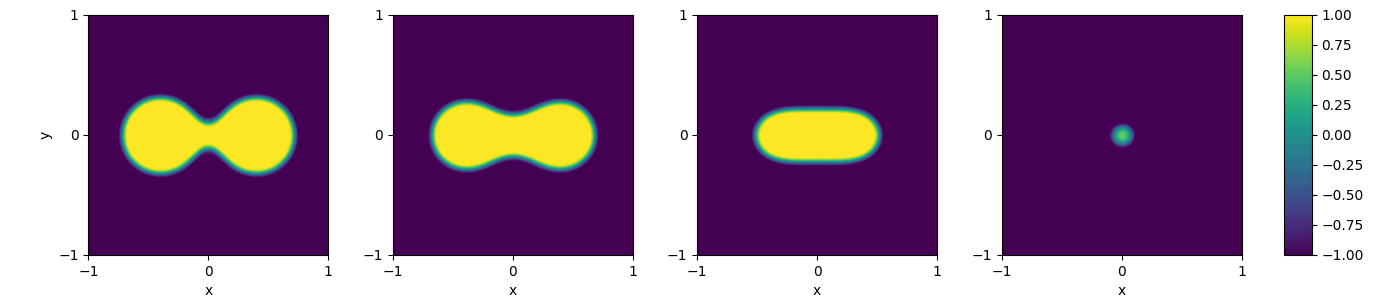}
\includegraphics[width=0.49\textwidth]{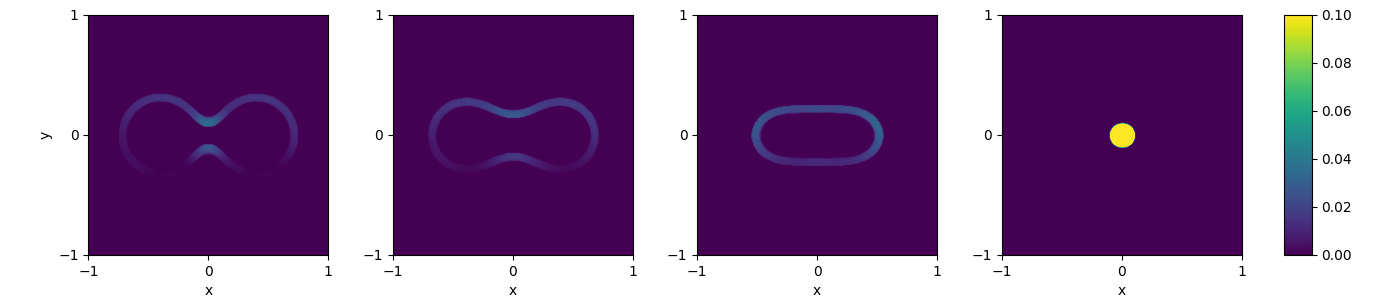}
\includegraphics[width=0.49\textwidth]{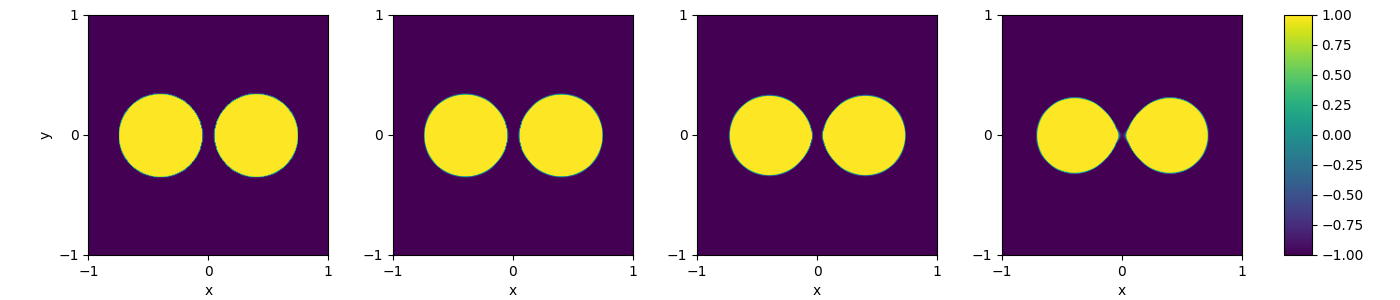}
\includegraphics[width=0.49\textwidth]{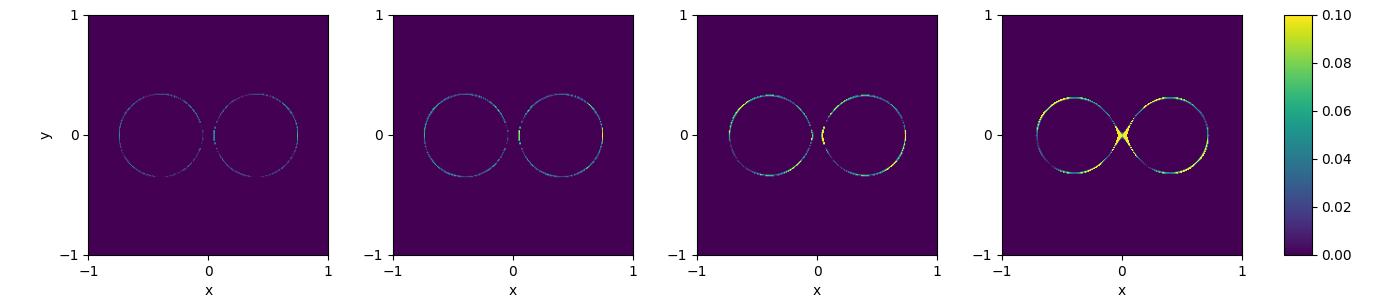}
\caption*{\scriptsize{Obstacle}}
\end{subfigure}
\caption{For two-dimensional merging bubbles example, plots of the NPF-Net prediction solutions (left columns) and the corresponding numerical errors (right columns) for solving the nonlocal AC equation with the regular and obstacle potentials. The rows, from top to bottom, correspond to results at the different values of $\delta=0.05,0.075,0.1$. The columns, from left to right,  display the results at four different times $t=10,20,50,100$.}
\label{fig:diff_delta_merging}
\end{figure}

For the nonlocal AC equation, the NPF-Net model is trained with $N_x=Ny=256$ and $\Delta t= 0.1$. For $N_x=N_y=512$ and $\Delta t = 0.001$, to test for accuracy, NPF-Net solutions are compared to the approximate solutions obtained using the numerical schemes discussed in Section~\ref{sec:DFF}. Figure \ref{fig:AC_2D_mesh_test} illustrates the errors when solving the nonlocal AC equation with the NPF-Net model and the Crank-Nicholson scheme. We observe that, for $\delta=0.05$ and $0.075$, the prediction errors from NPF-Net are similar to those generated by the Crank-Nicholson scheme throughout the simulation. However, for $\delta =0.1$ the prediction errors from NPF-Net are four times larger than those from the numerical simulations. 
\begin{table}[hb!]
\centering
\begin{tabular}{|cc|c|c|c|}
\hline
\multicolumn{2}{|c|}{GPU time (seconds)}                           & $\delta = 0.05$ & $\delta= 0.075$ & $\delta =0.1$ \\ \hline
\multicolumn{1}{|c|}{\multirow{2}{*}{Regular}}  & NPF-Net          & 1.546         & 1.584          & 1.553        \\ \cline{2-5} 
\multicolumn{1}{|c|}{}                          & Numerical scheme & 1.493         & 1.409          & 1.882        \\ \hline
\multicolumn{1}{|c|}{\multirow{2}{*}{Obstacle}} & NPF-Net          & 1.606         & 1.537          & 1.607        \\ \cline{2-5} 
\multicolumn{1}{|c|}{}                          & Numerical scheme & 2.084         & 2.290         & 1.490    \\ \hline   
\end{tabular}
\caption{For the Allen-Cahn equation, a comparison of GPU times (in seconds) for different $\delta$ values.}
\label{tab:AC_times}
\end{table}
To show the efficiency of NPF-Net, in Table \ref{tab:AC_times}, we show the GPU times for simulating solutions until $T=100$ with NPF-Net and a numerical scheme. NPF-Net shows a clear advantage in execution time, remaining largely unaffected by changes in $\delta$ or the type of potential function used. This consistency makes NPF-Net a more efficient and robust solution. 
Figure \ref{fig:diff_delta_merging} presents the predicted solution and associated errors at the times $t=10,20,50,100$ for the regular and obstacle potential functions. From top to bottom, the results are for solving the discretized nonlocal AC equations for $\delta = 0.05,0.075,0.1$. We observe that the two separate bubbles gradually shrink and merge into a smaller bubble which keeps shrinking and finally disappears at different times for the different choices of $\delta$. The prediction errors seem to only occur in the transition regions of two phases, which implies that phase shapes are accurately captured. 

\vskip5pt
\subsubsection{\textbf{Nonlocal Cahn-Hilliard equation}}
\begin{figure}[h!]
    \centering
    \begin{subfigure}{1\textwidth}
        \includegraphics[width=0.32\textwidth]{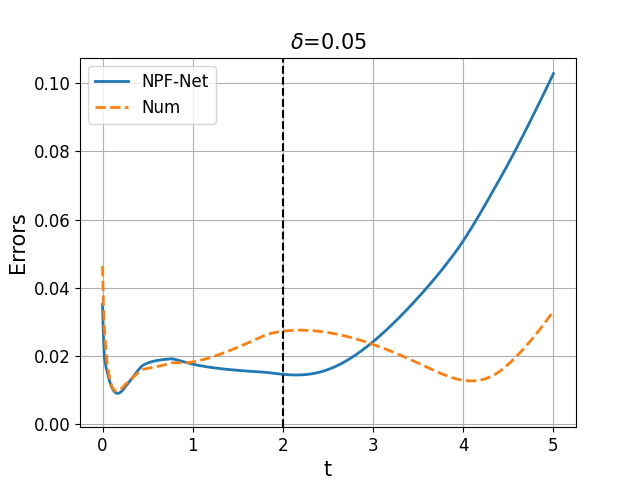}
        \includegraphics[width=0.32\textwidth]{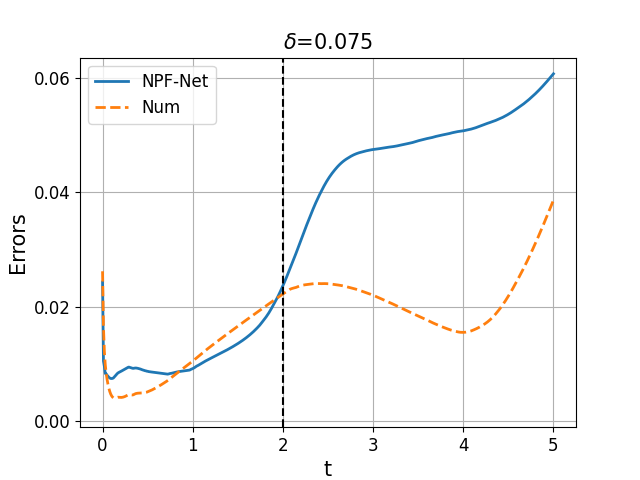}
        \includegraphics[width=0.32\textwidth]{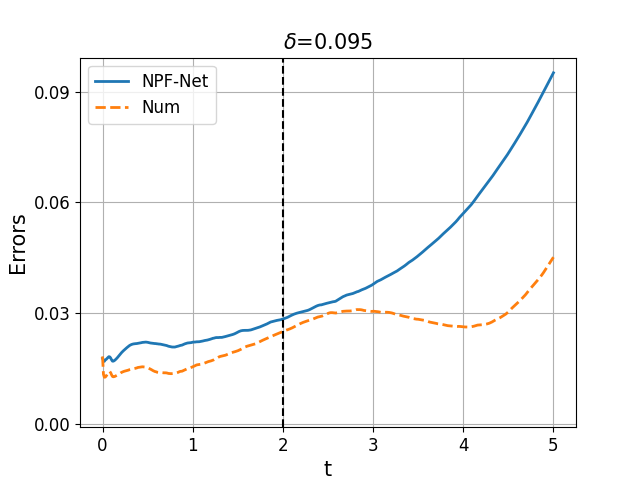}
        \caption*{\scriptsize{Regular}}
    \end{subfigure}
    \begin{subfigure}{1\textwidth}
        \includegraphics[width=0.32\textwidth]{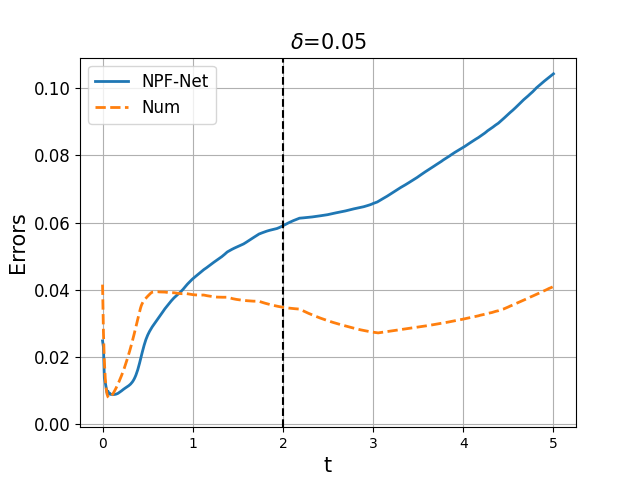}
        \includegraphics[width=0.32\textwidth]{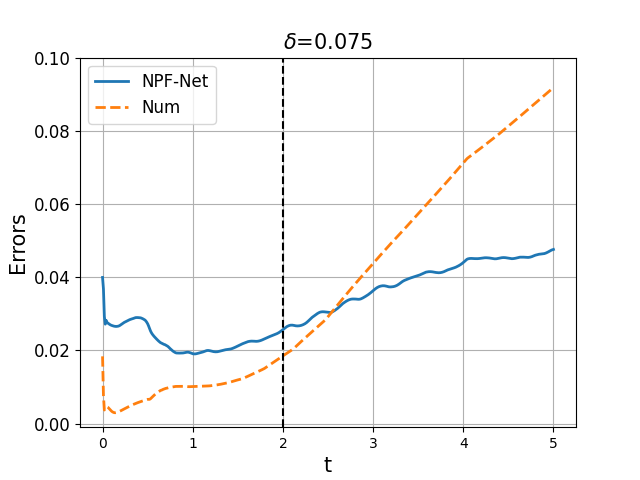}
        \includegraphics[width=0.32\textwidth]{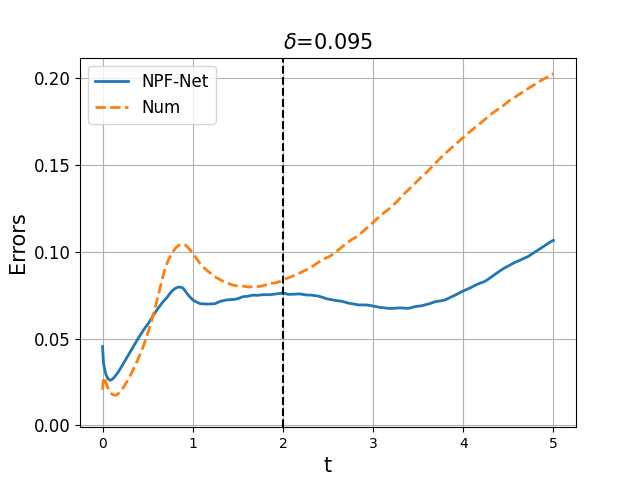}
        \caption*{\scriptsize{Obstacle}}
    \end{subfigure}
    \caption{Errors network simulations and numerical prediction with $N_x=N_y=256$ and $\Delta t=0.01$ for different choice of $\delta = 0.05,0.075, 0.095$. The vertical black dashed line indicates the end of the training time is $T_{train}=2$. }
    \label{fig:CH_2D_mesh_test}
\end{figure}
\begin{figure}[htb!]
\centering
\begin{subfigure}{1\textwidth}
\centering
\includegraphics[width=0.49\textwidth]{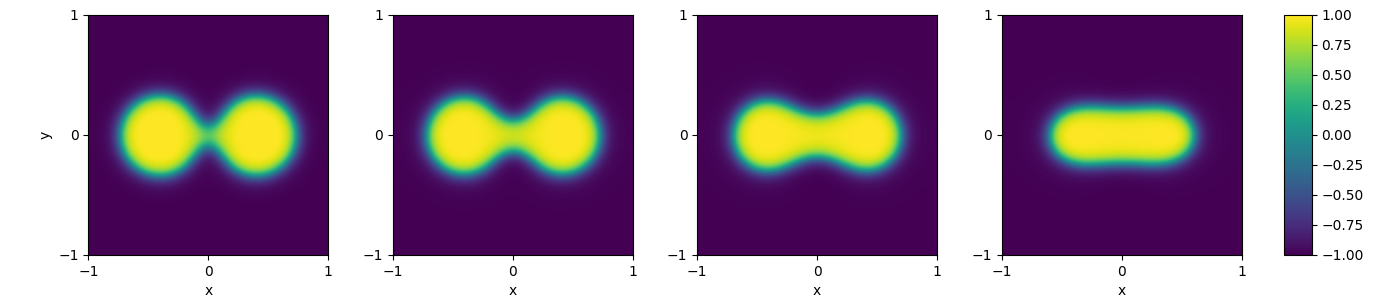}
\includegraphics[width=0.49\textwidth]{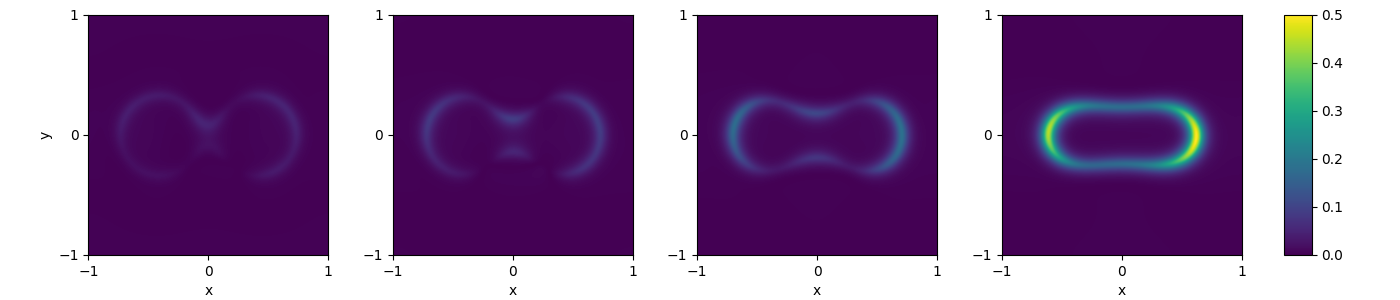}
\includegraphics[width=0.49\textwidth]{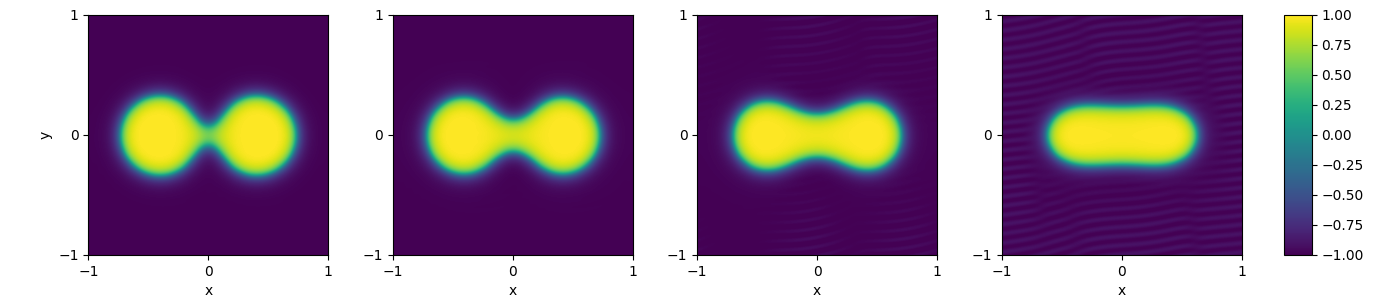}
\includegraphics[width=0.49\textwidth]{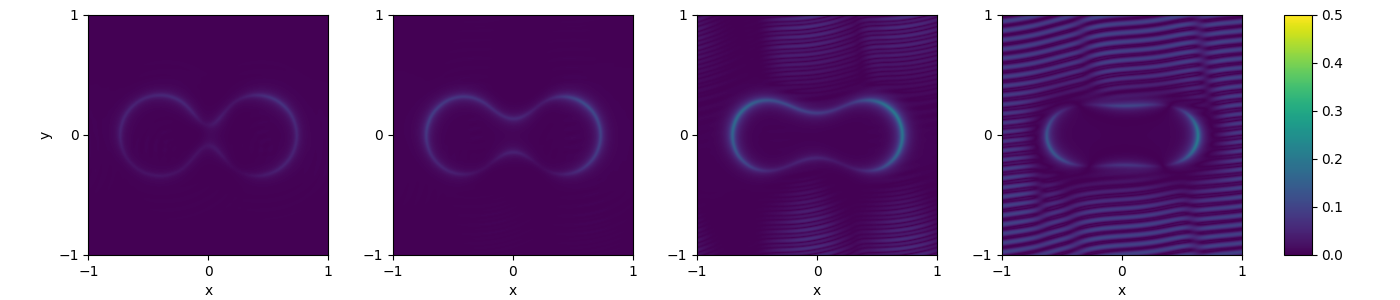}
\includegraphics[width=0.49\textwidth]{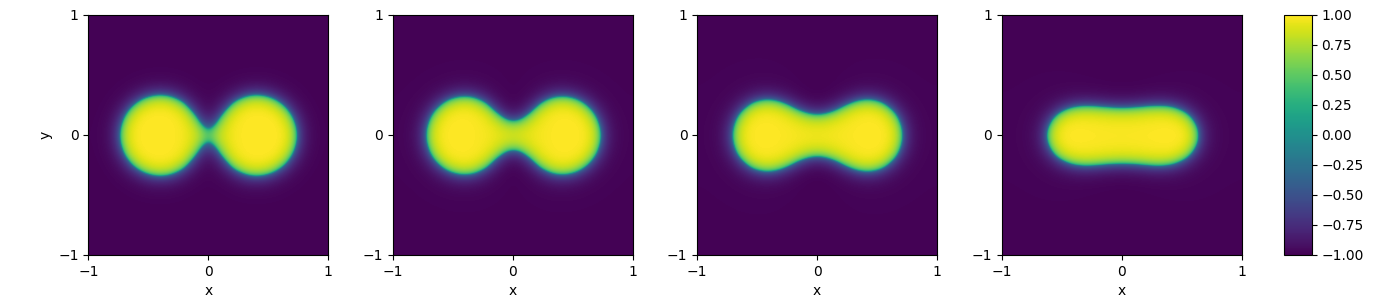}
\includegraphics[width=0.49\textwidth]{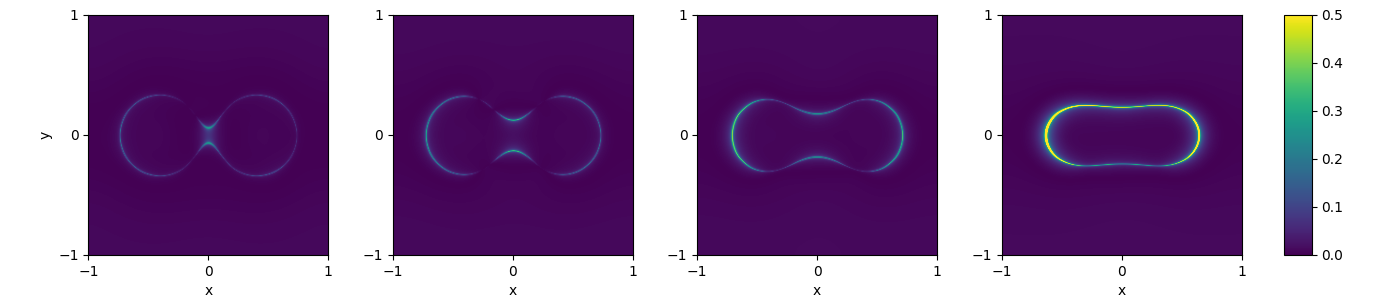}
\caption*{\scriptsize{Regular}}
\end{subfigure}
\begin{subfigure}{1\textwidth}
\centering
\includegraphics[width=0.49\textwidth]{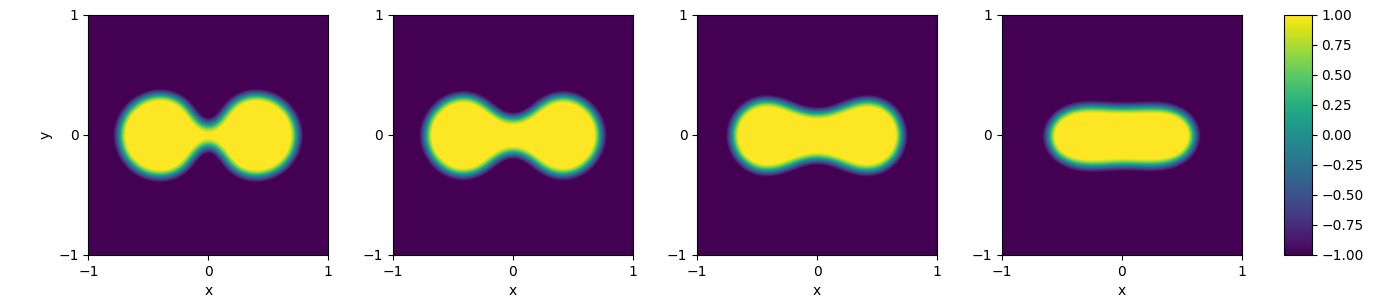}
\includegraphics[width=0.49\textwidth]{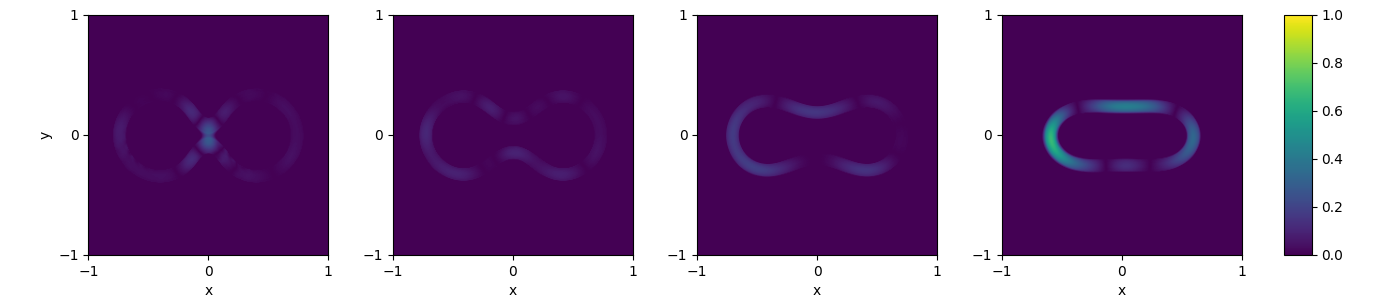}
\includegraphics[width=0.49\textwidth]{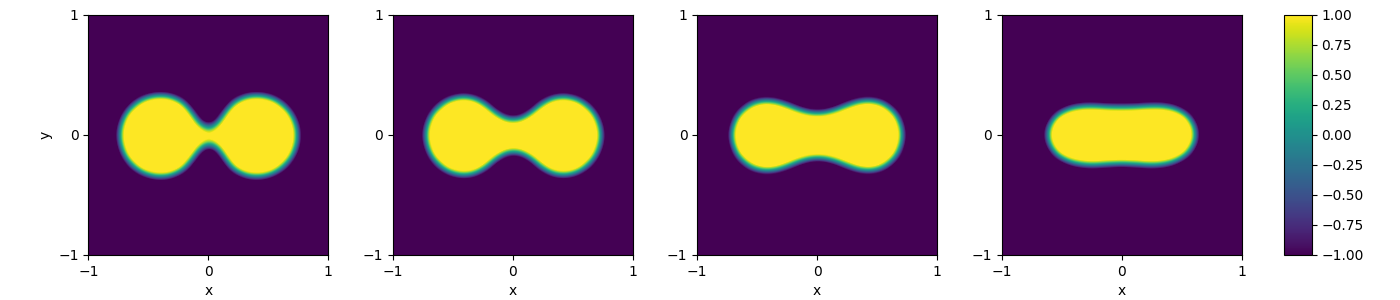}
\includegraphics[width=0.49\textwidth]{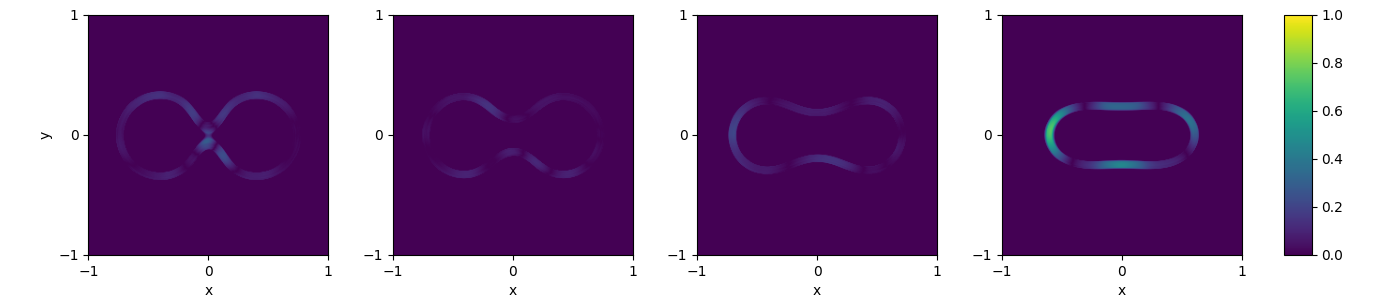}
\includegraphics[width=0.49\textwidth]{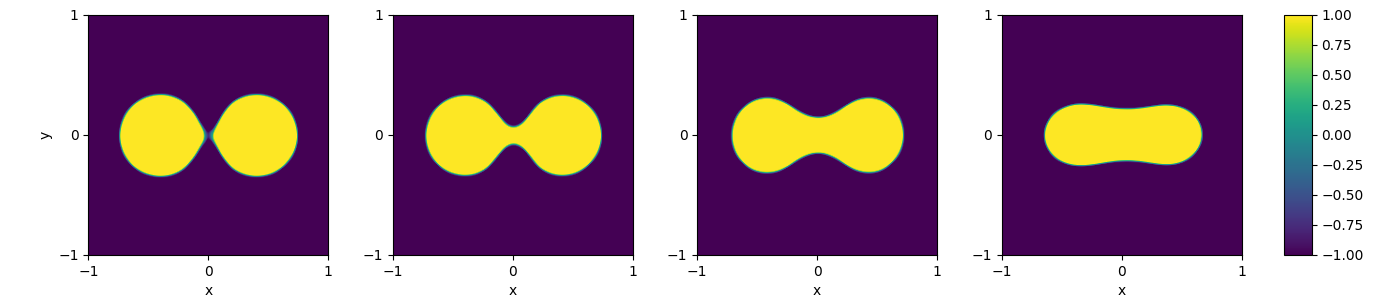}
\includegraphics[width=0.49\textwidth]{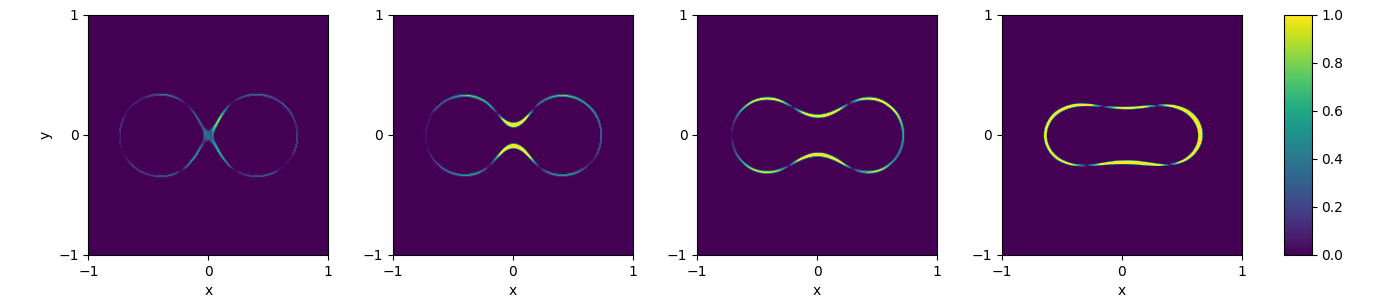}
\caption*{\scriptsize{Obstacle}}
\end{subfigure}
\caption{Plots of the NPF-Net prediction solutions (left columns) and the corresponding numerical errors (right columns) for solving the nonlocal CH equation with regular and obstacle potentials in the two-dimensional merging bubbles example. The rows, from top to bottom, correspond to results at different values of $\delta=0.05,0.075,0.095$. The columns display the results at four different times $t=0.5,1,2,5$.}
\label{fig:CH_diff_delta_merging}
\end{figure}

For the nonlocal CH equation, the NPF-Net model was trained with $N_x=N_y=256$, for $T_{train}=2$ with $\Delta t= 0.01$. For $N_x=N_y=512$ and $\Delta t = 1e-5$, to test for accuracy, NPF-Net solutions are compared to the approximate solutions obtained using the numerical schemes discussed in Section~\ref{sec:DFF}. Figure \ref{fig:CH_2D_mesh_test} illustrates the comparison of errors of NPF-Net and the numerical solver for different choices of $\delta$. We observe that the prediction errors for NPF-Net are comparable with that of the numerical solver. In Table \ref{tab:CH_times}, the GPU times for simulating the solutions up to $T=5$ are presented for both the numerical scheme and NPF-Net. The GPU time for the numerical scheme increases with larger $\delta$ due to the additional iterations required. However, the GPU times for NPF-Net remain stable, unaffected by changes in $\delta$ or the choice of potential function. Overall, NPF-Net is significantly more efficient in simulating the solution compared to the numerical scheme based on iterative methods. It is worth noting that the iterative method converges faster when solving the nonlocal AC equation compared to the nonlocal CH equation, resulting in lower GPU times for solving nonlocal AC than for nonlocal CH.
\begin{table}[h!]
\centering
\begin{tabular}{|cc|c|c|c|}
\hline
\multicolumn{2}{|c|}{GPU time (seconds)}                           & $\delta = 0.05$ & $\delta= 0.075$ & $\delta =0.095$ \\ \hline
\multicolumn{1}{|c|}{\multirow{2}{*}{Regular}}  & NPF-Net          & 0.816         & 0.782        & 0.763       \\ \cline{2-5} 
\multicolumn{1}{|c|}{}                          & Numerical scheme & 20.512        & 62.472         & 422.592      \\ \hline
\multicolumn{1}{|c|}{\multirow{2}{*}{Obstacle}} & NPF-Net          & 0.883    & 0.848         & 0.812     \\ \cline{2-5} 
\multicolumn{1}{|c|}{}                          & Numerical scheme & 15.647      & 46.877          & 242.920    \\ \hline   
\end{tabular}
\caption{For the Cahn-Hilliard equation, a comparison of GPU times (in seconds) for different $\delta$ values.}
\label{tab:CH_times}
\end{table}
Figure \ref{fig:CH_diff_delta_merging} presents the predicted solution and associated errors at the times $t=0.5,1,2,5$ for the regular and obstacle potential functions. From top to bottom, the results are for solving the nonlocal CH equations for $\delta = 0.05,0.075,0.095$. We observe that the two separate bubbles gradually merge into one bubble. 

\subsection{\textbf{Two-dimensional grain coarsening}}

Next, we use the trained NPF-Net to simulate the two-dimensional grain coarsening problem for which the initial conditions are chosen randomly for sharp-colored noise $u_0^2$ and white noise $u_0^3$ as illustrated in Figure \ref{fig:initials} and which are then applied to nonlocal AC and CH models. Both regular potential and obstacle potential are considered.

\subsubsection{\textbf{Nonlocal Allen-Cahn equation}}
\begin{figure}[h!]
\centering
\begin{subfigure}{1\textwidth}
\centering
\includegraphics[width=0.49\textwidth]{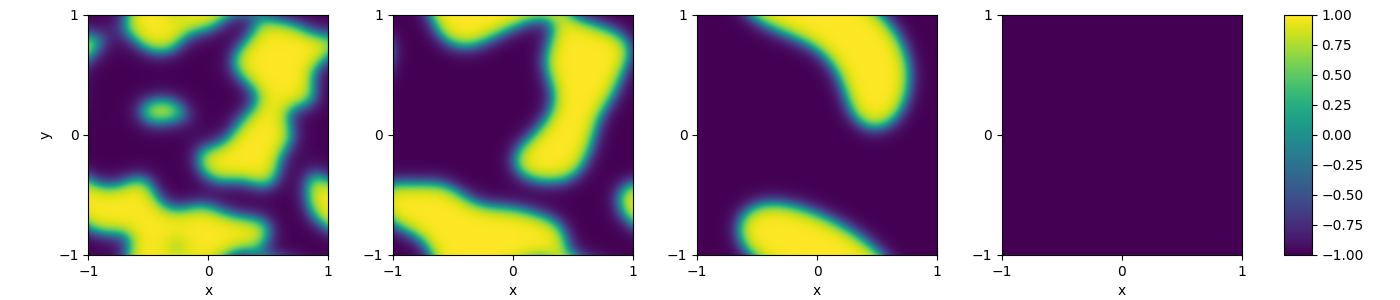}
\includegraphics[width=0.49\textwidth]{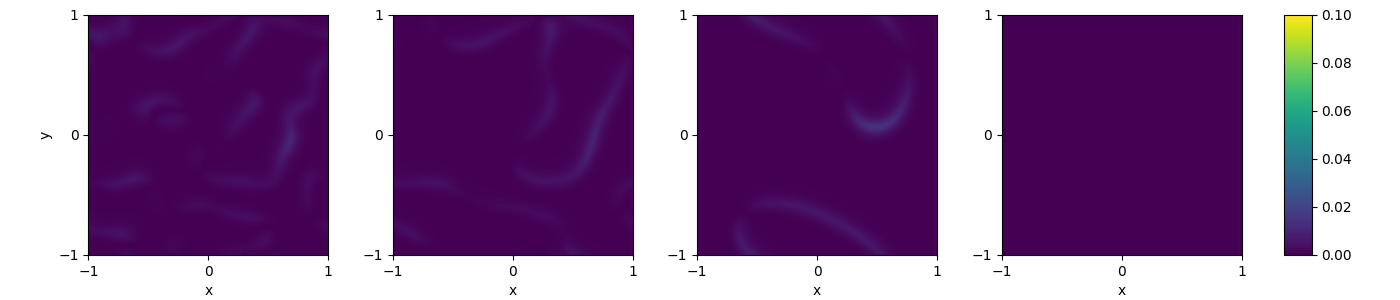}
\includegraphics[width=0.49\textwidth]{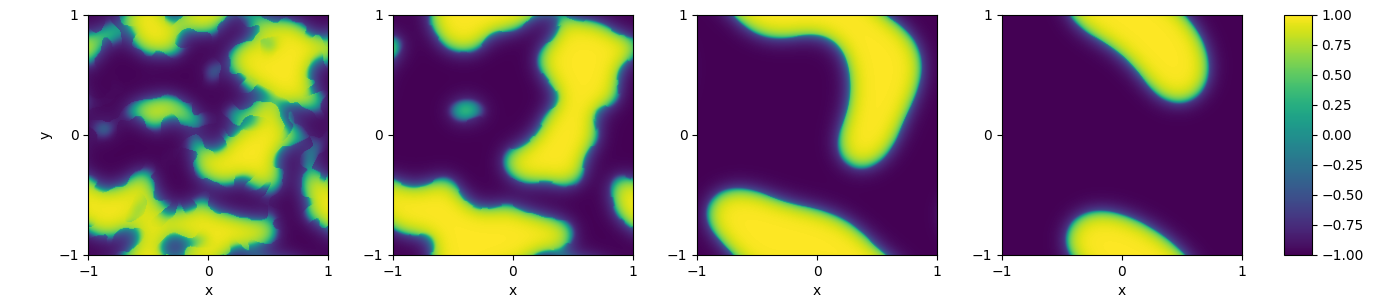}
\includegraphics[width=0.49\textwidth]{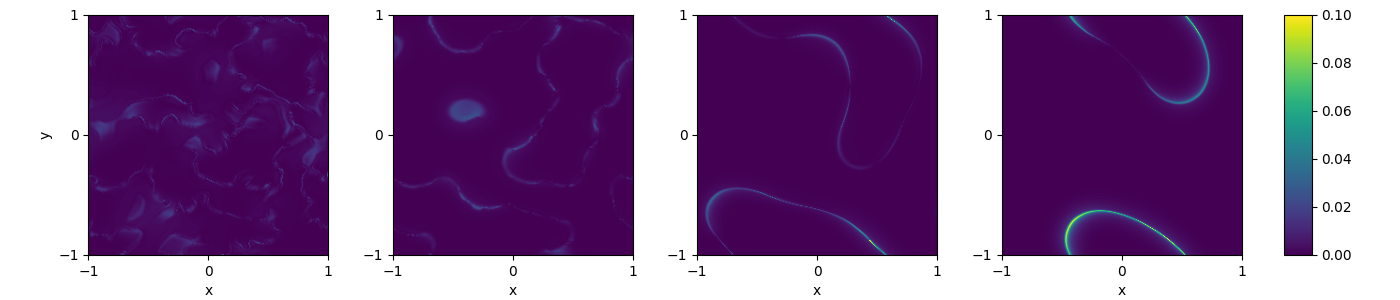}
\caption*{\scriptsize{Regular}}
\end{subfigure}
\begin{subfigure}{1\textwidth}
\centering
\includegraphics[width=0.49\textwidth]{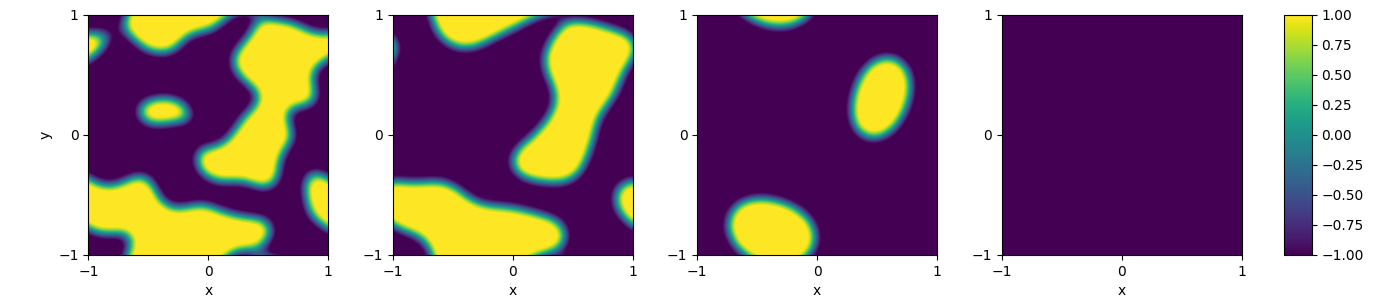}
\includegraphics[width=0.49\textwidth]{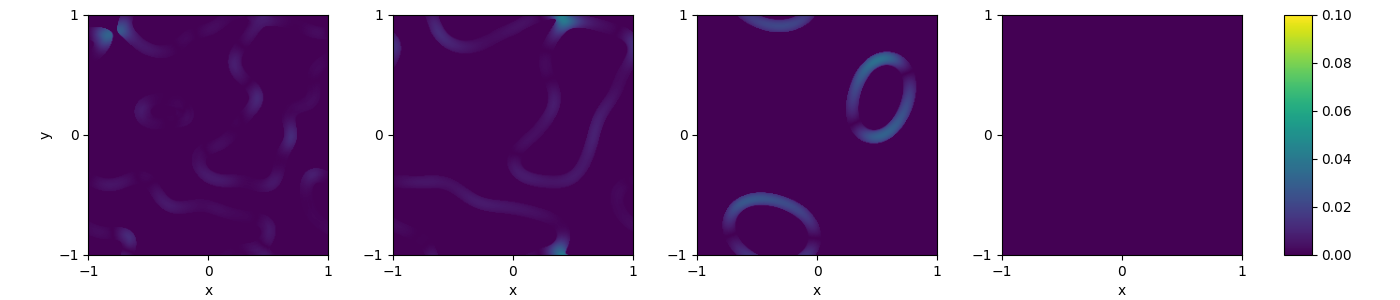}
\includegraphics[width=0.49\textwidth]{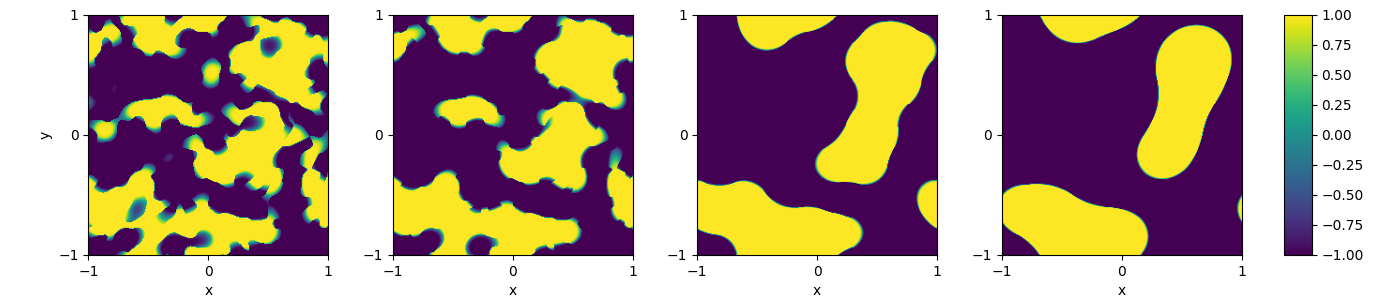}
\includegraphics[width=0.49\textwidth]{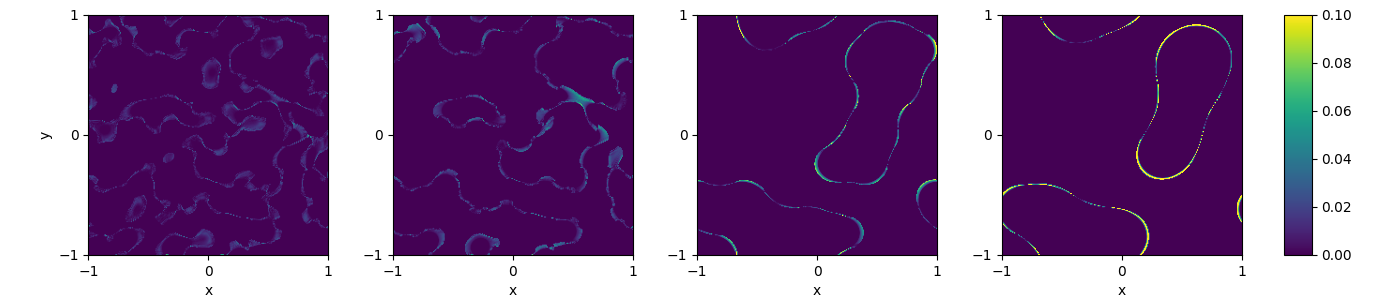}
\caption*{\scriptsize{Obstacle}}
\end{subfigure}
\caption{
Plots of the NPF-Net prediction solutions (left columns) and the corresponding numerical errors (right columns) for
solving the nonlocal CH equation with regular and obstacle potentials in the 2D grain coarsening example. Colored noise initial condition is used.  The rows, from Top to bottom, correspond to results at different values of $\delta=0.05,0.1$. The columns display the results at four different times $t = 5,10,50, 150$.}
\label{fig:diff_delta_colored_noise}
\end{figure}

\begin{figure}[ht!]
\centering
\begin{subfigure}{1\textwidth}
\centering
\includegraphics[width=0.49\textwidth]{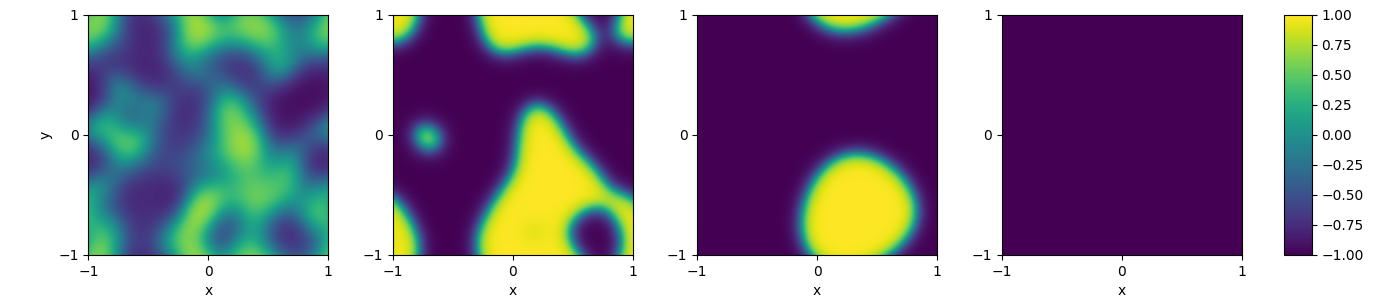}
\includegraphics[width=0.49\textwidth]{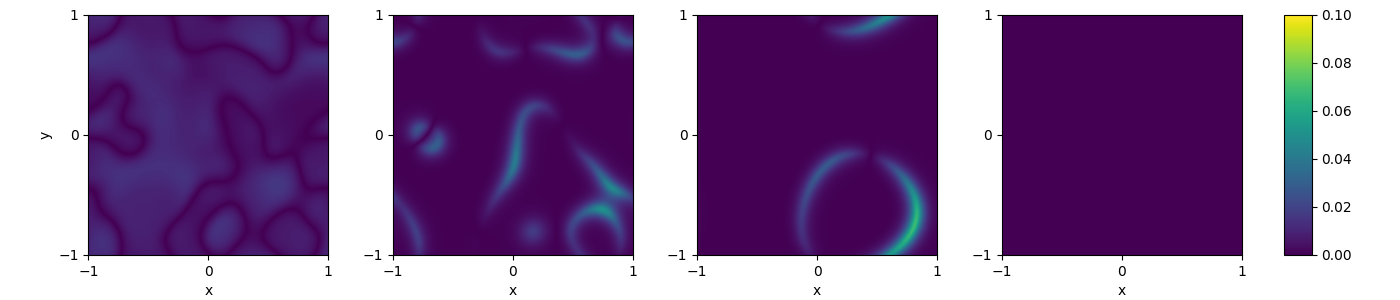}
\includegraphics[width=0.49\textwidth]{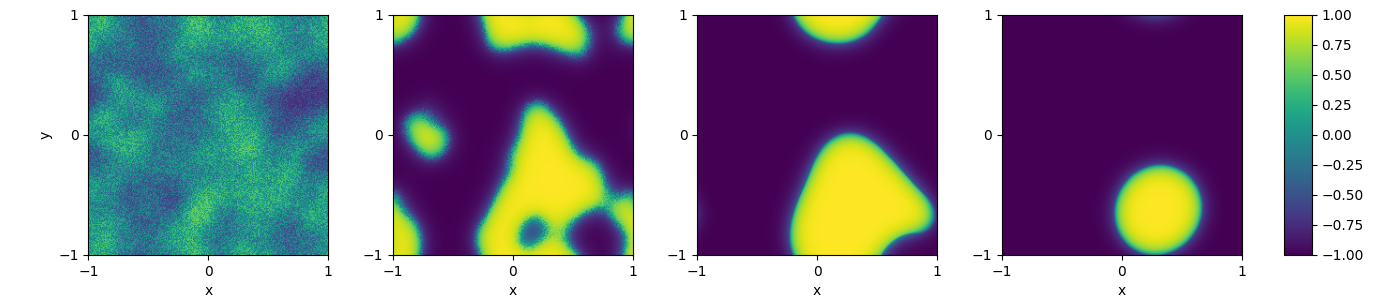}
\includegraphics[width=0.49\textwidth]{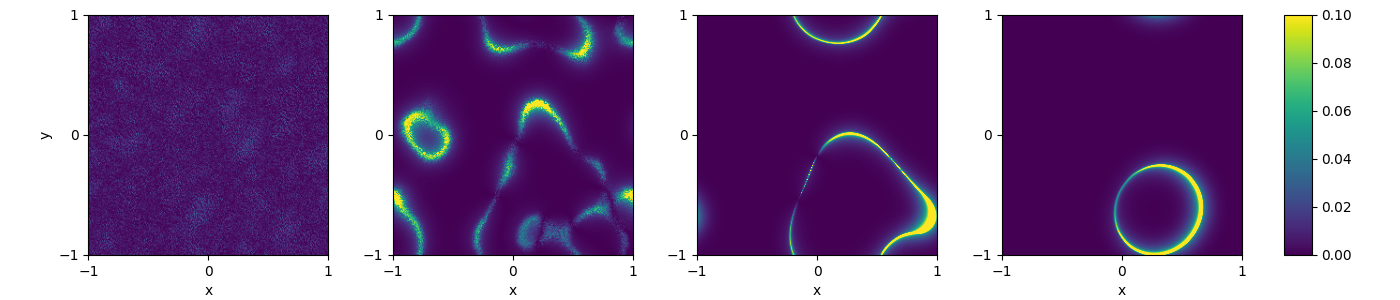}
\caption*{\scriptsize{Regular}}
\end{subfigure}
\begin{subfigure}{1\textwidth}
\centering
\includegraphics[width=0.49\textwidth]{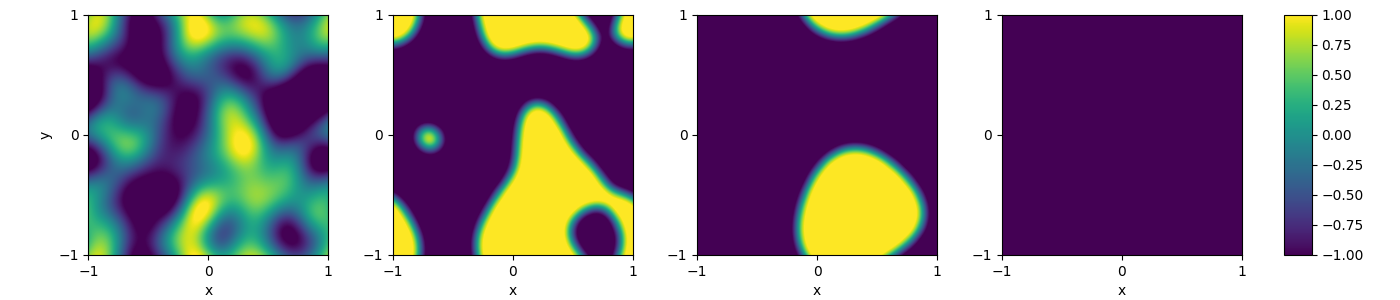}
\includegraphics[width=0.49\textwidth]{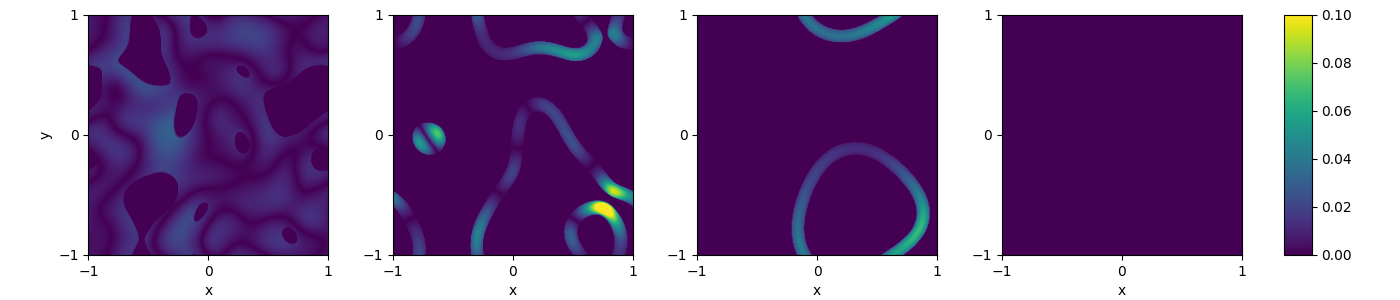}
\includegraphics[width=0.49\textwidth]{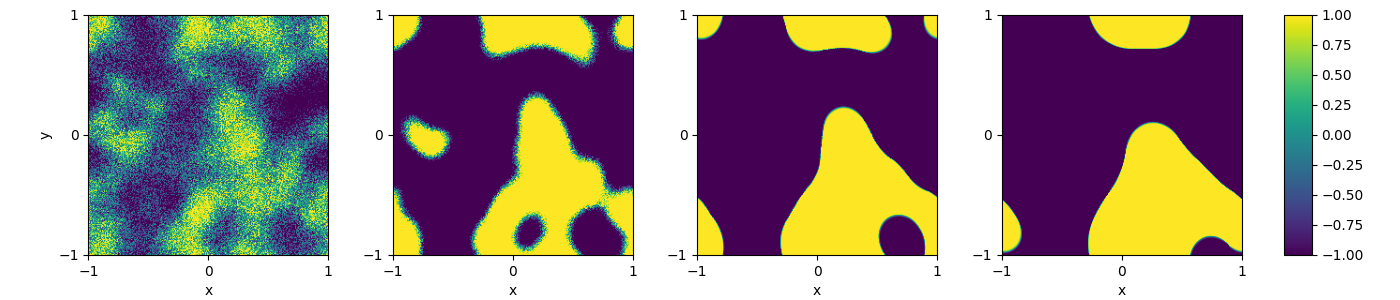}
\includegraphics[width=0.49\textwidth]{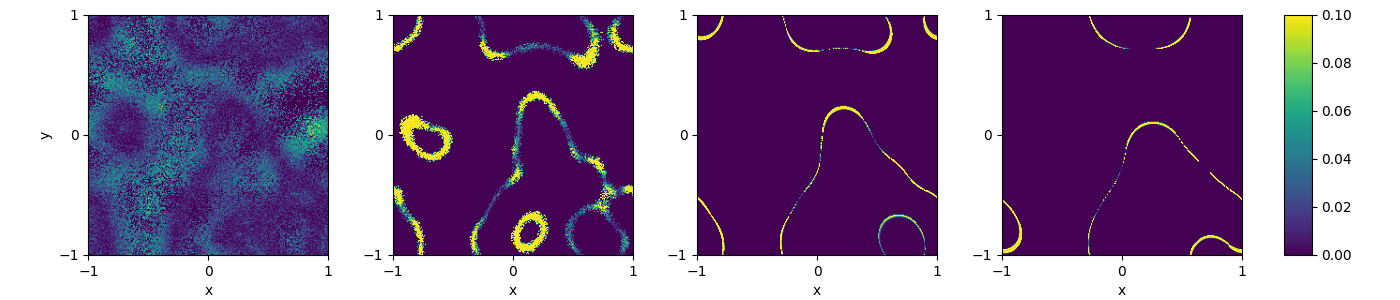}
\caption*{\scriptsize{Obstacle}}
\end{subfigure}
\caption{
Plots of the NPF-Net prediction solutions (left columns) and the corresponding numerical errors (right columns) for
solving the nonlocal CH equation with regular and obstacle potentials in the 2D grain coarsening example. White noise initial condition is used. The rows, from Top to bottom, correspond to results at different values of $\delta=0.05,0.1$. The columns display the results at four times $t = 5,10,50, 150$.}
\label{fig:diff_delta_white_noise}
\end{figure}

\begin{figure}[h!]
    \centering
    \begin{subfigure}{0.38\textwidth}
        \includegraphics[width=1\textwidth]{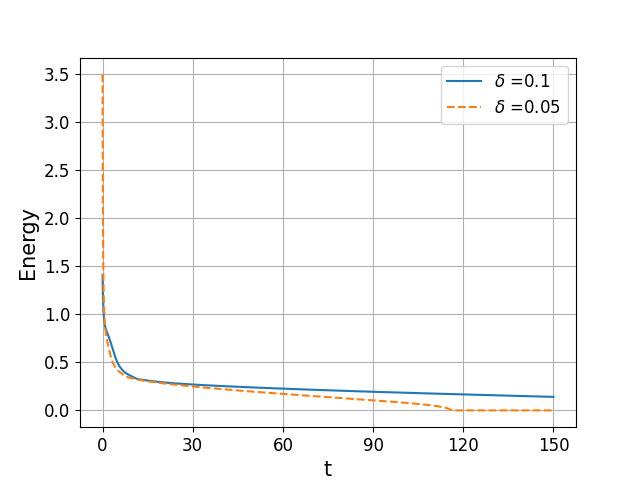}
        \caption*{\scriptsize{Regular}}
    \end{subfigure}
    \begin{subfigure}{0.38\textwidth}
        \includegraphics[width=1\textwidth]{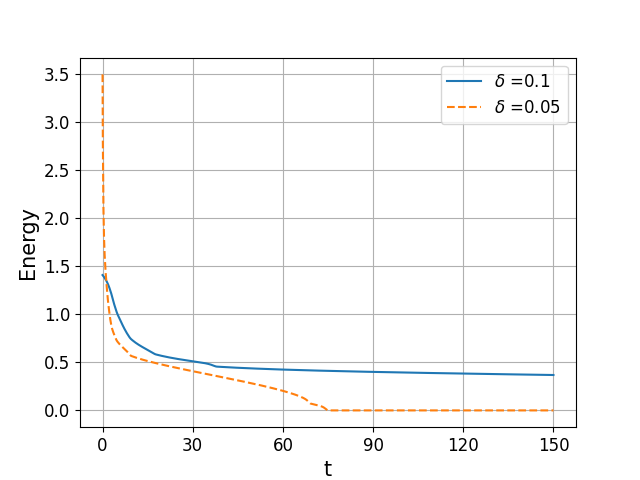}
        \caption*{\scriptsize{Obstacle}}
    \end{subfigure}
    \caption{Evolution of the nonlocal energy~\eqref{eq:GL_energy} for the solution of nonlocal AC equation obtained by the NN simulations with sharp-colored noise initial condition, ($\epsilon = 0.05$, $N_x=N_y = 256$, and $\Delta t = 0.1$).}
    \label{fig:Colored_energy}
\end{figure}

\begin{figure}[h!]
    \centering
    \begin{subfigure}{0.38\textwidth}        \includegraphics[width=1\textwidth]{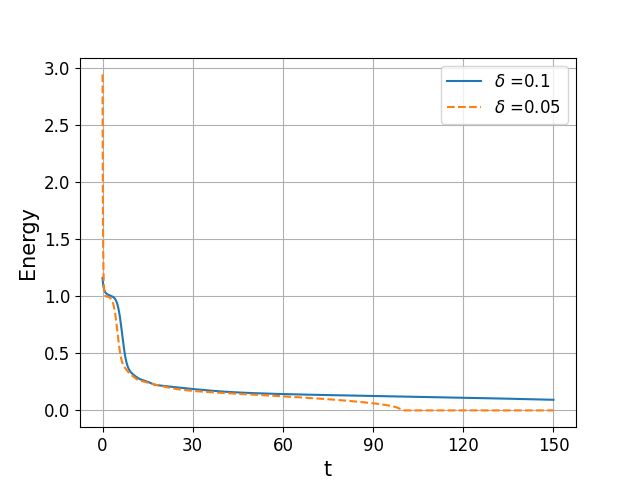}
        \caption*{\scriptsize{Regular}}
    \end{subfigure}
    \begin{subfigure}{0.38\textwidth}
        \includegraphics[width=1\textwidth]{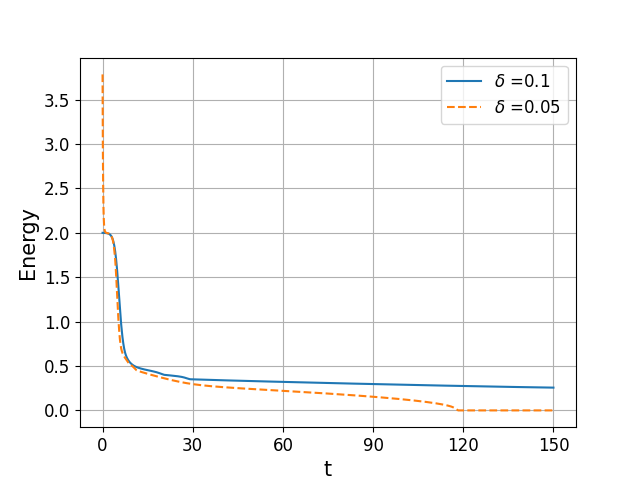}
        \caption*{\scriptsize{Obstacle}}
    \end{subfigure}
    \caption{Evolution of the nonlocal energy~\eqref{eq:GL_energy} for the solution of nonlocal AC equation obtained by the NN simulations with white noise initial condition, ($\epsilon = 0.05$, $N_x=N_y = 256$, and $\Delta t = 0.1$). }
    \label{fig:white_energy}
\end{figure}

For the nonlocal AC equation for the regular and obstacle potentials, in Figures \ref{fig:diff_delta_colored_noise} and \ref{fig:diff_delta_white_noise} which correspond to the initial conditions $u_0^2$ and $u_0^3$, respectively, we plot the predicted solutions obtained by NPF-Net and the corresponding prediction errors at the times $t=5,10,50,150$ (left to right) and, from top to bottom, the simulation solutions and errors at $\delta =0.05$ and 0.1. In the process of grain coarsening, one phase gradually becomes smaller and smaller and eventually vanishes in time. In Figure \ref{fig:Colored_energy} and \ref{fig:white_energy}, the evolution of the energy of the predicted solution with regular and obstacle potential are plotted respectively. We observe that in both cases the NN preserved the energy dissipation over time.

\subsubsection{\textbf{Nonlocal Cahn-Hilliard equation}}

\begin{figure}[h!]
\centering
\begin{subfigure}{1\textwidth}
\centering
\includegraphics[width=0.49\textwidth]{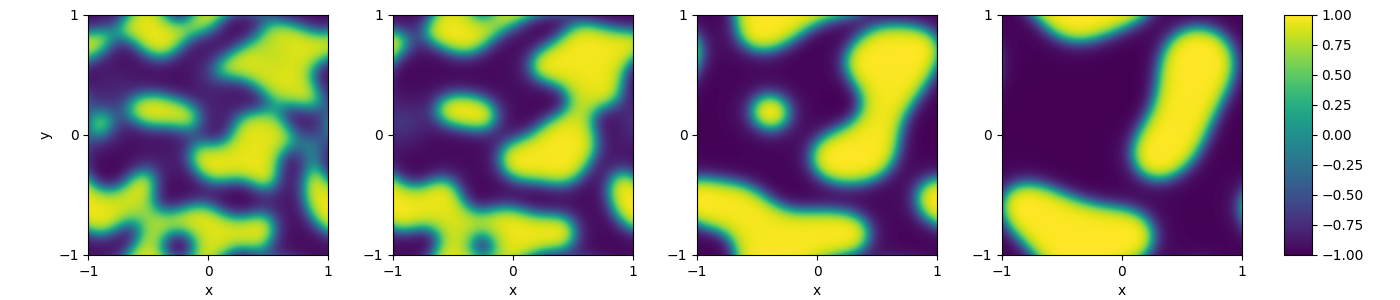}
\includegraphics[width=0.49\textwidth]{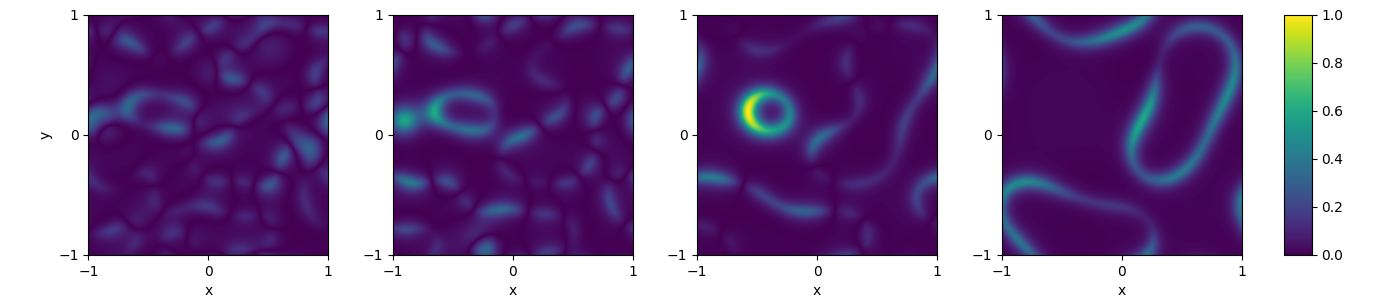}
\includegraphics[width=0.49\textwidth]{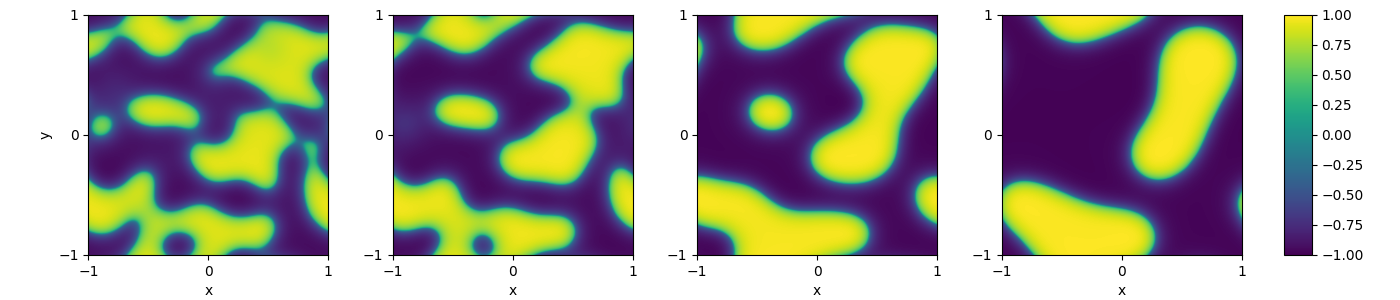}
\includegraphics[width=0.49\textwidth]{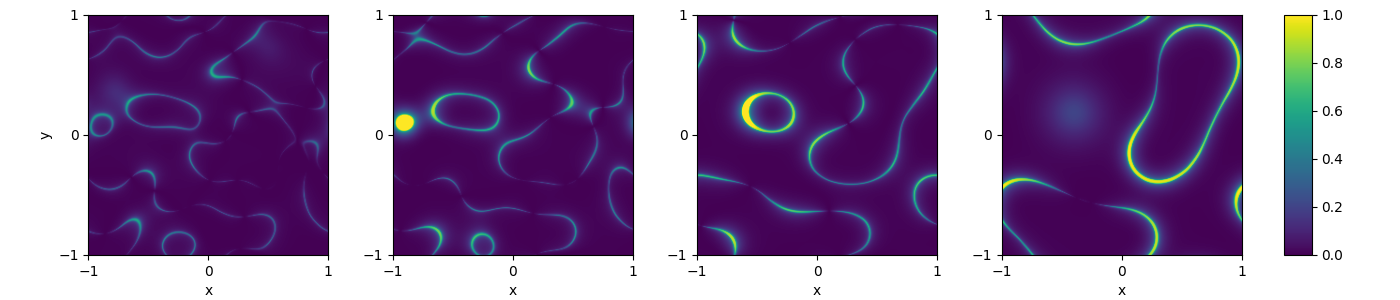}
\caption*{\scriptsize{Regular}}
\end{subfigure}
\begin{subfigure}{1\textwidth}
\centering
\includegraphics[width=0.49\textwidth]{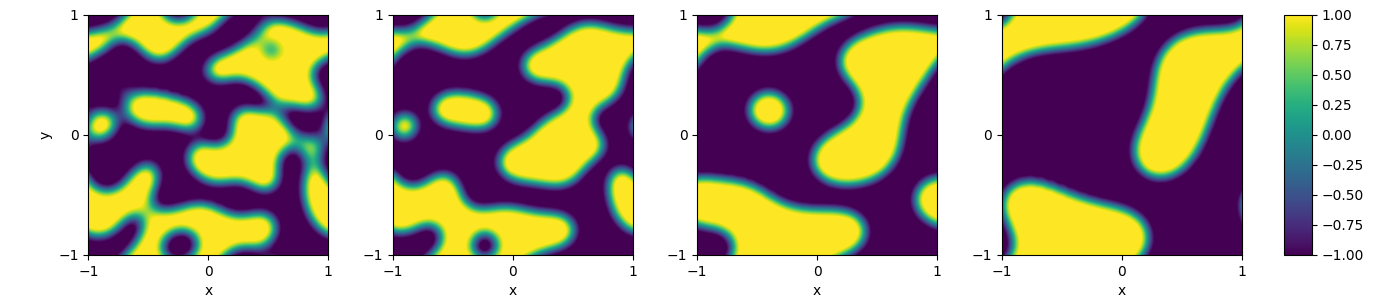}
\includegraphics[width=0.49\textwidth]{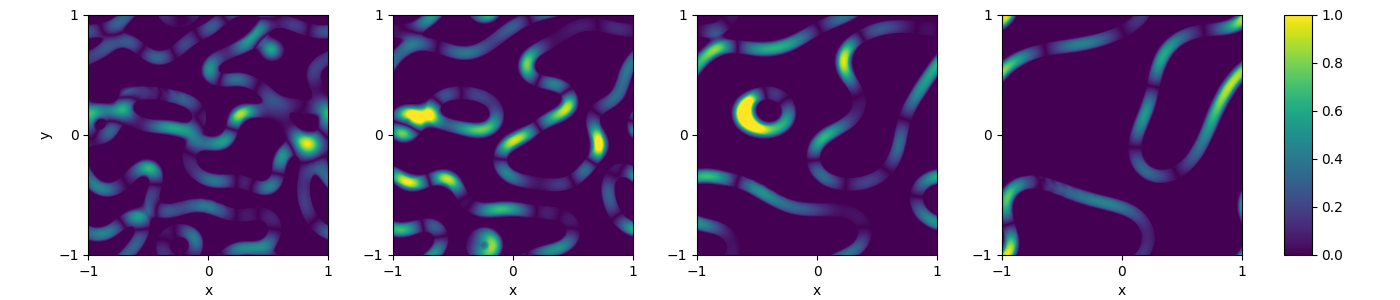}
\includegraphics[width=0.49\textwidth]{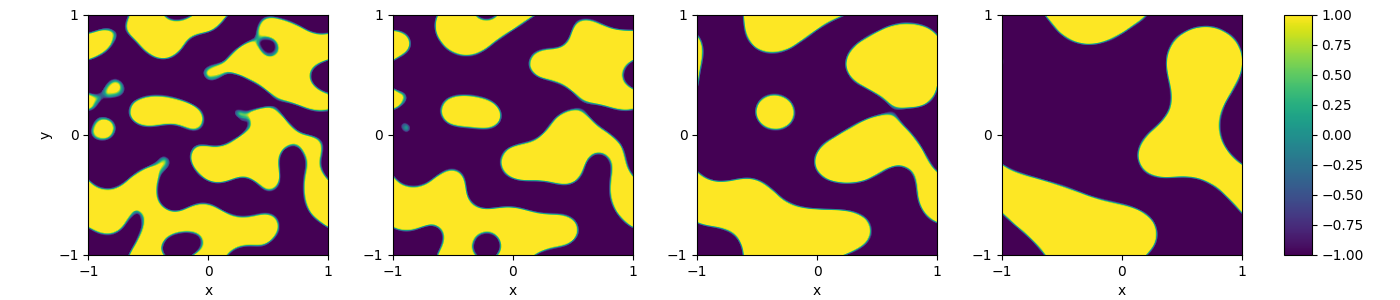}
\includegraphics[width=0.49\textwidth]{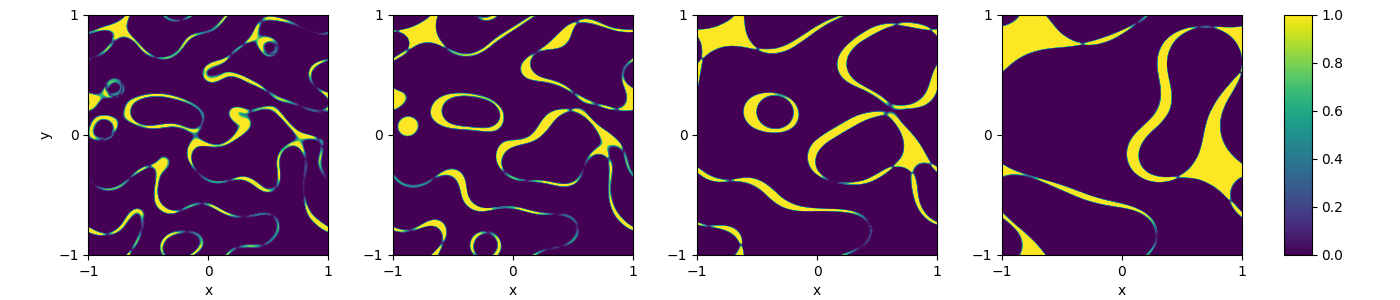}
\caption*{\scriptsize{Obstacle}}
\end{subfigure}
\caption{
Plots of the NPF-Net prediction solutions (left columns) and the corresponding numerical errors (right columns) for
solving the nonlocal CH equation with regular and obstacle potentials in the 2D grain coarsening example. Sharp-colored initial condition is used. From Top to bottom, the rows correspond to results at different values of $\delta=0.05,0.095$. The columns display the results at four different times $t = 0.5,1,2,5$.}
\label{fig:CH_diff_delta_sharp_noise}
\end{figure}

\begin{figure}[h!]
    \centering
    \begin{subfigure}{0.38\textwidth}
    \includegraphics[width=1\textwidth]{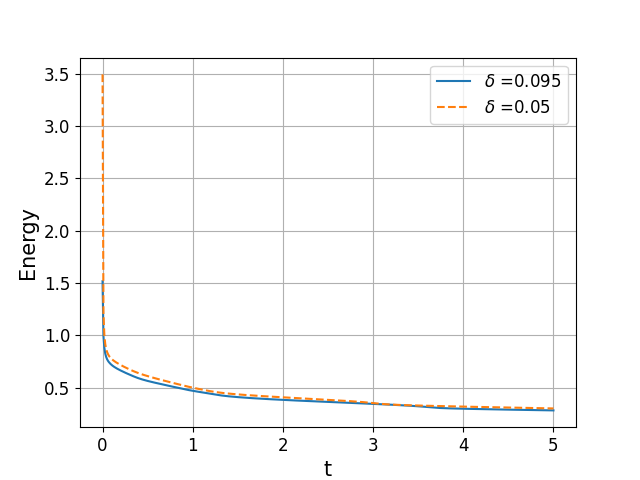}
        \caption*{\scriptsize{Regular}}
    \end{subfigure}
    \begin{subfigure}{0.38\textwidth}
        \includegraphics[width=1\textwidth]{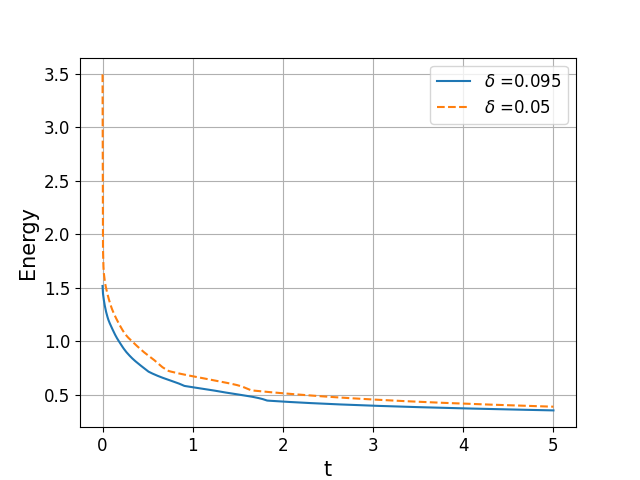}
        \caption*{\scriptsize{Obstacle}}
    \end{subfigure}
    \caption{Evolution of the nonlocal energy~\eqref{eq:GL_energy} for the solution of nonlocal CH equation obtained by the NN simulations with sharp-colored noise initial condition, ($\epsilon = 0.05$, $N_x=N_y = 256$, and $\Delta t = 0.01$).}
    \label{fig:CH_color_energy}
\end{figure}

For the nonlocal CH equation for the regular and obstacle potentials, in Figure~\ref{fig:CH_diff_delta_sharp_noise} which corresponds to the sharp-colored noise initial condition $u_0^2$, we plot the predicted solutions obtained by NPF-Net and the corresponding prediction errors at the times $t=0.5,1,2,5$ (left to right) and the two values of $\delta=0.05,0.095$ (top to bottom ). We observe a slight increase in errors over time for the obstacle potential function. However, despite the sharpness of the solution, our network model effectively learns and captures this behavior, preserving the sharp interfaces throughout the simulation. On Figure~\ref{fig:CH_color_energy} we also plot the evolution for the nonlocal energy~\eqref{eq:GL_energy}. We observe that the NN approach preserves energy decay over time for both regular and obstacle potential functions.

\subsection{\textbf{3D Allen-Cahn equation}}
\begin{figure}[h!]
\begin{subfigure}{0.98\textwidth}
\centering
\includegraphics[width=0.2\textwidth]{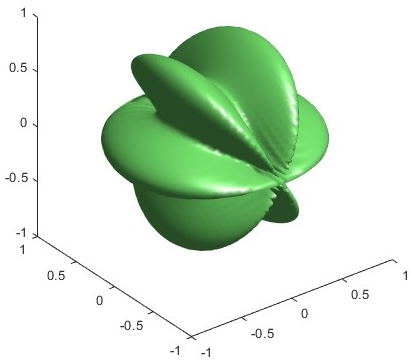}
\includegraphics[width=0.2\textwidth]{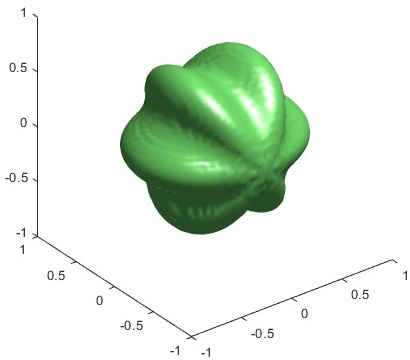}
\includegraphics[width=0.2\textwidth]{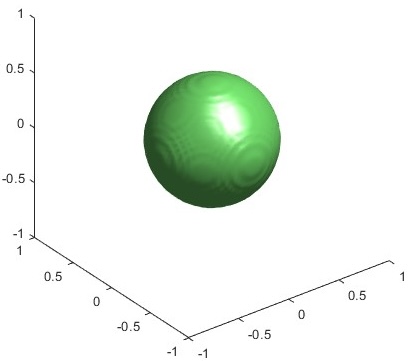}
\includegraphics[width=0.2\textwidth]{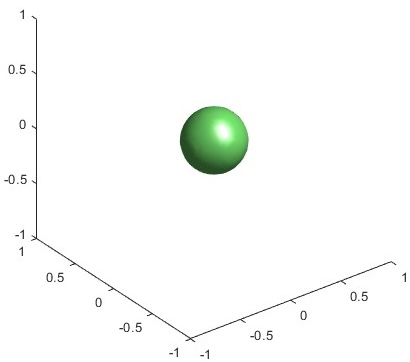}
\includegraphics[width=0.2\textwidth]{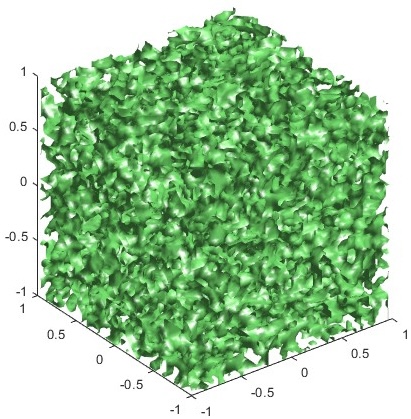}
\includegraphics[width=0.2\textwidth]{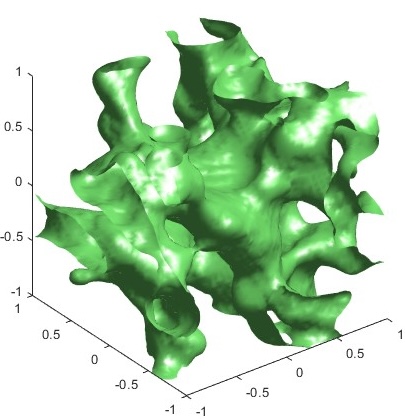}
\includegraphics[width=0.2\textwidth]{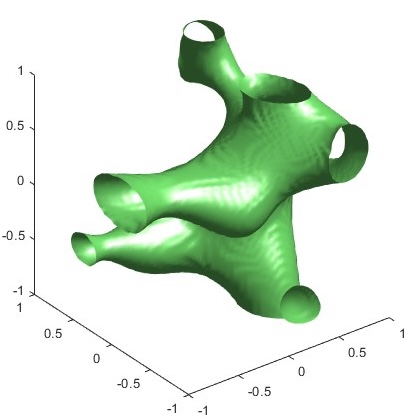}
\includegraphics[width=0.2\textwidth]{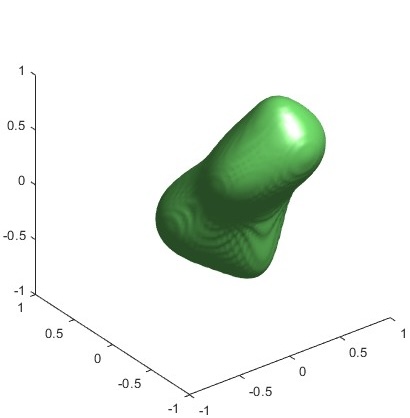}
\caption*{\scriptsize{Regular}}
\end{subfigure}
\begin{subfigure}{0.98\textwidth}
\centering
\includegraphics[width=0.2\textwidth]{3D/star_initial.jpg}
\includegraphics[width=0.2\textwidth]{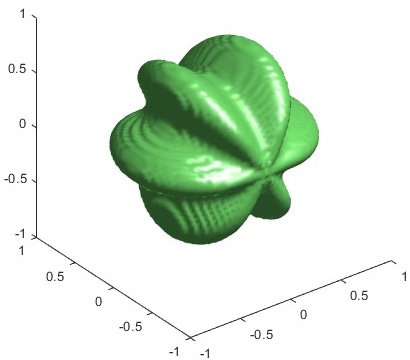}
\includegraphics[width=0.2\textwidth]{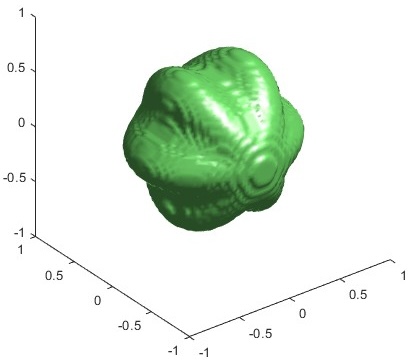}
\includegraphics[width=0.2\textwidth]{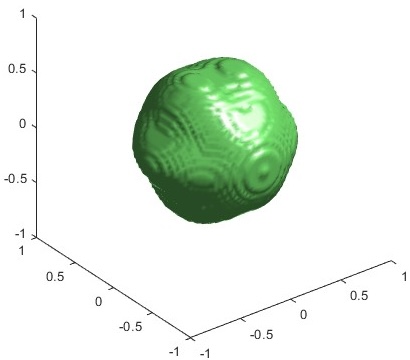}
\includegraphics[width=0.2\textwidth]{3D/initial.jpg}
\includegraphics[width=0.2\textwidth]{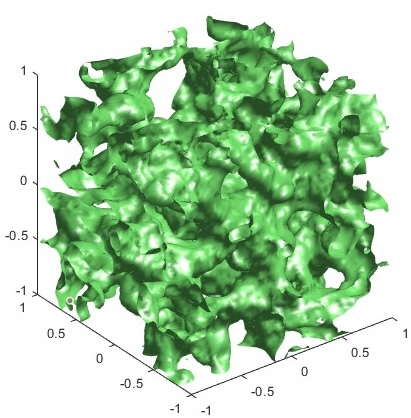}
\includegraphics[width=0.2\textwidth]{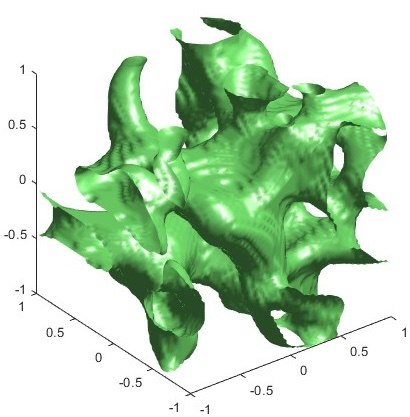}
\includegraphics[width=0.2\textwidth]{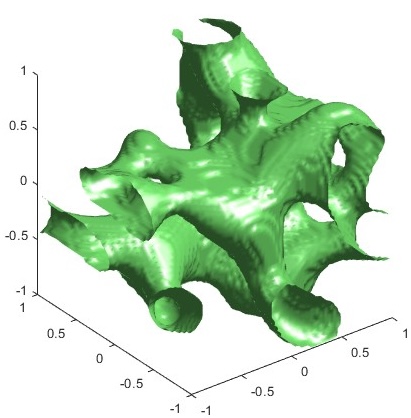}
\caption*{\scriptsize{Obstacle}}
\end{subfigure}
\centerline{
\begin{subfigure}{0.4\textwidth}
\centering
\includegraphics[width=1\textwidth]{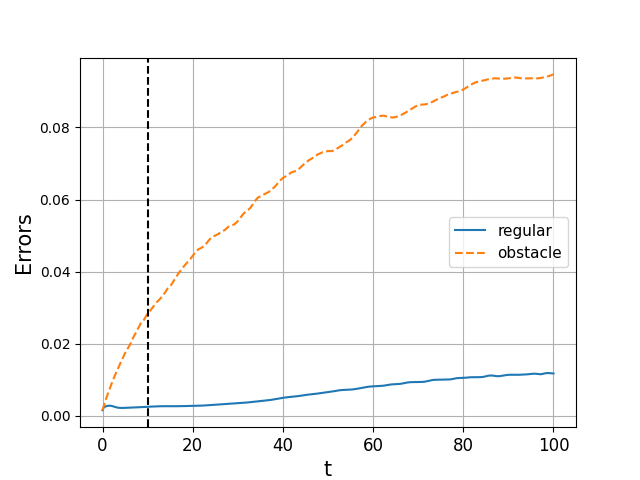}
\caption*{\scriptsize{Star-shaped initial}}
\end{subfigure}
\begin{subfigure}{0.4\textwidth}
\centering
\includegraphics[width=1\textwidth]{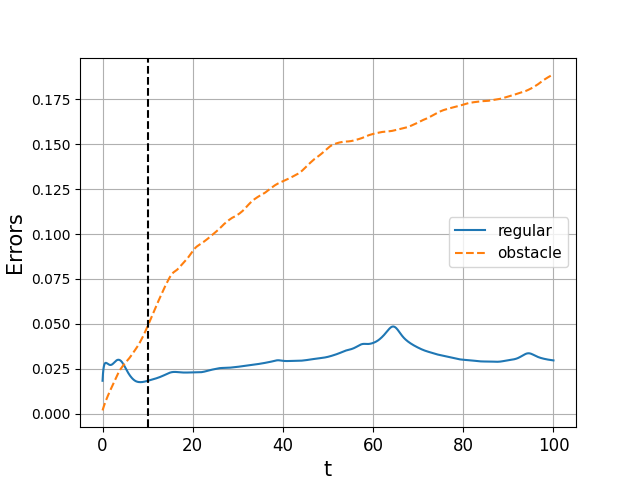}
\caption*{\scriptsize{Sharp-colored initial}}
\end{subfigure}}
\caption{The NPF-Net prediction solutions and relative $L_2$ errors for solving nonlocal AC equation with regular and obstacle potential function under two different types of initial conditions. Top four rows: plots of the iso-surface (value 0) of the predicted solution at times $T$ = 0,10,50,100; Bottom row: evolution of the relative $L_2$ errors of the predicted solution}
\label{fig:3D}
\end{figure}
To demonstrate the capability of our model, we apply the proposed method to 3D nonlocal AC problems with an interfacial parameter of $\epsilon = 0.05$. The same network architectures and training strategies used for the 2D problems are extended to construct NPF-Net for 3D problems. Consequently, the input $U_n$ is now a $1 \times N_x \times N_y \times N_z$ tensor, and the filter size in each convolutional layer changes to $K \times K \times K$ with $K=3$. We set $\delta = 0.1$, $\Delta t = 0.1$, and $T_{train} = 10$, while the spatial mesh size is set to $h = 2/64$.
To test NPF-Net for solving 3D nonlocal problems, we consider two types of initial conditions. One is the star-shaped initial condition, and the other is the sharp-colored noise initial condition. Figure~\ref{fig:3D} shows the iso-surface (value 0) of the predicted solutions and relative $L_2$ errors by NPF-Net at time instances $t = 0, 10, 50, 100$ for the nonlocal AC equation with both regular and obstacle potential functions. The results clearly illustrate the grain coarsening process in both cases. Due to the sharper interface in the obstacle potential function, the error is larger compared to the regular potential function.

\section{Concluding remarks}\label{sec:conclusion}
This paper introduces NPF-Net, an end-to-end deep learning method for solving nonlocal Allen-Cahn (AC) and Cahn-Hilliard (CH) phase-field models. Our approach effectively handles both diffuse and sharp interfaces, learns fully discrete operators without ground-truth data, and demonstrates excellent performance across several potential functions and initial conditions. The time-adaptive training strategy significantly reduces training time while maintaining accuracy and the model shows strong generalization capabilities. Our method is the incorporation of a bound limiter in the network architecture and the use of projection formulas in the loss function. These features allow NPF-Net to efficiently handle non-smooth potential functions and capture sharp interfaces which are challenging aspects in both traditional numerical methods and existing machine learning approaches. This capability enables more realistic modeling of complex material behaviors, particularly for systems with distinct phase boundaries or discontinuous processes.
Despite their promising performance, NPF-Net can be improved in several aspects.
For example, the current method is designed for a fixed horizon $\delta$, which means that NPF-Net needs to be retrained once $\delta$ is changed. This limits the generalization of NPF-Net. To resolve this issue, we will treat $\delta$ and/or other parameters in the target equation as additional inputs to the NPF-Net model such that the trained NPF-Net, as a function of $\delta$, can be used to solve the nonlocal AC and CH equations with different horizons. The network architecture and the training strategy need to be significantly modified to ensure the model can learn the relationship between $\delta$ and the solution dynamics across a wide range of $\delta$ values. Moreover, we will extend the method to solve nonlocal mass-conserving Cahn-Hilliard equations which are crucial for many physical applications. 

The NPF-Net strategy we have developed for the nonlocal AC and CH settings can be applied to many other applications involving phase-field model. Because of its flexibility and scope, it is natural to expect that the gains effected by NPF-Net for the AC and CH settings could extend to other applications. To provide further evidence about the improvement gains of NPF-Net, we will, in the future, apply that approach to other applications for which we have in hand nonlocal models and their discretizations and which incorporate phase fields. Specifically, we will first consider alloy solidification models
\cite{burkovska2023non,burkovska2021nonlocal,burkovska2023khoda} and
superconductivity models \cite{superconductivity2,superconductivity3,superconductivity4,superconductivity1} to which we apply NPF-Net methodology. These activities will provide further evidence about the gains effected by using that methodology.

\section{Acknowledgement}
This work is supported by the U.S. Department of Energy, Office of Science, Office of Advanced Scientific Computing Research, Applied Mathematics program, under the contracts ERKJ388 and ERKJ443. 
ORNL is operated by UT-Battelle, LLC., for the U.S. Department of Energy under Contract DE-AC05-00OR22725. {Lili Ju and Max Gunzburger acknowledge the support from the U.S. Department of Energy, Office of Science, Office of Advanced Scientific Computing Research program under grants DE‐SC0023171 and DE-SC0025527, respectively.}


\end{document}